\documentclass[11pt]{article}

\usepackage{acl}

\usepackage{times}
\usepackage{latexsym}

\usepackage[T1]{fontenc}

\usepackage[utf8]{inputenc}

\usepackage{microtype}

\usepackage{inconsolata}

\usepackage{graphicx}
\usepackage{amsmath}
\usepackage{multirow}

\usepackage{array}
\usepackage{booktabs}
\usepackage[most]{tcolorbox}
\usepackage{fancyvrb}
\usepackage{float}

\usepackage{url}
\newtcolorbox[auto counter, number within=section, list inside=prompts]{promptbox}[2][]{
    enhanced,
    colback=blue!2,
    colframe=blue!40!black,
    colbacktitle=blue!15!white,
    coltitle=blue!40!black,
    fonttitle=\bfseries\small,
    fontupper=\footnotesize\ttfamily,
    boxrule=0.6pt,
    arc=2pt,
    left=6pt, right=6pt, top=4pt, bottom=4pt,
    toptitle=2pt, bottomtitle=2pt,
    title={Prompt~\thetcbcounter\ -- #2},
    #1
}

\newtcolorbox[auto counter, number within=section, list inside=schemas]{schemabox}[2][]{
    enhanced,
    colback=green!2,
    colframe=green!35!black,
    colbacktitle=green!15!white,
    coltitle=green!35!black,
    fonttitle=\bfseries\small,
    boxrule=0.6pt,
    arc=2pt,
    left=6pt, right=6pt, top=4pt, bottom=4pt,
    toptitle=2pt, bottomtitle=2pt,
    title={Schema~\thetcbcounter\ -- #2},
    #1
}

\newtcolorbox[auto counter, number within=section, list inside=exhibits]{exhibitbox}[2][]{
    enhanced,
    colback=gray!4,
    colframe=gray!50!black,
    colbacktitle=gray!18!white,
    coltitle=black,
    fonttitle=\bfseries\small,
    boxrule=0.6pt,
    arc=2pt,
    left=6pt, right=6pt, top=4pt, bottom=4pt,
    toptitle=2pt, bottomtitle=2pt,
    title={Exhibit~\thetcbcounter\ -- #2},
    #1
}

\title{PSEBench: A Controllable and Verifiable Benchmark for Evaluating LLMs in Patient Safety Event Triage
}

\author{%
  Keqi Han\textsuperscript{1}\thanks{\ Equal contribution.}\thanks{\ Work done during internship at Scale AI.} \quad
  Ryan Young\textsuperscript{2}\footnotemark[1] \quad
  Annabel Strauss\textsuperscript{2} \quad
  Lindsey Hughes\textsuperscript{3} \quad
  Katharine M. Nesbitt\textsuperscript{3} \\[2pt]
  \bfseries
  Nicole Schueler\textsuperscript{3} \quad
  Che Ngufor\textsuperscript{3} \quad
  Carl Yang\textsuperscript{1} \quad
  Yuan Xue\textsuperscript{2} \quad
  Zhijun Yin\textsuperscript{4} \\[6pt]
  \textsuperscript{1}Emory University \quad
  \textsuperscript{2}Scale AI \quad
  \textsuperscript{3}Mayo Clinic \quad
  \textsuperscript{4}Vanderbilt University Medical Center%
}

\begin{document}
\maketitle
\begin{abstract}
Patient safety event triage, determining whether a clinical event is 
reportable under jurisdiction-specific policy, is a high-stakes task typically performed manually by patient safety experts. Although LLMs may support this workflow, reliable evaluation is limited by the lack of benchmarks to capture evidence-grounded policy reasoning, proactive information seeking for incomplete reports, and principled abstention in irreducibly ambiguous cases. We 
address this gap with a policy-grounded construction methodology centered 
on the \emph{clause card}, a structured representation that factorizes 
regulatory text into auditable decision specifications. Combining clause cards with 
anchor-driven instantiation and closed-loop verification, our  
scalable pipeline produces narratives with by-construction ground truth and naturally supports generating missing information and uncertain variants. We 
instantiate this method on Minnesota's 29 Reportable Adverse Health 
Events, producing PSEBench, a 5,074-case benchmark with an agentic 
evaluation environment. 
Evaluation on 15 representative LLMs reveals consistent capability trends, demonstrates the benchmark’s utility, and identifies actionable gaps toward reliable LLM-based patient safety event triage.
\end{abstract}

\section{Introduction}

\begin{figure*}[t]
    \centering
    \includegraphics[width=1\linewidth]{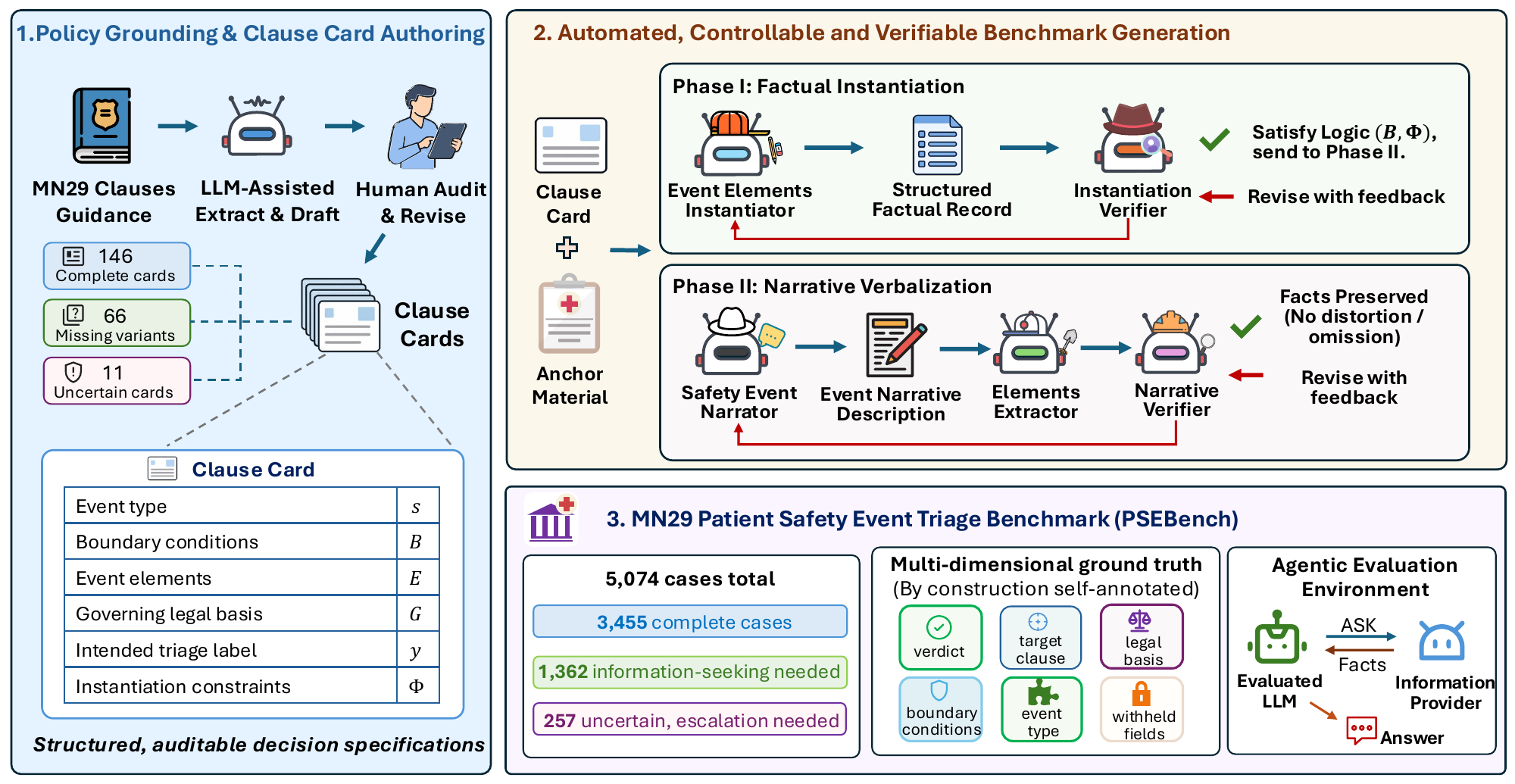}
    \caption{PSEBench: A Controllable and Verifiable Benchmark for Evaluating
LLMs in Patient Safety Event Triage.}
    \label{fig:framework}
\end{figure*}

Patient safety event (PSE) reporting systems are a
foundational component of hospital operation \citep{gong2022challenges}, helping detect safety problems and drive  quality improvement. These systems collect large volumes of free-text reports that describe clinical events, their immediate  
consequences, actions taken, and suspected causes. 
The reports require daily expert triage to determine whether each event is reportable under formal legal and administrative policy frameworks, 
such as Minnesota’s 29 Reportable Adverse Health Events 
(MN29) \citep{mdh_mn29_adverse_events}. 
Currently, this triage process remains largely manual, imposing substantial operational and 
cognitive burden on safety event  experts, which naturally motivates the exploration of 
large language models (LLMs) as potential autonomous triage assistants.

Prior PSE research has primarily focused on isolated descriptive classification tasks, such as predicting event type, harm severity \citep{wang2017classification,evans2020incident,schaeferle2025automating},
medication error category, or contributing factors \citep{boxley2023medication,tabaie2023contributing}. 
They do not capture the triage performed by safety event experts, which requires grounding decisions in policy and evidence, identifying missing decision-critical facts, and abstaining when the facts or policy criteria are ambiguous. As a result, realistic benchmarks are needed to evaluate whether LLMs can comprehensively support policy-grounded PSE triage rather than merely classify event reports.


However, building such a benchmark is challenging. First, policy-grounded triage requires rich supervision, and obtaining such annotations through conventional human expert annotation would be costly and difficult to scale. 
Second, 
although LLM-generated data and instruction synthesis have enabled scalable dataset construction~\citep{wang2023selfinstruct,long2024llms}, high-stakes regulatory triage requires stronger guarantees of policy faithfulness, auditability, and evidence traceability. 
Third, 
PSE triage is not always a single-turn decision. Some reports may omit facts required to determine reportability, while others may remain indeterminate even after the relevant facts are known because the governing policy is silent or ambiguous. These challenges align with recent interactive medical benchmarks that emphasize follow-up questioning and abstention under uncertainty~\citep{li2024mediq,Abstain}. 

To address these challenges, we propose a policy-grounded benchmark construction methodology that combines scalable generation with explicit controllability and verification. Our approach centers on human-audited \emph{clause cards}, structured decision specifications that decompose regulatory guidance into reportability criteria, boundary conditions, required facts, and expected verdicts.
Using these clause cards alongside diverse real-world clinical 
\emph{anchor materials}, an LLM generates event narratives, while automated verifiers check both the structured facts and generated text against the clause card. This closed-loop process reduces silent label drift, ensuring that the generated 
narratives faithfully inherit the triage verdict, legal basis, 
and decision-critical facts of its clause card without 
requiring expensive post-hoc manual annotation. 
Furthermore, the clause-card structure also enables systematic construction of \emph{missing-information} and \emph{uncertain} variants. Finally, we introduce a two-role \emph{agentic evaluation environment} in which the evaluated LLM interacts with an information provider over multiple turns, enabling assessment of follow-up questioning and abstention under uncertainty.

We instantiate this methodology to create the first 
comprehensive benchmark for patient safety event triage, 
grounded in Minnesota's 29 Reportable Adverse 
Health Events (MN29) policy. By seeding the 
generation with authentic medical incident 
reports, we produce a large-scale dataset 
comprising 5,074 cases across complete, 
missing-information, and uncertain scenarios, 
with each case bundled with rich multi-dimensional 
ground truth. We evaluate a suite of representative LLMs 
within our agentic environment. Our analysis reveals 
that while models achieve moderate triage accuracy, 
they still exhibit significant room for improvement in evidence grounding, 
proactive information seeking, and uncertainty-aware escalation, underscoring the 
gap between current LLM 
capabilities and the rigorous demands of real-world clinical safety event management.

\section{Method}

\subsection{Problem Formulation}
\label{sec:problem-formulation}

Given an event description $x$ (a free-text 
clinical narrative) and a reporting policy 
$\mathcal{P}$ (a corpus of statutory clauses and accompanying guidance), 
a triage agent must produce a structured output 
$o = (y,\, t,\, G,\, r)$ comprising: 
(i) a triage verdict $y \in \{\text{Reportable}, \text{Non-Reportable}, \text{Uncertain}\}$; 
(ii) the targeted reportable clause $t$ when $y = \text{Reportable}$; 
(iii) the cited legal basis $G$; and 
(iv) a rationale $r$ that articulates the reasoning. 

Our benchmark targets three case regimes: 
\textbf{(a) complete}, where $x$ contains all decision-critical facts without ambiguity; 
\textbf{(b) missing-information}, where decision-critical facts are absent 
from $x$ and must be recovered before committing to a 
\textsc{Reportable} or \textsc{Non-Reportable} verdict; and 
\textbf{(c) uncertain}, where $x$ is factually complete but touches on 
a policy silence or contradiction.

\subsection{Clause Cards}
\label{sec:clause-card}

Recent work has explored using LLMs to scale up
data creation \cite{wang2023selfinstruct}.
However, high-stakes 
domains such as PSE triage are governed by intricate rules. 
A single policy clause typically intertwines scope restrictions, factual triggers and explicit exceptions. 
Directly prompting an LLM to generate narratives 
from such dense text is unreliable and tends to introduce misalignment between surface text 
and underlying legal logic, leading to silent label drift \cite{long2024llms}. 

To overcome these limitations, we introduce \emph{clause cards}, a
structured decision specification that mediates between a policy clause
and the benchmark instances generated from it. A clause card explicitly
factorizes the latent decision logic of a clause into a set of
named, auditable variables. Each clause card describes one decision
region of the policy, meaning a set of cases that all share the same intended
label for the same legal reasons. It is reused as a generative
template from which many diverse event narratives can be produced.

Formally, a clause card is a tuple
$c \;=\; (s,\; B,\; E,\; \rho,\; y,\; G,\; \Phi),$
where (i) $s$ identifies the specific event type that the card
covers; (ii) $B$ is a set of \emph{boundary conditions}, which are
logical predicates that determine reportability; (iii) $E$ is a set of \emph{basic event elements},
which are concrete, factual slots that constitute the event; (iv) $\rho : B \to 2^{E}$ maps each boundary
condition to the basic event elements that concretize it; (v) $y$ is the intended triage
label of the card; (vi) $G$ is a set of canonical identifiers pointing
to the clause, guidance, or appendix passages that serve as the legal
basis for $y$; and (vii) $\Phi$ is a set of \emph{instantiation constraints}
over $E$, ensuring internal coherence among the basic event elements and
their consistency with $B$. 

By design, $B$ is necessary and jointly sufficient for $y$, 
giving the clause card a dual role in our pipeline. As a
\emph{control surface}, $B$ and $E$ expose exactly the levers a
generator may pull to produce diverse but label-preserving event
narratives. As a \emph{verification surface}, the same variables let
downstream verifiers automatically check whether a generated narrative
realizes the intended facts and supports the intended boundary-condition
truth values, and whether evaluated models cite legal evidence
consistent with $G$.

Clause cards are constructed through an
LLM-assisted, human-in-the-loop protocol to make
specification authoring scalable without sacrificing legal faithfulness.
We first fix the clause-card schema and prepare a detailed authoring guideline.
For each policy clause, the process begins with \emph{loyal extraction}, where the LLM isolates
relevant definitions, exceptions, and thresholds from the source policy
without interpretation. 
Then, the LLM performs \emph{structured drafting} to produce a set of candidate
clause cards that are mutually exclusive and exhaustive for the clause.
Afterward, humans rigorously audit and refine these drafts, 
ensuring that the boundary conditions $B$
are necessary and jointly sufficient for $y$, and that the event elements $E$
are purely factual and properly concretize $B$ via $\rho$. Thus, the LLM
amortizes the heavy lifting of structural drafting, while humans 
finalize the reusable decision specifications. This division of labor allows
the pipeline to remain auditable as it scales.


\subsection{Controllable and Verifiable Benchmark Generation}
\label{sec:generation-pipeline}

We decompose 
generation into a rigorous, multi-stage pipeline: factual instantiation, narrative 
realization, and automated verification. Because every generated narrative 
must satisfy the structural constraints of its underlying clause card, it inherits the card's triage label and legal evidence. This 
guarantees that the benchmark is strictly policy-grounded, controllable, 
and verifiable. Consequently, it eliminates the need for expensive, post-hoc manual 
annotation while providing the rich, multi-dimensional ground truth essential for 
evaluating LLMs.

\paragraph{Anchor Material as Generative Seed.}
To inject realism and diversity
without compromising the card's semantics, we 
introduce \emph{anchor materials}---real-world de-identified artifacts such
as incident reports, clinical case summaries, or publicly reported adverse events.
The anchor plays three complementary
roles. As a \emph{stylistic prior}, it supplies authentic clinical
register and document structure. As a \emph{scene anchor}, it grounds
the narrative in concrete operational details (settings, roles, devices,
workflows) that LLMs rarely produce unprompted. Furthermore, it
functions as a \emph{diversity seed}: 
by varying anchor materials while holding clause card $c$ fixed, 
we obtain 
different in-distribution realizations of the
card's factual slots $E$, 
dramatically increasing the benchmark's surface diversity. 

\paragraph{From Clause Card to Verified Narrative.}
The generation proceeds in two main phases. First, 
an \emph{Instantiator} takes a clause card and an anchor material, 
populating the basic event elements ($E$) with concrete, scenario-specific 
facts. At this stage, 
the output is purely a 
structured factual record. This record is immediately audited by 
an \emph{Instantiation Verifier}, which checks whether the generated 
facts strictly satisfy the intended truth values of the boundary conditions ($B$) 
and the instantiation constraints ($\Phi$). 
If an instantiation violates the card's decision logic, the verifier returns feedback on the failed conditions for the instantiator to revise. This process repeats until the instantiation passes verification or reaches a retry limit (e.g., three attempts). This ensures that only policy-consistent structured cases are used for narrative generation.

In the second phase, an \emph{Event Narrator} weaves the verified factual record into 
a cohesive clinical event narrative, with the anchor material providing stylistic and scene-level grounding. 
An LLM might omit 
critical details or hallucinate a nuance that shifts the event across a legal boundary. 
To prevent this silent label drift, we introduce a closed-loop verification step. 
An \emph{Extractor} reads the generated narrative and attempts to parse out the 
underlying event elements. A \emph{Narrative Verifier} then cross-checks 
these extracted facts against the original instantiated record to ensure 
the narrative faithfully encodes the intended event without omission or distortion.
Only narratives that survive this automated verification 
are admitted into the benchmark.

\paragraph{Missing-Information Variants:} 
In practice, initial reports are often incomplete, requiring safety experts to seek additional information (e.g., from electronic health records (EHRs)) before making a triage decision. To evaluate this ability, a missing-information case is constructed by withholding a decision-critical subset of the basic event elements $E' \subseteq E$, such that the remaining evidence is compatible with both reportable and non-reportable interpretations. The narrative is then re-rendered and verified using only the remaining fields. 
Each clause card specifies such policy-grounded omissions through a \texttt{missing\_information\_variants} field, authored under the same LLM-assisted, human-in-the-loop protocol. Since the withheld elements are explicitly tracked, each case provides ground truth for the missing facts that an LLM agent should request.
    

    
\paragraph{Uncertain (Gray-Zone) Cases:} 
Some events naturally fall into regulatory gray zones where the policy text is inherently ambiguous or contradictory. In such cases, an LLM agent should escalate it for human review rather than force a binary reportable/non-reportable decision.  
To evaluate this ability, we introduce cases where the clinical facts are completely known, yet the legal classification remains irreducibly ambiguous. These cases are governed by dedicated \emph{uncertain clause cards}, which share the exact same structural schema as standard clause cards but are purposefully designed to capture specific policy silences or contradictions. Like the missing variants, they are authored via LLM-assisted extraction and subsequent human audit. Since the underlying schema is identical, uncertain cases seamlessly reuse the standard two-phase generation pipeline.


\subsection{MN29 Patient Safety Triage Benchmark}
\label{sec:mn29-benchmark}

We instantiated our pipeline using Minnesota's 29 
Reportable Adverse Health Events (MN29) \citep{mdh_mn29_adverse_events} as the policy testbed. 
MN29 is a comprehensive, state-mandated reporting system.
With detailed official guidance, MN29 provides a highly complex, 
legally binding framework ideal for stress-testing policy-grounded triage. 
Following the LLM-assisted human-in-the-loop authoring protocol of 
Section~\ref{sec:clause-card}, we produce a library of 146 clause cards for complete cases, 66 missing-information 
variants attached to subsets of these cards, and 11 uncertain 
clause cards that capture canonical gray-zone fact patterns. All cards have 
undergone human audit.

\paragraph{Sourcing Anchor Materials.}
We source anchor materials from the 2023 and 2024 \emph{medical accident 
information} releases in the Japan Council for Quality 
Health Care (JQ) database \citep{jq_medsafe_database}, 
which publishes de-identified structured incident reports submitted by Japanese 
healthcare facilities. 
Each report is first 
translated by an LLM from Japanese to English and coarsely 
classified into one of the 29 MN clause categories. This 
classification is intentionally a topical pre-sort rather than a reportability judgment. At generation time, anchors for a given clause card are sampled only from its matching topical bucket. This 
soft topical alignment keeps the anchor's scene broadly consistent with the kind of event that the clause card targets, which we found materially improves both 
the generation efficiency and the realism of the 
final narratives.

\paragraph{Generated Benchmark Cases.}
Due to the budget constraints, for each clause card, we sampled 25 distinct anchor materials 
to drive the generation pipeline (with missing variants 
inheriting the anchors of their parent complete cases). 
This process yielded a benchmark comprising 
3,455 complete cases, 1,362 missing-information cases, and 257 uncertain cases.
The generation success rates---defined as the proportion of 
initiated generations that successfully survived the 
closed-loop verification process---were 94.7\%, 89.5\%, 
and 93.5\% for complete, missing, and uncertain cases, respectively.
Failures are candidates rejected by the verifier because the
instantiated facts or narrative text did not satisfy the intended
specification. 

Notably, our benchmark cases are not 
released as bare \emph{(narrative, label)} pairs. 
Instead, each case is paired with structured ground-truth annotations derived from clause cards, including the final triage verdict, event type, governing policy basis, and boundary conditions that determine the decision logic. Missing-information variants also explicitly track the withheld fields. These rich, 
by-construction annotations directly enable 
comprehensive multi-dimensional evaluation of LLMs 
without requiring any post-hoc manual labeling. Generation details can be found in Appendix \ref{app:benchmark-construction}.

\paragraph{Safety Expert Validation.}
To validate benchmark quality, two patient safety experts jointly reviewed a sample of 90 generated cases (30 each of complete, missing, and uncertain). On a 1--5 scale, the narratives achieved high ratings for clinical realism (mean 4.50--4.63) and internal plausibility (mean 3.87--4.20) across all types. Crucially, the experts strongly concurred with the pipeline's by-construction ground truth, agreeing with 28/30 complete-case verdicts, 29/30 missing-case indeterminacies, and 30/30 uncertain-case ambiguities. This indicates our pipeline produces realistic, policy-faithful triage scenarios. 

\subsection{Agentic Evaluation Environment}
\label{sec:eval-env}


\paragraph{Two-Role Architecture.}
We model the triage process as an interaction between two LLM-driven roles. 
The first, $\mathcal{A}_{\text{eval}}$, is the LLM under evaluation: 
given an event narrative and the policy text, it ultimately produces a triage verdict (Reportable,
Non-Reportable, or Uncertain), the targeted clause, cited legal evidence, and a rationale. 
$\mathcal{A}_{\text{eval}}$ is equipped with an interface to query an external information source for additional facts. 
The second, $\mathcal{A}_{\text{info}}$, is an automated, 
stateless oracle simulating an EHR query interface.  $\mathcal{A}_{\text{info}}$ holds the \emph{canonical basic event elements} from the clause card for a given case. 
It is instructed to answer \emph{only} queries that align with these known 
facts, and declare "unknown" for any queries outside this scope. 
This prevents $\mathcal{A}_{\text{info}}$  from improvising 
hallucinated information and 
ensuring that $\mathcal{A}_{\text{eval}}$'s reasoning remains strictly grounded in the benchmark's controlled facts.

$\mathcal{A}_{\text{eval}}$ interacts with $\mathcal{A}_{\text{info}}$ through a multi-turn loop. 
At each turn, $\mathcal{A}_{\text{eval}}$ autonomously chooses one of two actions: \texttt{ASK} a specific 
factual question, which is routed to $\mathcal{A}_{\text{info}}$ and the response is 
appended to the conversation, or \texttt{ANSWER} with a final structured 
output committing to a verdict. The decision of whether to ask, 
what to ask, and when to stop asking is entirely $\mathcal{A}_{\text{eval}}$'s discretion, 
and the environment imposes only a turn budget rather than a 
fixed query schedule. We log the full interaction trajectory, including all 
questions and provider responses, and 
the final structured answer. 

\paragraph{Multi-Dimensional Evaluation Metrics.}
Because every case in our benchmark is generated from a clause card 
with comprehensive ground-truth annotations by construction, we can 
evaluate the agent's performance 
using a suite of hard-coded metrics and LLM-as-a-judge scorers. 
We measure \textbf{triage accuracy (M1)} by comparing 
$\mathcal{A}_{\text{eval}}$'s final verdict against the intended verdict over all cases, and 
\textbf{reportable-clause accuracy (M2)} by checking whether the agent identifies the correct target clause on gold-\textsc{Reportable} cases.
For evidence grounding, we assess whether $\mathcal{A}_{\text{eval}}$ 
cites the correct governing legal basis, \textbf{evidence citation F1 (M3)}, and whether 
its rationale successfully hits the necessary boundary conditions, \textbf{boundary condition hit (M4)} through an LLM scorer. 
For missing-information variants, we track whether $\mathcal{A}_{\text{eval}}$ actively queried $\mathcal{A}_{\text{info}}$ by \textbf{missing case detection F1 (M5)}, and
whether its queries successfully targeted the withheld fields, \textbf{missing slot identification F1 (M6)}. We assess \textbf{uncertainty detection F1 (M7)} 
by measuring the $\mathcal{A}_{\text{eval}}$'s ability to correctly escalate 
gray-zone events as "Uncertain". Finally, we report \textbf{reportable-detection F1 (M8)}. This multi-dimensional suite provides 
a highly granular operational profile of an agent's true triage readiness. The full definitions, computation procedures, and judge 
prompts can be found in Appendix \ref{app:metrics}.

\begin{table*}[ht]
\centering
\small
\setlength{\tabcolsep}{4pt}
\renewcommand{\arraystretch}{1.05}
\begin{tabular}{lcccccccc}
\hline
 & \textbf{M1} & \textbf{M2} & \textbf{M3} & \textbf{M4} & \textbf{M5} & \textbf{M6} & \textbf{M7} & \textbf{M8} \\
\textbf{Model} & \textbf{Verdict} & \textbf{Clause} & \textbf{Evidence} & \textbf{Boundary} & \textbf{Missing} & \textbf{Missing Slot} & \textbf{Uncertain} & \textbf{Reportable} \\
 & \textbf{Acc.} & \textbf{Acc.} & \textbf{Citation F1} & \textbf{Hit Rate} & \textbf{Detect.\,F1} & \textbf{Identify. F1} & \textbf{Detect.\,F1} & \textbf{Detect.\,F1} \\
\hline
\multicolumn{9}{l}{\emph{Closed-source frontier}} \\
GPT-5.5 & 94.6 & \textbf{98.5} & \textbf{74.3} & \textbf{88.0} & 82.2 & 77.3 & \textbf{79.7} & 91.7 \\
GPT-5 & 90.2 & 98.0 & 66.4 & 83.9 & 77.7 & 75.0 & 12.8 & 86.9 \\
Claude Opus 4.7 & 92.3 & 97.3 & 67.4 & 78.5 & 83.4 & \textbf{77.6} & 57.1 & 90.9 \\
Claude Sonnet 4.6 & 84.5 & 97.9 & 65.8 & 85.8 & 70.1 & 76.2 & 49.2 & 81.6 \\
Gemini 3.1 Pro & \textbf{95.3} & 97.8 & 65.5 & 78.6 & \textbf{88.2} & 77.4 & 74.0 & \textbf{94.0} \\
Gemini 2.5 Flash & 85.8 & 96.7 & 64.9 & 77.5 & 70.5 & 71.6 & 31.3 & 82.4 \\
\hline
\multicolumn{9}{l}{\emph{Open-source frontier}} \\
DeepSeek-R1 & 83.7 & 95.3 & 56.3 & 76.5 & 66.5 & 66.3 & 36.3 & 82.0 \\
Qwen3-235B & 76.7 & 93.3 & 60.6 & 81.6 & 54.7 & 69.3 & 18.3 & 74.4 \\
GPT-OSS-120B & 79.0 & 94.3 & 45.4 & 68.5 & 36.5 & 71.6 & 10.0 & 71.4 \\
\hline
\multicolumn{9}{l}{\emph{Small models}} \\
GPT-5-nano & 71.9 & 87.2 & 41.1 & 72.5 & 35.3 & 61.0 & 10.1 & 69.4 \\
Llama-3.1-8B & 52.4 & 59.7 & 30.7 & 62.1 & 46.7 & 24.6 & 10.5 & 59.5 \\
Mistral-Small-3.2 (24B) & 60.5 & 89.5 & 49.4 & 71.4 & 27.7 & 53.5 & 2.2 & 60.5 \\
\hline
\multicolumn{9}{l}{\emph{Medical-specialty}} \\
HuatuoGPT-o1 (8B) & 45.5 & 48.8 & 13.6 & 58.7 & 0.3 & 0.0 & 2.3 & 51.9 \\
MedGemma (27B) & 51.3 & 77.4 & 43.8 & 73.8 & 0.6 & 37.5 & 0.0 & 54.2 \\
HuatuoGPT-o1 (70B) & 61.1 & 84.1 & 51.7 & 67.4 & 2.3 & 56.6 & 3.6 & 60.5 \\
\hline
\end{tabular}

\caption{Overall performance of 15 LLMs on the MN29 Patient Safety Triage benchmark. 
All scores are percentages \%; per-column best scores are in \textbf{bold}. 
}

\label{tab:overall_results}
\end{table*}

\section{Experiments and Results}
\label{sec:experiments}

\subsection{Experimental Setup}
\label{sec:exp-setup}

\paragraph{Tested LLMs.} We evaluate 15
representative LLMs spanning four families:
(i)~\emph{closed-source frontier} models: GPT-5.5, GPT-5,
Claude Opus 4.7, Claude Sonnet 4.6, Gemini 3.1 Pro, and Gemini 2.5
Flash; (ii)~\emph{open-source frontier} models: DeepSeek-R1,
Qwen3-235B, and GPT-OSS-120B; (iii)~\emph{small} general-purpose
models: GPT-5-nano, Llama-3.1-8B, and Mistral-Small-3.2 (24B); and
(iv)~\emph{medical-specialty} models that have undergone clinical
continued pre-training or instruction tuning: HuatuoGPT-o1
(8B and 70B) \citep{chen2024huatuogpto1medicalcomplexreasoning} and MedGemma (27B). To isolate the agent's reasoning
behavior from confounding choices, the $\mathcal{A}_{\text{info}}$ and the LLM judges are fixed across
runs to GPT-5.2. Details are listed in
Appendix \ref{app:llms-infra}.


\subsection{Overall Performance}
\label{sec:exp-main}

Table~\ref{tab:overall_results} reports the  performance of
all 15 LLMs. Overall, the results are consistent with established model capability trends, with the highest-ranked systems corresponding to the most capable and recently released models, validating the reasonability of our benchmark construction pipeline.

Our evaluation reveals a stark divergence between models' static classification capabilities and their interactive triage readiness. Closed-source frontier models (e.g., Gemini 3.1 Pro, GPT-5.5) dominate the leaderboard, achieving high verdict accuracy; however, this top-line metric masks severe inconsistencies in agentic behaviors. For instance, while GPT-5 and GPT-5.5 both achieve strong verdict accuracy, GPT-5 almost entirely fails to escalate ambiguous cases (M7 drops from $79.7\%$ to $12.8\%$). Open-source frontier models (e.g., DeepSeek-R1, Qwen3-235B) approach the verdict accuracy of mid-tier closed models (M1 in the mid-to-high 70s and 80s) but suffer precipitous drops in proactive information-seeking, with missing detection (M5) scores falling as low as $36.5\%$. Finally, small general-purpose models ($<30$B) lag uniformly across all dimensions and exhibit degenerate interactive profiles. Together, these results demonstrate that as models scale down or shift from proprietary to open-weight families, their capacity for proactive information-seeking and principled abstention degrades far faster than their basic verdict accuracy, showing that simple classification performance is an unreliable proxy for real-world triage readiness. Additional results and analysis are provided in Appendix \ref{app:ext-results} .

We also observe that medical-specialty models are ill-suited for policy-grounded PSE triage. 
All three models (HuatuoGPT-o1 8B/70B, MedGemma
27B) sit at the bottom of the leaderboard. Their verdict
accuracy (M1: 45.5--61.1) is comparable to small general-purpose
models, but their weaknesses are more pronounced on metrics that distinguish safety-management triage from standard clinical QA: missing-information handling (M5/M6) and uncertainty-aware abstention (M7). Scaling within the Huatuo family improves verdict accuracy modestly, with Huatuo-70B increasing M1 by $15.6$ points over Huatuo-8B, 
but does not unlock proactive information seeking or
uncertainty-aware abstention---behaviors that an agent
must exhibit to be deployed safely. These results suggest that medical-specialty training may help with static clinical QA but does not by itself produce the interactive, uncertainty-aware behavior required for policy-grounded PSE triage.


\subsection{Multi-Dimensional Diagnosis}
\label{sec:exp-multidim}

\paragraph{Verdict and Clause Accuracy.}
Performance by case type reveals a sharp degradation as the task moves beyond complete reports. As shown in Table~\ref{tab:m1_by_casetype}, even the strongest model, GPT-5.5, drops from 97.8\% accuracy on complete cases to 66.5\% on uncertain cases. The decline is also severe for smaller or domain-specialized models: Llama-3.1-8B falls from 57.5\% to 29.2\%, and HuatuoGPT-o1-8B drops from 52.9\% to 1.9\%. 
Table~\ref{tab:behavioral_profile} partially explains this pattern, suggesting an over-reporting bias, where models tend to force a ``Reportable'' verdict rather than abstain under ambiguity. In addition, Verdict Accuracy (M1) consistently trails Clause Accuracy (M2), indicating that identifying the precise governing policy clause is easier than reaching the correct reportability outcome.

\begin{table}[t]
\centering
\small
\setlength{\tabcolsep}{4pt}
\renewcommand{\arraystretch}{1.05}
\begin{tabular}{lccc}
\hline
\textbf{Model} & \textbf{Complete} & \textbf{Missing} & \textbf{Uncertain} \\
& {\footnotesize ($n{=}3{,}455$)} & {\footnotesize ($n{=}1{,}362$)} & {\footnotesize ($n{=}257$)} \\
\hline
\multicolumn{4}{l}{\emph{Closed-source frontier}} \\
GPT-5.5 & 97.8 & 91.9 & \textbf{66.5} \\
GPT-5 & 97.0 & 88.7 & 7.8 \\
Claude Opus 4.7 & 95.8 & 89.9 & 56.8 \\
Claude Sonnet 4.6 & 88.3 & 82.0 & 46.3 \\
Gemini 3.1 Pro & \textbf{98.6} & \textbf{92.4} & 65.4 \\
Gemini 2.5 Flash & 92.7 & 79.2 & 26.8 \\
\hline
\multicolumn{4}{l}{\emph{Open-source frontier}} \\
DeepSeek-R1 & 92.4 & 70.0 & 37.7 \\
Qwen3-235B & 82.6 & 72.2 & 21.0 \\
GPT-OSS-120B & 88.8 & 67.9 & 6.2 \\
\hline
\multicolumn{4}{l}{\emph{Small models}} \\
GPT-5-nano & 82.0 & 57.3 & 12.5 \\
Llama-3.1-8B & 57.5 & 43.8 & 29.2 \\
Mistral-Small-3.2 & 71.8 & 42.9 & 1.6 \\
\hline
\multicolumn{4}{l}{\emph{Medical-specialty}} \\
HuatuoGPT (8B) & 52.9 & 34.9 & 1.9 \\
MedGemma (27B) & 60.3 & 38.0 & 0.0 \\
HuatuoGPT (70B)  & 72.9 & 42.2 & 2.3 \\
\hline
\end{tabular}
\caption{Verdict accuracy (M1) stratified by case type. Numbers are percentages; per-column best is in \textbf{bold}.}
\label{tab:m1_by_casetype}
\end{table}

\paragraph{Evidence Grounding.}

Evaluation shows that obtaining the correct triage verdict does not guarantee policy-faithful reasoning. 
As shown in Table~\ref{tab:overall_results}, strong models like GPT-5.5, Claude Opus 4.7, and Gemini 3.1 Pro 
achieve impressive overall verdict accuracy (exceeding 92\%), yet their performance consistently drops 
on Evidence citation (M3) and Boundary condition hit rate (M4). 
This gap emphasizes that evaluating PSE 
triage purely via final verdict accuracy overstates 
true readiness; models often guess the right verdict 
for the wrong or hallucinated reasons. Requiring 
robust evidence grounding is essential to build clinical trust.

\paragraph{Proactive Information Seeking.}
\label{sec:exp-information-seeking}

\begin{figure}[t]
    \centering
    \includegraphics[width=\linewidth]{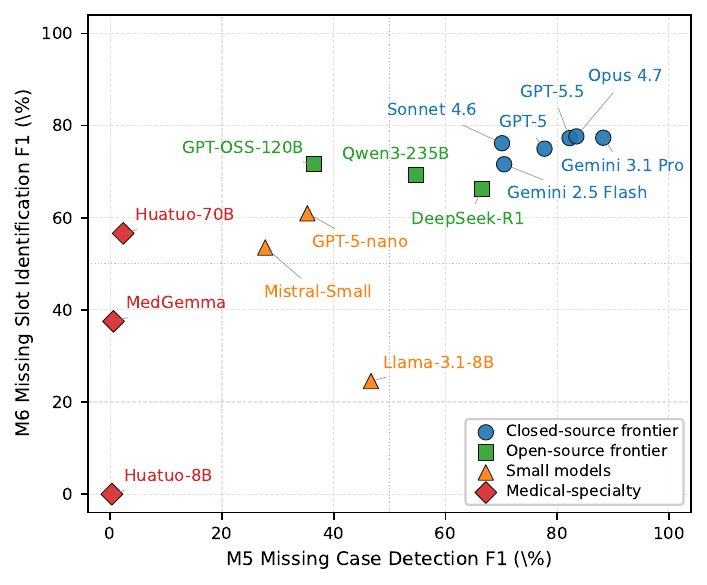}
    \vspace{-0.2in}
    \caption{Agentic Information Seeking: Missing case detection vs.\,Missing slot identification. Models in the upper right actively ask and successfully target withheld facts. Models in the lower right ask indiscriminately. Models on the left commit without asking.}
    \label{fig:m5_m6_scatter}
\end{figure}

When incident reports omit decision-critical facts, a 
deployable triage agent must proactively query Information Provider 
rather than hallucinating an assumption. We measure 
the capability to \emph{decide whether to ask} via Missing Detection F1 (M5) 
and the capability to \emph{target the right factual gap} via Missing Slot Identification 
F1 (M6). Figure~\ref{fig:m5_m6_scatter} maps 
the models across these two dimensions. The closed-source frontier models 
(e.g., GPT-5.5, Gemini 3.1 Pro) occupy the 
upper-right quadrant, successfully recognizing 
missing facts and formulating targeted questions: 
Table~\ref{tab:behavioral_profile} shows that they 
refuse to ask on only $9.4$--$16.0\%$ of missing cases. 
In contrast, Llama-3.1-8B exhibits an 
``ask indiscriminately'' failure mode: it asks on most 
missing cases (no-ask$=13.7\%$), yet 
its low missing slot detection score ($24.6\%$ in Table~\ref{tab:overall_results}) 
reveals that these queries are poorly targeted to the 
withheld facts. At the other extreme, medical-specialty models 
occupy the far-left edge with a severe ``commit-without-asking'' 
prior: HuatuoGPT-o1 (8B), MedGemma, and even the much larger 
HuatuoGPT-o1 (70B) issued zero {\tt ASK}s on $99.9\%$, $99.7\%$, 
and $98.8\%$ of missing cases respectively. As discussed in 
Section~\ref{sec:exp-main}, this suggests that current clinical 
instruction-tuning pipelines inadvertently penalize clarifying 
behaviors, favoring confident, single-turn responses even when 
the input is dangerously underspecified.

\paragraph{Uncertainty Awareness}
The ability to escalate ambiguous, gray-zone events (M7) 
is the most polarized capability. 
Only 
the strongest models demonstrate meaningful uncertainty 
awareness (GPT-5.5 at 79.7\%, Gemini 3.1 Pro at 74.0\%). 
Most other models, including GPT-5 (12.8\%) and all 
medical-specialty models ($\leq 3.6\%$), fail entirely on this dimension. Right block of Table~\ref{tab:behavioral_profile} reveals 
the dominant failure mode: \emph{overconfidence toward 
``Reportable''}. When presented 
with an irreducible policy ambiguity, models overwhelmingly 
force a ``Reportable'' verdict. HuatuoGPT-o1 (8B), for 
instance, incorrectly classified 230 out of 257 uncertain 
cases as Reportable; Llama-3.1-8B did so for 167 cases. 
This ``high sensitivity, low specificity'' bias is 
particularly problematic in PSE triage, as a high 
false-alarm rate quickly erodes the utility of an 
automated triage assistant, overwhelming safety managers with spurious reportable flags.

\begin{table}[t]
    \centering
    \footnotesize
    \setlength{\tabcolsep}{4pt}
    \renewcommand{\arraystretch}{1.05}
    \begin{tabular}{lrrr}
    \hline
    \multirow{2}{*}{\textbf{Model}} & \textbf{Miss.} & \multicolumn{2}{c}{\textbf{Uncertain (\%)}} \\
    \cline{3-4}
    & \textbf{no-ask\%} & \textbf{$\to$Unc} & \textbf{$\to$Rep} \\
    \hline
    \multicolumn{4}{l}{\emph{Closed-source frontier}} \\
    GPT-5.5            & 16.0 & 66.5 & 32.3 \\
    GPT-5              & 26.2 &  7.8 & 72.4 \\
    Claude Opus 4.7    & 12.0 & 56.8 & 40.1 \\
    Claude Sonnet 4.6  & 21.0 & 46.3 & 52.1 \\
    Gemini 3.1 Pro     &  9.4 & 65.4 & 23.7 \\
    Gemini 2.5 Flash   & 40.2 & 26.8 & 66.9 \\
    \hline
    \multicolumn{4}{l}{\emph{Open-source frontier}} \\
    DeepSeek-R1        & 46.2 & 37.7 & 54.9 \\
    Qwen3-235B         & 33.6 & 21.0 & 69.3 \\
    GPT-OSS-120B       & 77.1 &  6.2 & 75.5 \\
    \hline
    \multicolumn{4}{l}{\emph{Small models}} \\
    GPT-5-nano         & 77.5 & 12.5 & 82.9 \\
    Llama-3.1-8B       & 13.7 & 29.2 & 65.0 \\
    Mistral-S-3.2 (24B)& 82.0 &  1.6 & 90.3 \\
    \hline
    \multicolumn{4}{l}{\emph{Medical-specialty}} \\
    HuatuoGPT (8B)     & 99.9 &  1.9 & 89.5 \\
    MedGemma (27B)     & 99.7 &  0.0 & 98.4 \\
    HuatuoGPT (70B)    & 98.8 &  2.3 & 87.2 \\
    \hline
    \end{tabular}
    \caption{Behavioral profile across the 15 models. \textbf{Miss.\ no-ask\%}: fraction of the 1{,}362 missing cases on which the model issued no {\tt ASK} at all. \textbf{Uncertain $\to$Unc / $\to$Rep}: percentage of the 257 uncertain cases routed to \textsc{Uncertain} or \textsc{Reportable} 
    }
    \label{tab:behavioral_profile}
\end{table}

\section{Related Work}

\paragraph{NLP for patient safety incident reports.}
Prior NLP work on patient safety incident reports has mainly targeted descriptive classification, including event type, severity \citep{wang2017classification,evans2020incident,schaeferle2025automating}, medication-error categories, contributing factors \citep{boxley2023medication,tabaie2023contributing} and trend analysis \citep{young2019systematic,chen2024human}. These tasks support surveillance and quality improvement, but do not address jurisdiction-specific, policy-grounded triage decisions involving legal evidence, missing facts, and abstention.


\paragraph{LLM-assisted benchmark construction.}
LLMs have been widely used to synthesize instructions and evaluation data \citep{wang2023selfinstruct,honovich2023unnatural,long2024llms}. However, in regulated medical settings, a narrative can be clinically fluent while failing to satisfy the legal conditions. PSEBench addresses this limitation with clause cards and closed-loop verification, which preserve intended labels, evidence, and decision logic.

\paragraph{Interactive clinical evaluation and uncertainty.}
Recent medical evaluation work increasingly emphasizes interaction,
information gathering, and uncertainty handling beyond static QA.
AgentClinic \citep{schmidgall2026agentclinic} and MAI-DxO
\citep{nori2025sequential} advocate clinically realistic evaluation;
MediQ \citep{li2024mediq} and MedAgentBench
\citep{jiang2025medagentbench} evaluate follow-up questioning and
patient-record interfaces; and MedAbstain studies abstention under
medical uncertainty \citep{Abstain}.
To the best of our knowledge, there is no prior 
public benchmark for PSE reportability 
triage that jointly evaluates policy-grounded verdicts, 
legal evidence grounding, 
targeted information seeking over missing facts, and explicit 
recognition of policy-level uncertainty.



\section{Conclusions}

We presented PSEBench, a policy-grounded benchmark for evaluating LLMs in patient safety event triage. Built on auditable clause cards and closed-loop verification, PSEBench generates cases with by-construction ground truth and supports multi-dimensional evaluation. Experiments with 15 LLMs reveal that high verdict accuracy frequently masks critical failures in evidence grounding, proactive information seeking, and reasoning under policy ambiguity. Overall, PSEBench provides a scalable and verifiable testbed for developing safer, more policy-aligned LLM applications for patient safety triage workflows.

\section*{Limitations}

First, although PSEBench uses real-world de-identified safety events as anchor materials to improve clinical realism and scene diversity, the final event narratives are still synthetically generated. Real-world incident reports and raw electronic health record (EHR) notes can exhibit extreme structural disorder, severe grammatical noise, and idiosyncratic abbreviations that PSEBench may not fully capture. At the same time, strict by-construction alignment with complex policy logic requires a controlled generation process, making this trade-off necessary for the present benchmark.

Second, our multi-turn evaluation for \textsc{missing} cases abstracts information seeking into an idealized conversational protocol through the \texttt{Ask} tool. This design allows us to isolate and measure an LLM's intrinsic epistemic awareness, query formulation, and inquiry strategy, without confounding these abilities with interface-specific retrieval failures. However, it does not evaluate whether an agent can navigate a live or highly unstructured EHR system to retrieve the needed information in practice. Future work could extend PSEBench with more realistic tool-use environments that require agents to search, filter, and synthesize evidence from heterogeneous EHR artifacts.

Finally, PSEBench is currently grounded specifically in the MN29 policy framework. While MN29 is a mature, representative state-mandated reporting system, institutional or national guidelines elsewhere may vary in their specific taxonomy and boundary conditions. Nonetheless, our core benchmark construction methodology---factorizing opaque policies into verifiable clause cards and utilizing a closed-loop generation pipeline---is highly generalizable and can be readily adapted to other complex regulatory or compliance frameworks. Future work could instantiate the same construction methodology under additional regulatory frameworks, further testing the generality of clause cards and closed-loop verification for broader compliance-oriented triage tasks.

\section*{Ethical considerations}
We use Minnesota’s 29 Reportable Adverse Health Events policy documents and publicly available JQ de-identified incident-report releases as source materials, following their respective public access and use terms. The released PSEBench code and generated benchmark artifacts will be distributed under the license specified in the accompanying repository.




\bibliography{references}


\appendix

\section{Benchmark Construction Details}
\label{app:benchmark-construction}

This section documents the artifacts and procedures that operationalize
the clause-card-centered construction pipeline described in
\S\ref{sec:clause-card} of the main paper. It covers (i) the full
clause-card schema, (ii) the LLM-assisted
authoring prompts, (iii) the case-generation pipeline prompts, (iv) a
brief description of the MN29 policy and the anchor-material corpus we
draw clinical context from, (v) dataset statistics, (vi) generation
retry-rate statistics, and (vii) one end-to-end generation example.

\subsection{Clause Card Schema}
\label{app:clause-card-schema}

A clause card is serialized as a single JSON document. The top-level
schema is given in Schema~\ref{schema:clause_card}; the role of each
field is documented as an inline comment beside it. Together, the
JSON fields encode the clause-card tuple $(s, B, E, \rho, y, G, \Phi)$
of the main paper: \path{event_type} carries the event-type tag $s$
that identifies which decision region the card covers,
\path{boundary_conditions} is $B$, \path{basic_event_elements} is $E$,
the \path{corresponding_basic_event_elements} list inside each
boundary condition encodes the mapping $\rho:B\!\to\!2^{E}$, the
intended verdict $y$ is determined by $s$,
\path{governing_legal_basis.value} is the legal-basis set $G$ used as
the gold reference for evidence-citation (M3), and
\path{constraints_on_basic_event_elements_instantiation} encodes the
instantiation predicates $\Phi$. The auxiliary \path{clause_id} is
the MN29 sub-clause this card lives under and is used only for
bookkeeping (one clause typically has several cards covering
disjoint decision regions).

\begin{figure*}[t]
\begin{schemabox}[label=schema:clause_card]{Clause-card JSON template}
\begin{Verbatim}[fontsize=\footnotesize, commandchars=\\\{\}]
\{
  "clause_card_id":         <stable identifier of the card>,
  "clause_card_definition": <single-paragraph summary of the decision region the card covers>,
  "clause_id":              <identifier of the MN29 sub-clause this card belongs to>,
  "event_type":             "Reportable" | "Non_Reportable" | "Uncertain",

  "fixed_fields": \{                            // factual constants for this decision region
    "governing_legal_basis": \{                 // gold legal-basis set G; reference for metric M3
      "value":   [<MN29 clause / recommendation / appendix identifier>, ...],
      "meaning": <short rationale linking the listed identifiers to the verdict>
    \},

    "boundary_conditions": \{            // set B; necessary and jointly sufficient for the verdict
      <bc_name>: \{
        "value":   true | false,                // intended truth value under the card's verdict
        "meaning": <definition with MN29 inclusion / exclusion criteria>,
        "corresponding_basic_event_elements": [
          <element name that concretizes this condition (mapping rho)>, ...
        ]
      \},
      ...
    \}
  \},

  "basic_event_elements": \{                    // set E; one entry per element
    <element_name>: \{
      "field_type":         "string" | "string_or_null" | "enum",
      "meaning":            <what concrete fact this element captures>,
      "allowed_content":    <guidance for admissible realizations (illustrative, not literal)>,
      "disallowed_content": <realizations that would violate B or leak the verdict>
    \},
    ...
  \},

  "constraints_on_basic_event_elements_instantiation": [
    <instantiation predicate Phi: coherence with E, consistency with B, lexical hygiene>, ...
  ]
\}
\end{Verbatim}
\end{schemabox}
\end{figure*}

For clause cards whose \path{event_type} is \textsc{Reportable} or
\textsc{Non\_Reportable}, the card may optionally be extended with a
\path{missing_information_variants} block (Schema~\ref{schema:missing_variants}).
Each variant specifies one principled way the case can become
factually under-specified while remaining realistic, and supplies the
gold reference for the missing-information detection (M5) and the
slot-identification (M6) metrics. \textsc{Uncertain} cards are not
paired with missing variants and therefore never carry this block.

\begin{figure*}[t]
\begin{schemabox}[label=schema:missing_variants]{Missing-information-variants block (Reportable / Non\_Reportable cards only)}
\begin{Verbatim}[fontsize=\footnotesize, commandchars=\\\{\}]
"missing_information_variants": [
  \{
    "missing_variant_id": <stable identifier of this missing-information variant>,
    "summary":            <one-line description of the masked initial event,
                           used by the Missing-Narrator to know what to write>,
    "masked_boundary_conditions": [
      <boundary condition whose truth value is intentionally not determinable
       from the masked narrative>, ...
    ],
    "masked_basic_event_elements": [
      <basic event element withheld from the masked narrative;
       gold reference for the slot-identification metric M6>, ...
    ],
    "why_reasonable_in_real_world":
      <justification that this missing scenario could plausibly arise in real reporting workflows>,
    "why_not_classifiable_from_remaining_event_description":
      <justification that the residual narrative is genuinely under-determined
       and consistent with at least two verdicts>
  \},
  ...
]
\end{Verbatim}
\end{schemabox}
\end{figure*}

\subsection{LLM-Assisted Authoring Prompts}
\label{app:authoring-prompts}

This subsection collects the 4 LLM prompts that drive the
clause-card authoring pipeline. The first is a \emph{loyal-extraction}
prompt that consolidates, for one target clause, all parts of
MN29 that bear on the clause's reportability semantics into a single
self-contained reference document (Prompt~\ref{prompt:loyal-extraction}).
The next three are \emph{structured-drafting} prompts that consume the
loyal-extraction output and propose, respectively, candidate
\textsc{Complete Cases} clause cards, missing
variants for a given parent card, and \textsc{Uncertain} clause cards
(written up in the next pages). Clause card authoring is conducted by GPT-5.2.

\paragraph{Loyal-extraction prompt.}
The loyal-extraction step is run once per target clause, in a
fresh chat session, with a frontier LLM. The model is asked to copy
verbatim (i) the target clause from \texttt{MN29\_clause.md} and
(ii) every passage from \texttt{MN29\_guidance.md} that materially
affects whether an event is reportable under the target clause meets
the necessary or excluded conditions. Crucially, the model is
\emph{not} allowed to paraphrase, summarize, or interpret the
extracted text --- only to copy. The resulting
\texttt{loyal\_extraction.md} (one per clause) is then used as
the authoritative clause-and-guidance context for every downstream
authoring and verification stage. The full prompt is given in
Prompts~\ref{prompt:loyal-extraction}--\ref{prompt:loyal-extraction-cont}.

\begin{figure*}[t]
\begin{promptbox}[label=prompt:loyal-extraction]{Loyal-extraction prompt (part 1/2): role, inputs, task, relevance criteria, copying rules}
\textbf{Role.}\\
You are a meticulous regulatory-text analyst. Your job is to assemble
a single, self-contained reference document that contains every
piece of MN29 text that materially affects the reportability
semantics of one target clause.\\[4pt]

\textbf{Inputs.}\\
You are given two documents (provided verbatim below):\\
\ \ 1.\ \texttt{TARGET\_CLAUSE}: the identifier and verbatim text of one specific MN29 clause (the ``target clause'').\\
\ \ 2.\ \texttt{MN29\_GUIDANCE}: the full text of the MN29 Guidance document, in its original Markdown structure (top-level sections \emph{General Recommendation N}, \emph{<Family> Event Recommendation N}, and the question/issue subsections under each recommendation).\\[4pt]

\textbf{Task.}\\
Produce one Markdown document, \texttt{loyal\_extraction.md}, that contains:\\
\ \ (i) the verbatim text of the target clause, and\\
\ \ (ii) the verbatim text of every guidance passage that bears on the target clause's reportability semantics.\\[2pt]

A guidance passage ``bears on the target clause's reportability semantics'' if at least one of the following holds:\\
\ - it defines a term that the target clause relies on (e.g.\ ``serious injury'', ``associated with'', ``hospital'');\\
\ - it enumerates inclusion or exclusion criteria that gate the target clause's verdict;\\
\ - it scopes the target clause to a population, setting, time window, or care context;\\
\ - it lists examples of events that are or are not intended to be captured by the target clause;\\
\ - it carves out boundary conditions (e.g.\ ``reasonable differences in clinical judgment'') whose presence or absence flips the verdict;\\
\ - it cross-references the target clause from any \emph{General Recommendation} or \emph{Event Recommendation}.\\[4pt]

\textbf{Strict copying rules.}\\
\ - You may only \emph{copy} text from \texttt{TARGET\_CLAUSE} and \texttt{MN29\_GUIDANCE}. Do not paraphrase, summarize, condense, generalize, interpret, or annotate.\\
\ - Preserve the original wording, punctuation, line breaks, inline emphasis, bullets, and indentation of every excerpt.\\
\ - For each excerpt, copy the entire enclosing block of the source document (e.g.\ the entire \emph{General Recommendation N} section, the entire question/issue subsection); do not cut a recommendation in half.\\
\ - If a recommendation is irrelevant to the target clause, omit it entirely; do not stub it out with placeholders.\\
\ - Do not invent, infer, reorder, or merge content across recommendations.\\
\ - Do not include your own commentary, rationale, or section summaries.\\[2pt]

\emph{The output format, self-checks, and the concrete inputs are continued in Prompt~\ref{prompt:loyal-extraction-cont}.}
\end{promptbox}
\end{figure*}

\begin{figure*}[t]
\begin{promptbox}[label=prompt:loyal-extraction-cont]{Loyal-extraction prompt (part 2/2): output format, self-checks, inputs}
\textbf{Output format.}\\
Produce exactly one Markdown document with the structure below. Use the exact section headings shown, in the order shown:\\
\hrulefill\\
\# \{Family Name\} Clause \{N\}: Step 1 Loyal Extraction\\[2pt]
\#\# Source 1: \texttt{MN29\_clause.md}\\[2pt]
\#\#\# \{Target clause heading, copied verbatim\}\\[2pt]
\{Verbatim text of the target clause.\}\\[6pt]
\#\# Source 2: \texttt{MN29\_guidance.md}\\[2pt]
\#\#\# \{Heading of first relevant guidance section, copied verbatim\}\\[2pt]
\{Verbatim text of that section.\}\\[2pt]
\#\#\# \{Heading of second relevant guidance section, copied verbatim\}\\[2pt]
\{Verbatim text of that section.\}\\[2pt]
... (one heading per relevant guidance section, in the order they appear in \texttt{MN29\_GUIDANCE}) ...\\
\hrulefill\\[4pt]

\textbf{Checks to perform before responding.}\\
\ - Every excerpt under \emph{Source 2} appears verbatim in the provided \texttt{MN29\_GUIDANCE}.\\
\ - Every excerpt under \emph{Source 2} is genuinely relevant to the target clause under at least one of the criteria above.\\
\ - The relative ordering of excerpts under \emph{Source 2} matches their order in \texttt{MN29\_GUIDANCE}.\\
\ - The document contains no content other than the verbatim excerpts and the prescribed section headings.\\[4pt]

\textbf{Inputs follow below.}\\[2pt]
\texttt{TARGET\_CLAUSE} = \{\{target\_clause\_block\}\}\\
\texttt{MN29\_GUIDANCE} = \{\{mn29\_guidance\_full\_text\}\}
\end{promptbox}
\end{figure*}

\paragraph{Structured-drafting prompt for Complete Cases clause cards.}
The first structured-drafting prompt takes one target clause's
loyal-extraction document and the clause-card schema from
\S\ref{app:clause-card-schema} as input, and asks a frontier LLM to
propose a set of candidate clause cards that, taken together, cover
the clause's decision space cleanly. The output is a JSON array of
cards: some carry \path{event_type} = \textsc{Reportable} and seed
reportable cases for the clause, others carry
\path{event_type} = \textsc{Non\_Reportable} and seed non-reportable
cases. The prompt is purposefully restrictive: it requires that the
proposed cards be \emph{mutually exclusive} (no two cards admit the
same factual situation). The prompt does \emph{not} ask for any
\path{missing_information_variants} --- missing variants are
authored by a separate prompt (Prompt~\ref{prompt:missing-drafting})
once each parent card has been finalized. The full prompt is given
in Prompts~\ref{prompt:cc-drafting}--\ref{prompt:cc-drafting-cont}.

\begin{figure*}[t]
\begin{promptbox}[label=prompt:cc-drafting]{Structured-drafting prompt for Complete Cases clause cards (part 1/2): role, inputs, task, coverage requirements}
\textbf{Role.}\\
You are a senior patient-safety regulator drafting the decision specifications that govern MN29 triage for one target clause. You write only specifications, not narratives.\\[4pt]

\textbf{Inputs.}\\
You are given three documents (provided verbatim below):\\
\ \ 1.\ \texttt{TARGET\_CLAUSE\_ID}: a stable identifier of the target MN29 clause (e.g.\ \path{CareManagement_1_MedicationError}).\\
\ \ 2.\ \texttt{LOYAL\_EXTRACTION}: the \texttt{loyal\_extraction.md} produced for this clause; it is the authoritative clause-and-guidance context and supersedes any prior knowledge you may have about MN29.\\
\ \ 3.\ \texttt{CLAUSE\_CARD\_SCHEMA}: the clause-card JSON schema. Every output card must conform to this schema.\\[4pt]

\textbf{Task.}\\
Propose a set of candidate \emph{clause cards} for the target clause. Each card factorizes one decision region of the clause into auditable variables, in the exact schema given by \texttt{CLAUSE\_CARD\_SCHEMA}. Cards may carry one of two event types:\\
\ - \textsc{Reportable}: the card describes a class of events that the target clause makes reportable; the conjunction of its boundary conditions is necessary and jointly sufficient for the reportable verdict under the target clause.\\
\ - \textsc{Non\_Reportable}: the card describes a class of events that fall within the target clause's topical scope but are excluded from reporting by an explicit guidance carve-out (e.g.\ unknown-allergy reaction, reasonable clinical-judgment difference, below-threshold outcome, different cause).\\[2pt]

You may propose one or more cards of each type.\\[4pt]

\textbf{Coverage requirements.}\\
\ - \emph{Mutually exclusive.} Any realistic event compatible with the target clause's topical scope must satisfy the boundary conditions of \emph{at most one} card you propose. Two cards are not mutually exclusive merely because their text differs; their truth-value patterns over the union of all boundary conditions must differ.\\
\ - \emph{Source-grounded.} Each card's \path{governing_legal_basis} must list only identifiers that actually appear in \texttt{LOYAL\_EXTRACTION}, and the truth-value pattern of its boundary conditions must be implied by those listed passages. Do not import recommendations that are not in \texttt{LOYAL\_EXTRACTION}.\\
\ - \emph{Minimal necessary.} Each card's set of boundary conditions should be the smallest set that is necessary and jointly sufficient for the verdict; do not pad with conditions that are always true or trivially follow from other conditions on the same card.\\[4pt]

\textbf{Content requirements per card.}\\
\ - \path{clause_card_id}: prefix is \path{CardRep_<n>} for \textsc{Reportable} cards and \path{CardNonrep_<n>} for \textsc{Non\_Reportable} cards, suffixed by the target clause identifier (e.g.\ \path{CardRep_1_CareManagement_1}).\\
\ - \path{clause_card_definition}: a single-paragraph, self-contained description of the decision region the card covers, written so that a reader who has not seen the boundary conditions can still tell which events fall under this card.\\
\ - \path{boundary_conditions}: each entry is a single-axis predicate (true / false) whose \path{meaning} reproduces the inclusion / exclusion criteria from \texttt{LOYAL\_EXTRACTION} rather than paraphrasing them. The conjunction of the entries' \path{value}s is necessary and jointly sufficient for the card's verdict. Every boundary condition lists at least one \path{corresponding_basic_event_elements}.\\
\ - \path{basic_event_elements}: each element is a neutral factual slot that concretizes at least one boundary condition. Element \path{meaning} fields must be written as fact-slot definitions (``what concrete fact does this slot carry'') --- not as in-boundary declarations (``the patient is harmed''). Different elements must not overlap; in particular, do not let one slot silently encode another slot's content.\\
\ - \path{constraints_on_basic_event_elements_instantiation}: list every predicate over elements that must hold for any case generated from this card to be internally coherent and to realize the intended boundary-condition truth values. Include any banned phrases that would leak the verdict.\\[2pt]

\emph{Schema details, quality checks, and the concrete inputs are continued in Prompt~\ref{prompt:cc-drafting-cont}.}
\end{promptbox}
\end{figure*}

\begin{figure*}[t]
\begin{promptbox}[label=prompt:cc-drafting-cont]{Structured-drafting prompt for Complete Cases clause cards (part 2/2): output format, self-checks, inputs}
\textbf{Output format.}\\
Return exactly one JSON document with the shape\\
\hrulefill\\
\{\\
\ \ "target\_clause\_id": "<value of TARGET\_CLAUSE\_ID>",\\
\ \ "clause\_cards": [\\
\ \ \ \ \{ ... a single clause card conforming to CLAUSE\_CARD\_SCHEMA, with the \path{missing_information_variants} block omitted ... \},\\
\ \ \ \ \{ ... another clause card ... \},\\
\ \ \ \ ...\\
\ \ ]\\
\}\\
\hrulefill\\[2pt]
Do not emit any text outside this JSON document --- no commentary, no Markdown wrapper, no rationale.\\[4pt]

\textbf{Checks to perform before responding.}\\
\ - Every \path{governing_legal_basis.value} identifier appears verbatim in \texttt{LOYAL\_EXTRACTION}.\\
\ - For each card, the truth-value pattern of its boundary conditions is implied by the passages it cites, and no listed passage contradicts the card's verdict.\\
\ - The set of cards is decision-space exhaustive within the target clause's topical scope: every realistic event in scope satisfies the boundary conditions of at least one card.\\
\ - Each card's \path{basic_event_elements} are mutually non-overlapping, neutral fact slots, with no slot silently encoding another slot's content and no slot leaking the verdict.\\
\ - Each card's \path{constraints_on_basic_event_elements_instantiation} list is consistent with its boundary conditions (no constraint forces a slot value that contradicts a boundary-condition truth value).\\
\ - The JSON validates as a single document; identifiers, enum values, and field names exactly match \texttt{CLAUSE\_CARD\_SCHEMA}.\\[4pt]

\textbf{Inputs follow below.}\\[2pt]
\texttt{TARGET\_CLAUSE\_ID} = \{\{target\_clause\_id\}\}\\
\texttt{LOYAL\_EXTRACTION} = \{\{loyal\_extraction\_markdown\}\}\\
\texttt{CLAUSE\_CARD\_SCHEMA} = \{\{clause\_card\_json\_schema\}\}
\end{promptbox}
\end{figure*}

\paragraph{Structured-drafting prompt for missing-information
variants.}
The second structured-drafting prompt is run once per finalized
\textsc{Complete Case} clause card to author
that card's \path{missing_information_variants} block. The prompt
takes the parent card and the same loyal-extraction document used
to draft it, and asks a frontier LLM to enumerate principled ways
the case can become factually under-specified while staying
realistic. Each variant identifies (i) a subset of the parent
card's \path{basic_event_elements} to withhold from the narrative
and (ii) the boundary conditions whose truth value is consequently
no longer determinable from the visible narrative. The full
prompt is given in
Prompts~\ref{prompt:missing-drafting}--\ref{prompt:missing-drafting-cont}.

\begin{figure*}[t]
\begin{promptbox}[label=prompt:missing-drafting]{Structured-drafting prompt for missing-information variants (part 1/2): role, inputs, task, variant requirements}
\textbf{Role.}\\
You are a senior patient-safety regulator. Your job is to enumerate the principled ways in which a real reporting workflow may receive only part of the facts that one specific clause card relies on for its verdict, and to describe each such ``missing'' case in the structured form required by the benchmark.\\[4pt]

\textbf{Inputs.}\\
You are given four documents (provided verbatim below):\\
\ \ 1.\ \texttt{PARENT\_CARD}: one finalized clause card whose \path{event_type} is either \textsc{Reportable} or \textsc{Non\_Reportable}; it is the parent of the variants you are about to author. Treat its \path{boundary_conditions} and \path{basic_event_elements} as the authoritative vocabulary you must reuse verbatim --- do not invent new boundary-condition names or element names.\\
\ \ 2.\ \texttt{SIBLING\_CARDS}: the full list of finalized clause cards on the same clause as \texttt{PARENT\_CARD} (including \texttt{PARENT\_CARD} itself). This is what you must consult to decide whether the residual narrative of a candidate variant is also consistent with an alternative verdict drawn from another card on the clause. You may not invent a sibling card --- only cards in \texttt{SIBLING\_CARDS} count.\\
\ \ 3.\ \texttt{LOYAL\_EXTRACTION}: the \texttt{loyal\_extraction.md} for the parent card's clause. It is the authoritative clause-and-guidance context and supersedes any prior knowledge you may have about MN29.\\
\ \ 4.\ \texttt{MISSING\_VARIANTS\_SCHEMA}: the JSON schema for the \path{missing_information_variants} block. Every variant you propose must conform to this schema.\\[4pt]

\textbf{Task.}\\
Propose a set of missing-information variants for \texttt{PARENT\_CARD}. Each variant describes one realistic reporting scenario in which a specific subset of the parent card's basic event elements is not yet known to the reporter, so that the corresponding boundary conditions are temporarily not determinable from the visible narrative.\\[4pt]

\textbf{Variant requirements.}\\
\ - \emph{Genuine under-determination.} After the masked elements are removed, the remaining visible facts must be consistent with the parent card's verdict \emph{and} with at least one alternative verdict drawn from another card on the same clause. State which alternative card(s) the residual narrative also fits in \path{why_not_classifiable_from_remaining_event_description}. If the remaining facts still uniquely imply the parent verdict, the variant is not under-determined and you must not propose it.\\
\ - \emph{No through-leak.} The visible slots that remain after masking must not, individually or in combination, still imply any masked boundary condition's truth value via plain reading of the narrative. In particular, no visible slot's \path{meaning} or \path{allowed_content} may reproduce, paraphrase, or trivially entail a masked slot.\\
\ - \emph{Boundary--element coherence.} For each variant, \path{masked_boundary_conditions} and \path{masked_basic_event_elements} must be consistent with the parent card's $\rho$ mapping: every masked boundary condition's \path{corresponding_basic_event_elements} must be a subset of \path{masked_basic_event_elements}; conversely, no basic event element is masked unless it concretizes at least one masked boundary condition. Do not mask a boundary condition while leaving any of the elements that concretize it visible.\\
\ - \emph{Realistic missingness.} \path{why_reasonable_in_real_world} must name a concrete reason the masked facts could plausibly be absent from an early or partial report at the named care facility.\\
\ - \emph{Identifier hygiene.} \path{missing_variant_id} is snake\_case, starts with \path{missing_}, names the masked content semantically (e.g.\ \path{missing_outcome_severity_review}), and is unique within the parent card.\\
\ - \emph{Vocabulary fidelity.} Every name appearing in \path{masked_boundary_conditions} or \path{masked_basic_event_elements} must appear verbatim in \texttt{PARENT\_CARD}; never introduce a new boundary-condition or element name here.\\[2pt]

\emph{The output format, self-checks, and the concrete inputs are continued in Prompt~\ref{prompt:missing-drafting-cont}.}
\end{promptbox}
\end{figure*}

\begin{figure*}[t]
\begin{promptbox}[label=prompt:missing-drafting-cont]{Structured-drafting prompt for missing-information variants (part 2/2): output format, self-checks, inputs}
\textbf{Output format.}\\
Return exactly one JSON document with the shape\\
\hrulefill\\
\{\\
\ \ "parent\_clause\_card\_id": "<value of PARENT\_CARD.clause\_card\_id>",\\
\ \ "missing\_information\_variants": [\\
\ \ \ \ \{ ... a single variant conforming to MISSING\_VARIANTS\_SCHEMA ... \},\\
\ \ \ \ \{ ... another variant ... \},\\
\ \ \ \ ...\\
\ \ ]\\
\}\\
\hrulefill\\[2pt]
The \path{missing_information_variants} array is exactly what will be inserted into the parent card. Do not re-emit the parent card. Do not emit any text outside this JSON document --- no commentary, no Markdown wrapper, no rationale.\\[4pt]

\textbf{Checks to perform before responding.}\\
\ - Every \path{masked_boundary_conditions} entry appears verbatim in \texttt{PARENT\_CARD.boundary\_conditions}; every \path{masked_basic_event_elements} entry appears verbatim in \texttt{PARENT\_CARD.basic\_event\_elements}.\\
\ - For each variant, the residual narrative described in \path{summary} is consistent with the parent verdict \emph{and} with at least one other verdict supported by a different card in \texttt{SIBLING\_CARDS}; the alternative is named by its \path{clause_card_id} in \path{why_not_classifiable_from_remaining_event_description}.\\
\ - For each variant, no visible-slot \path{meaning} or \path{allowed_content} reproduces, paraphrases, or trivially entails any masked slot.\\
\ - \path{missing_variant_id} values are unique within the array and follow the \path{missing_} snake\_case convention.\\
\ - The JSON validates as a single document; field names and value types exactly match \texttt{MISSING\_VARIANTS\_SCHEMA}.\\[4pt]

\textbf{Inputs follow below.}\\[2pt]
\texttt{PARENT\_CARD} = \{\{parent\_clause\_card\_json\}\}\\
\texttt{SIBLING\_CARDS} = \{\{all\_clause\_cards\_for\_this\_clause\_json\_array\}\}\\
\texttt{LOYAL\_EXTRACTION} = \{\{loyal\_extraction\_markdown\}\}\\
\texttt{MISSING\_VARIANTS\_SCHEMA} = \{\{missing\_variants\_json\_schema\}\}
\end{promptbox}
\end{figure*}

\paragraph{Structured-drafting prompt for \textsc{Uncertain} clause
cards.}
The third structured-drafting prompt produces \textsc{Uncertain}
clause cards for a target clause. Unlike \textsc{Complete Case} cards, an \textsc{Uncertain} card does
\emph{not} describe a class of events whose verdict is decidable
once all facts are known. Instead, it describes a class of events
whose facts are entirely known yet whose legal classification
remains irreducibly ambiguous, because MN29 itself is silent,
self-contradictory, or carves out a qualitative exclusion without
quantitative bounds. The full prompt is given in
Prompts~\ref{prompt:uncertain-drafting}--\ref{prompt:uncertain-drafting-cont}.

\begin{figure*}[t]
\begin{promptbox}[label=prompt:uncertain-drafting]{Structured-drafting prompt for \textsc{Uncertain} clause cards (part 1/2): role, inputs, task, card requirements}
\textbf{Role.}\\
You are a senior patient-safety regulator drafting the decision specifications that govern \textsc{Uncertain} cases for one target MN29 clause. An \textsc{Uncertain} case is one in which all clinical facts are known and yet the legal classification under the target clause remains genuinely unresolvable because the clause text or its guidance is silent, internally contradictory, or carves out an exclusion without quantitative bounds. You write only specifications, not narratives.\\[4pt]

\textbf{Inputs.}\\
You are given four documents (provided verbatim below):\\
\ 1.\ \texttt{TARGET\_CLAUSE\_ID}: a stable identifier of the target MN29 clause.\\
\ 2.\ \texttt{LOYAL\_EXTRACTION}: the \texttt{loyal\_extraction.md} produced for this clause. It is the authoritative clause-and-guidance context. Every ambiguity you exploit must be traceable to a specific passage of \texttt{LOYAL\_EXTRACTION}.\\
\ 3.\ \texttt{EXISTING\_CARDS}: the full list of finalized \textsc{Reportable} and \textsc{Non\_Reportable} clause cards on this clause. Every \textsc{Uncertain} card you propose must describe a scenario that no card in \texttt{EXISTING\_CARDS} already resolves.\\
\ 4.\ \texttt{CLAUSE\_CARD\_SCHEMA}: the clause-card JSON schema. Every \textsc{Uncertain} card you output must conform to this schema with \path{event_type} set to \textsc{Uncertain}.\\[4pt]

\textbf{Task.}\\
Propose a set of candidate \textsc{Uncertain} clause cards for the target clause. Each card factorizes one ambiguity region of the clause into auditable variables in the schema given by \texttt{CLAUSE\_CARD\_SCHEMA}. Cards differ from each other by which specific policy ambiguity they exploit; do not propose two cards that exploit the same ambiguity.\\[4pt]

\textbf{Card requirements.}\\
\ - \emph{Policy-grounded ambiguity.} Each card's \path{governing_legal_basis.value} must cite exactly the passages in \texttt{LOYAL\_EXTRACTION} whose silence, internal contradiction, or unbounded qualitative carve-out is the source of the ambiguity.\\
\ - \emph{Distinct from existing cards.} For each card, the scenario described by the conjunction of its boundary conditions must \emph{not} be a special case of, equivalent to, or trivially decidable from, any card in \texttt{EXISTING\_CARDS}.\\
\ - \emph{Substantive boundary conditions.} Each boundary condition is a single-axis predicate whose \path{meaning} names a concrete clinical or organizational fact pattern (e.g.\ ``a formal multidisciplinary medication-safety review with two specialties formally taking opposing positions on the same yes/no question''). Do not use meta-conditions such as ``the case is uncertain'' or ``the policy is ambiguous'' as boundary conditions.\\
\ - \emph{Realistic scenarios.} The fact pattern must be one a real reporter at a covered facility could plausibly encounter; it must not be an artificially engineered edge case.\\
\ - \emph{No verdict-leaking vocabulary.} The card's \path{constraints_on_basic_event_elements_instantiation} must forbid, across the eventual generated narrative, any phrase that would either resolve the ambiguity from the surface text or signal it to a downstream model. At minimum, ban the strings ``uncertain'', ``uncertainty'', ``indeterminate'', ``escalate'', ``escalation'', ``inconclusive'', ``reportable'', ``non-reportable'', and any direct citation of MN29 (statute / recommendation / appendix numbers).\\
\ - \emph{No \path{missing_information_variants}.} \textsc{Uncertain} cards are not paired with missing variants; omit this block entirely.\\[2pt]

\emph{The output format, self-checks, and the concrete inputs are continued in Prompt~\ref{prompt:uncertain-drafting-cont}.}
\end{promptbox}
\end{figure*}

\begin{figure*}[t]
\begin{promptbox}[label=prompt:uncertain-drafting-cont]{Structured-drafting prompt for \textsc{Uncertain} clause cards (part 2/2): output format, self-checks, inputs}
\textbf{Output format.}\\
Return exactly one JSON document with the shape\\
\hrulefill\\
\{\\
\ \ "target\_clause\_id": "<value of TARGET\_CLAUSE\_ID>",\\
\ \ "clause\_cards": [\\
\ \ \ \ \{ ... a single clause card conforming to CLAUSE\_CARD\_SCHEMA, with \path{event_type} = "Uncertain" and the \path{missing_information_variants} block omitted ... \},\\
\ \ \ \ \{ ... another clause card ... \},\\
\ \ \ \ ...\\
\ \ ]\\
\}\\
\hrulefill\\[2pt]
Do not emit any text outside this JSON document --- no commentary, no Markdown wrapper, no rationale.\\[4pt]

\textbf{Checks to perform before responding.}\\
\ - Every \path{governing_legal_basis.value} identifier appears verbatim in \texttt{LOYAL\_EXTRACTION}.\\
\ - For each card, the scenario described by the conjunction of its boundary conditions is not a special case of, equivalent to, or trivially decidable from any single card in \texttt{EXISTING\_CARDS}.\\
\ - Different cards in your output exploit different policy ambiguities; no two cards rely on the same silence / contradiction / unbounded carve-out.\\
\ - Each card's \path{constraints_on_basic_event_elements_instantiation} list bans, at minimum, the verdict-leaking vocabulary enumerated above, plus the verbatim text of any unbounded carve-out phrase the card relies on.\\
\ - Every card carries \path{event_type} = \textsc{Uncertain} and omits the \path{missing_information_variants} block.\\
\ - The JSON validates as a single document; identifiers, enum values, and field names exactly match \texttt{CLAUSE\_CARD\_SCHEMA}.\\[4pt]

\textbf{Inputs follow below.}\\[2pt]
\texttt{TARGET\_CLAUSE\_ID} = \{\{target\_clause\_id\}\}\\
\texttt{LOYAL\_EXTRACTION} = \{\{loyal\_extraction\_markdown\}\}\\
\texttt{EXISTING\_CARDS} = \{\{existing\_reportable\_nonreportable\_cards\_json\_array\}\}\\
\texttt{CLAUSE\_CARD\_SCHEMA} = \{\{clause\_card\_json\_schema\}\}
\end{promptbox}
\end{figure*}

\subsection{Generation pipeline prompts}
\label{app:gen-pipeline}

Once a clause card has been finalized through the authoring
process and human auditing
(\S\ref{sec:generation-pipeline}, \S\ref{app:authoring-prompts}), the case generator turns
the card into an end-to-end fact-and-narrative pair through the
two main phases described in \S\ref{sec:generation-pipeline}.
\textsc{Phase 1 -- factual instantiation} couples the
\emph{Instantiator} with the \emph{Instantiation Verifier} to
produce one verified structured fact record per case.
\textsc{Phase 2 -- narrative realization} then couples the
\emph{Event Narrator} with an \emph{Extractor} and a
\emph{Narrative Verifier} to turn that fact record into a
verified free-text narrative. For \textsc{missing-information}
variants, a follow-up pass re-renders and re-verifies the same
fact record on the visible-fact subset using a
\emph{Missing-Variant Narrator} together with the same Extractor
and a dedicated \emph{Missing-Variant Verifier}. Verifier stages
do not retry on their own; when a verifier returns
\verb|"pass": false|, its \verb|"issues"| array is fed back into
the next attempt of the producer stage upstream, up to three
attempts per stage. Cases that exhaust three attempts at any
stage are dropped. The
full prompt for each role is given in
Prompts~\ref{prompt:instantiator}--\ref{prompt:missing-verifier-cont}
that follow. All generation roles are based on GPT-5.2.

\paragraph{Shared background paragraph.}
Every pipeline prompt opens with the same one-paragraph background
that anchors the task in MN29:
``\emph{This pipeline is part of a structured analysis workflow
for Minnesota's 29 Reportable Adverse Health Events framework. In
this workflow, `reportability' means whether a concrete healthcare
event would need to be reported under a specific adverse-event
category according to the relevant clause text and companion
guidance.}'' We refer to this as the \textsc{Mn29 Background}
block in the prompts below.

\paragraph{Retry-feedback block.}
The Instantiator, Event Narrator, and Missing-Variant Narrator
additionally support a \textsc{Retry Feedback} block that is
empty on the first attempt and, on any subsequent attempt,
contains the \verb|"issues"| list returned by the upstream
verifier on the last failed attempt, together with the previous
failed candidate verbatim. This is what closes the loop between a
verifier \verb|"fail"| and the next producer attempt. The exact
wording of each role's retry block is given inside that role's
prompt below.

\paragraph{Instantiator.}
The Instantiator takes the target clause card, the
\texttt{loyal\_extraction.md} for the clause, and a piece of
anchored real-world material, and produces a single concrete
\path{slot_values} object that fills every
\path{basic_event_elements} slot of the card with a medically
plausible value while honoring every boundary condition and every
\path{constraints_on_basic_event_elements_instantiation} entry on
the card. The output is a structured fact record, not yet a
narrative. The full prompt
is given in Prompt~\ref{prompt:instantiator}.

\begin{figure*}[t]
\begin{promptbox}[label=prompt:instantiator]{Instantiator prompt}
\textbf{\# Background.} You are helping with a research pipeline that generates synthetic healthcare event cases. \textsc{<Mn29 Background>}.\\[4pt]

\textbf{\# Task and concepts.}\\
\ - \texttt{Clause Card} is the structured specification of one precise event type that the generated case must satisfy.\\
\ - \texttt{Basic event elements} are the fact slots that must be instantiated.\\
\ - \texttt{Clause and guidance context} contains the governing clause text and companion guidance relevant to this event type. Treat it as authoritative domain context.\\
\ - \texttt{Anchored material} is a realism, scene anchor and style seed. It may be a clinical note, an incident excerpt, a case summary, or other domain-relevant material.\\
\ - Examples inside the clause card may appear in slot descriptions, allowed-content notes, disallowed-content notes, or instantiation constraints. Treat such examples as illustrative reference points that clarify the semantic boundary of the target event type. They are not a preferred sampling pool unless explicitly stated in the field definition that you can only choose from them.\\

Your job is to instantiate one complete structured event for the target clause card. The result must be a concrete factual case, not a narrative and not an analysis.\\[4pt]

\textbf{\# What you will receive.}\\
\ - \texttt{Target clause card}: the event type that must be satisfied, including the card definition, boundary conditions, basic event elements, and instantiation constraints.\\
\ - \texttt{Clause and guidance context}: source-derived context that explains the legal and policy meaning behind the target event type.\\
\ - \texttt{Anchored material}: use the scene anchor only as auxiliary context. You should use it as scene setting to help choose: realistic setting, specialty context, procedure family, terminology or surrounding circumstances.\\
\ - \texttt{Retry feedback} (optional): if this is not the first attempt, you will also receive the most recent failed candidate and the verification issues that it triggered. Use that feedback to correct the next candidate.\\[4pt]

\textbf{\# Behavioral guidance.}\\
1.\ Fill every \path{basic_event_element} exactly once.\\
2.\ Produce a concrete and medically plausible fact value for each slot.\\
3.\ Make the final structured case semantically consistent with the clause-card definition, boundary conditions, and instantiation constraints.\\
4.\ Use the clause card as the semantic contract. If the anchored material conflicts with the clause card, follow the clause card.\\
5.\ If the clause card includes example facts or example phrasings, treat them as illustrative only. Do not simply copy them, lightly paraphrase them, or default to them when other semantically valid concrete instantiations are available under the anchored setting.\\
6.\ Use the anchored material and the surrounding clause/guidance context to diversify the generated case as long as the final case still fits the target clause card.\\
7.\ Use concrete facts rather than abstract labels. For example, prefer a specific side, body part, site, level, medication, device, injury, or action instead of vague phrases such as ``wrong site'' or ``correct side.''\\
8.\ Do not leave required facts vague just to sound general.\\
9.\ If \texttt{Retry feedback} is provided, treat it as required corrective guidance for the next attempt.\\
10.\ Do not write an event narrative, explanation, or commentary.\\
11.\ Do not add any top-level key other than \path{slot_values}.\\
12.\ Do not write any reportability statements in the instantiated fact value, such as \emph{reportable}, \emph{non-reportable}, \emph{should be reported}, \emph{should not be reported}, \emph{below/above reporting threshold} or uses equivalent direct classification language.\\[4pt]

\textbf{\# Input information.}\\
\#\# Target clause card: \texttt{\{\{target\_clause\_card\_json\}\}}\\
\#\# Clause and guidance context: \texttt{\{\{loyal\_extraction\_markdown\}\}}\\
\#\# Anchored material: \texttt{\{\{anchored\_material\_text\}\}}\\
\textsc{Retry Feedback} block (empty on first attempt; on retries, includes the previous failed \path{slot_values} JSON and the upstream verifier's \path{issues} array).\\[4pt]

\textbf{\# Output format.}\\
Return JSON only in exactly this shape:\\
\verb|{ "slot_values": { "<slot_name_1>": "...",|\\
\verb|                   "<slot_name_2>": "...", ... } }|\\
Include every required slot name exactly once; use \texttt{null} only if the slot's \path{field_type} allows null; do not add extra keys.
\end{promptbox}
\end{figure*}

\paragraph{Instantiation Verifier.}
The Instantiation Verifier checks the candidate
\path{slot_values} object against the target clause card and the
clause-and-guidance context. It does not edit the candidate; it
returns a strict \texttt{pass}\,/\,\texttt{fail} verdict and, on
\texttt{fail}, a concrete \verb|"issues"| list that the
Instantiator consumes on its next attempt. The full prompt is
given in Prompt~\ref{prompt:inst-verifier}.

\begin{figure*}[t]
\begin{promptbox}[label=prompt:inst-verifier]{Instantiation Verifier prompt}
\textbf{\# Background.} You are verifying one candidate structured healthcare event. \textsc{<Mn29 Background>}.\\[4pt]

\textbf{\# Task and concepts.}\\
\ - \texttt{Clause Card} defines one precise event type.\\
\ - \texttt{Basic event elements} are neutral fact slots. Their \path{meaning} explains what each slot represents.\\
\ - \texttt{Candidate slot\_values} is the proposed structured instantiation that must be checked.\\
\ - \texttt{Clause and guidance context} provides the governing domain context that explains how the event type should be understood.\\

Your job is to decide whether the candidate instantiation truly fits the target clause card. This is a strict verification task, not a rewriting task.\\[4pt]

\textbf{\# What you will receive.}\\
\ - \texttt{Target clause card}: the event type that the candidate must satisfy.\\
\ - \texttt{Clause and guidance context}: authoritative supporting context for interpreting the clause card.\\
\ - \texttt{Candidate slot\_values}: the structured event to verify.\\[4pt]

\textbf{\# Behavioral guidance.}\\
1.\ Check whether the candidate fits the clause-card definition as a whole.\\
2.\ Check whether the candidate is consistent with each boundary condition.\\
3.\ Check whether the candidate satisfies the instantiation constraints.\\
4.\ Check whether the slot values are concrete rather than vague pseudo-concrete labels.\\
5.\ Be strict about semantic fit, but judge semantic equivalence rather than exact wording.\\[4pt]

\textbf{\# Input information.}\\
\#\# Target clause card: \texttt{\{\{target\_clause\_card\_json\}\}}\\
\#\# Clause and guidance context: \texttt{\{\{loyal\_extraction\_markdown\}\}}\\
\#\# Candidate \path{slot_values}: \texttt{\{\{candidate\_slot\_values\_json\}\}}\\[4pt]

\textbf{\# Output format.}\\
Return JSON only in this shape: \verb!{ "pass": true | false, "issues": [] }!. Return \verb|pass: true| only if the candidate fully satisfies the target clause card; otherwise return \verb|pass: false| and list the concrete problems in \verb|issues|. Do not add keys other than \verb|pass| and \verb|issues|.
\end{promptbox}
\end{figure*}

\paragraph{Event Narrator.}
The Event Narrator turns the verified \path{slot_values} record
into a free-text incident description that reads like a real
clinical write-up. It is given a neutral \path{meaning}-only view
of the basic-event-element schema (no card-level metadata that
would betray the verdict), the canonical \path{slot_values}, the
clause text and companion guidance for this clause, and the same
anchor material that grounded the Instantiator. It must preserve
every load-bearing fact while forbidding any verdict language
(\emph{reportable}, \emph{non-reportable}, clause numbers, etc.)
in the surface text. The full prompt is given in
Prompt~\ref{prompt:event-narrator}.

\begin{figure*}[t]
\begin{promptbox}[label=prompt:event-narrator]{Event Narrator prompt}
\textbf{\# Background.} You are a narrator for a synthetic healthcare-event generation pipeline. \textsc{<Mn29 Background>}.\\[4pt]

\textbf{\# Task and concepts.}\\
\ - \texttt{Basic event elements} are neutral fact slots. Their \path{meaning} explains what each fact represents.\\
\ - \texttt{Canonical structured event} is the factual ground truth that you must convey. It is the core source of event facts. Do not omit or contradict any of its facts. You may paraphrase or reframe the facts to make the narrative more realistic and authentic, as long as the original meaning remains unchanged and semantically consistent.\\
\ - \texttt{Clause and guidance context} is provided only to help you stay aligned with the event type and the surrounding domain context.\\
\ - \texttt{Anchored material} is real-world text that serves as a realism and style reference. It may be a clinical note, an incident excerpt, a case summary, or other domain-relevant material. Draw on it to make your narrative concrete, specific, and natural-sounding: borrow its level of detail, clinical vocabulary, sentence rhythm, and writing register.\\

Your role is to act as a narrator or recorder. Based on the structured facts, write one coherent, natural, realistic healthcare event description, as if it were a real event write-up from a clinical or safety-reporting context. The output should feel indistinguishable from a real event write-up authored by a clinician, nurse, or safety reporter.\\[4pt]

\textbf{\# What you will receive.}\\
\ - \texttt{Basic event element meanings}: the neutral meaning of each fact slot.\\
\ - \texttt{Canonical structured event}: the core facts that must be conveyed.\\
\ - \texttt{Clause and guidance context}: domain context that helps you understand the event type.\\
\ - \texttt{Anchored material}: real-world text to draw on for writing style, clinical vocabulary, and level of detail. If the anchored material itself is an event description, use it to emulate realistic writing style, tone, clinical vocabulary, and level of detail.\\
\ - \texttt{Retry feedback} (optional): if this is not the first attempt, you will also receive the most recent failed narrative and the verification issues that it triggered.\\[4pt]

\textbf{\# Behavioral guidance.}\\
1.\ Preserve the factual meaning of the canonical structured event. Do not omit or substantially alter load-bearing facts.\\
2.\ Draw on the anchored material for writing style, clinical vocabulary, and level of concrete detail. Your narrative should be as specific and natural as the kind of text found in the anchor.\\
3.\ If \texttt{Retry feedback} is provided, treat it as required corrective guidance for the next attempt.\\
4.\ Do not mention reportability, non-reportability, clause numbers, clause-card ids, legal conclusions, or recommendations about whether the event should be reported.\\
5.\ Write in a natural, specific, and realistic clinical style, not like a rigid template, checklist, or legal summary.\\
6.\ Beyond the core facts in the canonical structured event, you may draw on, reuse or modify other related background information from the anchored material so that the resulting event description feels more realistic and authentic.\\
7.\ Adopt the perspective of (or portray yourself as) a nurse, physician, or another healthcare professional involved in or knowledgeable about the event (e.g.\ pharmacist, therapist, technician, radiology/lab staff, rehabilitation physician). The narrative may be written in either first-person or third-person, as long as it reflects a realistic clinical voice. Focus on producing a natural, coherent description of what occurred, similar to documentation or a clinical note, rather than a rigid or overly formal incident analysis report.\\[4pt]

\textbf{\# Input information.}\\
\#\# Basic event element meanings: \texttt{\{\{bee\_skeleton\_json\}\}}\\
\#\# Canonical \path{slot_values}: \texttt{\{\{canonical\_slot\_values\_json\}\}}\\
\#\# Clause and guidance context: \texttt{\{\{loyal\_extraction\_markdown\}\}}\\
\#\# Anchored material: \texttt{\{\{anchored\_material\_text\}\}}\\
\textsc{Retry Feedback} block (empty on first attempt; on retries, includes the previous failed narrative and the upstream verifier's \path{issues} array).\\[4pt]

\textbf{\# Output format.}\\
Return JSON only in this shape: \verb|{ "event_narrative": "..." }|. Return exactly one event description; the narrative must be specific and realistic; do not add keys other than \verb|event_narrative|.
\end{promptbox}
\end{figure*}

\paragraph{Extractor.}
The Extractor reads the generated narrative and re-emits one
value per basic-event-element slot, using only the narrative as
its source and the same neutral \path{meaning}-only schema view.
It is intentionally given no clause-card metadata, no boundary
conditions, no canonical \path{slot_values}, and no verdict
hints, so its output is a faithful structured projection of what
the narrative actually says. The same Extractor prompt is reused
in both phases: its output feeds the Narrative Verifier in the
complete path and the Missing-Variant Verifier in the missing
path. The full prompt is given in Prompt~\ref{prompt:extractor}.

\begin{figure*}[t]
\begin{promptbox}[label=prompt:extractor]{Extractor prompt}
\textbf{\# Background.} You are extracting structured facts from a healthcare event description. \textsc{<Mn29 Background>}.\\[4pt]

\textbf{\# Task and concepts.}\\
\ - \texttt{Basic event element skeleton} is the extraction schema.\\
\ - \path{meaning} is the primary, authoritative description of what that slot represents.\\
\ - Your role is objective information extraction only.\\

Your job is to read the event narrative and extract one value for each slot in the provided schema based on the content of the event narrative (the narrative itself is the only source of extracted values).\\[4pt]

\textbf{\# Behavioral guidance.}\\
1.\ Use each field's \path{meaning} as the primary extraction target.\\
2.\ If a slot is not stated, not recoverable, or genuinely ambiguous, return \verb|null| for that slot.\\
3.\ Do not invent facts to make the narrative fit any slot.\\[4pt]

\textbf{\# Input information.}\\
\#\# Basic event element skeleton: \texttt{\{\{bee\_skeleton\_json\}\}}\\
\#\# Event narrative: \texttt{\{\{event\_narrative\_text\}\}}\\[4pt]

\textbf{\# Output format.}\\
Return JSON only in this shape: \verb|{ "slot_values": { "<slot_name_1>": null, ... } }|. Include every slot exactly once inside \verb|slot_values|; use \verb|null| when the narrative does not support a value; do not add keys other than \verb|slot_values|.
\end{promptbox}
\end{figure*}

\paragraph{Narrative Verifier.}
The Narrative Verifier compares the canonical \path{slot_values}
against the Extractor's reconstruction from the narrative and
decides whether the narrative preserved every load-bearing fact.
It also enforces the no-verdict-leak rule: even if every slot is
preserved, the narrative fails verification if it explicitly
says the event \emph{is reportable}, \emph{non-reportable},
\emph{should be reported}, or uses any equivalent direct
classification language. The full prompt is given in
Prompt~\ref{prompt:narrative-verifier}.

\begin{figure*}[t]
\begin{promptbox}[label=prompt:narrative-verifier]{Narrative Verifier prompt}
\textbf{\# Background.} You are verifying whether a narrated healthcare event preserved a canonical structured event. \textsc{<Mn29 Background>}.\\[4pt]

\textbf{\# Task and concepts.}\\
\ - \texttt{Canonical slot\_values} is the factual ground truth that the narrator was supposed to express.\\
\ - \texttt{Event narrative} is the actual generated healthcare event description that you must inspect directly.\\
\ - \texttt{Extracted slot\_values} is the structured interpretation recovered from the resulting narrative.\\
\ - \texttt{Basic event element meanings} explain the neutral meaning of each slot.\\

Your job is to compare the canonical structured event with the extracted structured event and decide whether the narrative preserved the intended facts.\\[4pt]

\textbf{\# What you will receive.}\\
\ - \texttt{Basic event element meanings}: neutral interpretations of the slots being compared.\\
\ - \texttt{Canonical slot\_values}: the source facts.\\
\ - \texttt{Event narrative}: the actual generated narrative whose wording you must inspect directly.\\
\ - \texttt{Extracted slot\_values}: the facts recovered from the generated narrative.\\[4pt]

\textbf{\# Behavioral guidance.}\\
1.\ Judge semantic equivalence, not exact surface wording.\\
2.\ Pass only if the extracted \path{slot_values} preserved the canonical facts in substance.\\
3.\ Fail if a load-bearing fact was omitted, materially altered, contradicted, or replaced by a materially different fact.\\
4.\ Fail if the narrative explicitly states that the event is \emph{reportable}, \emph{non-reportable}, \emph{should be reported}, \emph{should not be reported}, or uses equivalent direct classification language.\\
5.\ Be specific in \verb|issues| when you identify a problem. If you find multiple issues, list them all.\\[4pt]

\textbf{\# Input information.}\\
\#\# Basic event element meanings: \texttt{\{\{bee\_skeleton\_json\}\}}\\
\#\# Canonical \path{slot_values}: \texttt{\{\{canonical\_slot\_values\_json\}\}}\\
\#\# Event narrative: \texttt{\{\{event\_narrative\_text\}\}}\\
\#\# Extracted \path{slot_values} from the narrative: \texttt{\{\{extracted\_slot\_values\_json\}\}}\\[4pt]

\textbf{\# Output format.}\\
Return JSON only in this shape: \verb!{ "pass": true | false, "issues": [] }!. Return \verb|pass: true| only if the extracted facts preserve the canonical facts and the narrative does not contain verdict-leaking language; otherwise return \verb|pass: false| and explain the concrete mismatches in \verb|issues|.
\end{promptbox}
\end{figure*}

\paragraph{Missing-Variant Narrator.}
The Missing-Variant Narrator produces a \textsc{missing} variant
of an already-accepted complete case by re-rendering its visible
facts under the masking pattern specified by the clause card. It
sees only the visible-fact subset, never the withheld facts, and
must convey those visible facts faithfully while leaving the
withheld facts naturally unstated. The full prompt spans
Prompts~\ref{prompt:missing-narrator}--\ref{prompt:missing-narrator-cont}.

\begin{figure*}[t]
\begin{promptbox}[label=prompt:missing-narrator]{Missing-Variant Narrator prompt (part 1/2): role, inputs, behavioral guidance}
\textbf{\# Background.} You are a narrator for a synthetic healthcare-event generation pipeline. \textsc{<Mn29 Background>}.\\[4pt]

\textbf{\# Task and concepts.}\\
\ - \texttt{Basic event elements} are neutral fact slots. Their \path{meaning} explains what each fact represents.\\
\ - \texttt{Structured event facts} are the factual ground truth that you must convey. It is the core source of visible facts. Do not omit or contradict any of its facts. You may paraphrase or reframe the facts to make the narrative more realistic and authentic, as long as the original meaning remains unchanged and semantically consistent.\\
\ - \texttt{Missing variant summary} describes the intended incompleteness pattern for this narrative. It is a hidden writing constraint, not text that should be repeated or explicitly explained inside the narrative.\\
\ - \texttt{Masked fact slots} are fact dimensions that were intentionally withheld. They are provided to you so you know what kinds of information must stay entirely out of the narrative.\\
\ - \texttt{Clause and guidance context} is provided only to help you stay aligned with the event type and surrounding domain context.\\
\ - \texttt{Anchored material} is real-world text that serves as a realism and style reference, used the same way as in the Event Narrator prompt.\\

Your role is to act as a narrator or recorder. Based only on the structured facts you are given, write one coherent, natural, realistic healthcare event description, as if it were a real event write-up. The output should feel indistinguishable from a real event write-up authored by a clinician, nurse, or safety reporter.\\[4pt]

\textbf{\# What you will receive.}\\
\ - \texttt{Basic event element meanings}: the neutral meaning of each visible fact slot that you are allowed to use.\\
\ - \texttt{Structured event facts}: the visible facts that must be conveyed.\\
\ - \texttt{Missing variant summary}: a description of the intended incompleteness pattern.\\
\ - \texttt{Masked fact slots and meanings}: the slot names and neutral meanings for the facts that must remain absent from the narrative.\\
\ - \texttt{Masked boundary conditions}: each masked boundary condition's \path{meaning} and \path{masked_condition_value}. The \path{masked_condition_value} is given to you only as metadata about the underlying case --- it is NOT something you should reveal through the narrative; your job is to keep its truth value unresolvable from the visible content.\\
\ - \texttt{Clause and guidance context}, \texttt{Anchored material}, \texttt{Retry feedback} (same semantics as the Event Narrator prompt).\\[2pt]

\emph{The behavioral-guidance rules, input placeholders, and output format are continued in Prompt~\ref{prompt:missing-narrator-cont}.}
\end{promptbox}
\end{figure*}

\begin{figure*}[t]
\begin{promptbox}[label=prompt:missing-narrator-cont]{Missing-Variant Narrator prompt (part 2/2): behavioral guidance, inputs, output format}
\textbf{\# Behavioral guidance.}\\
1.\ Preserve the factual meaning of the structured facts you are given. Do not omit or substantially alter load-bearing facts.\\
2.\ Draw on the anchored material for writing style, clinical vocabulary, and level of concrete detail.\\
3.\ If \texttt{Retry feedback} is provided, treat it as required corrective guidance for the next attempt.\\
4.\ Use the \texttt{Missing variant summary} only as an internal writing constraint that helps you decide what should remain naturally unstated.\\
5.\ Treat the \texttt{Masked fact slots and meanings} as a strict do-not-mention list: do not include facts, explanations, caveats, or uncertainty statements that would reveal or directly discuss those masked dimensions.\\
6.\ \emph{Boundary-condition preservation.} Treat each entry in \texttt{Masked boundary conditions} as a creation-time constraint, not only as something the verifier will check after the fact. For each masked boundary condition, before committing any sentence, mentally ask: ``would a reasonable, clinically literate reader of this sentence be able to confidently determine the truth value of this boundary condition?'' If yes, rewrite the sentence so the answer becomes ``no'' or ``only with substantial speculation.''\\
7.\ Do not explicitly say that some information is missing, unknown, unavailable, pending, not yet determined, not documented, or still under review if that statement would point to facts that were intentionally omitted.\\
8.\ Do not directly discuss the absence of omitted facts. The correct behavior is to simply not mention them.\\
9.\ Do not mention reportability, non-reportability, clause numbers, clause-card ids, legal conclusions, or recommendations about whether the event should be reported.\\
10.\ Write in a natural, specific, and realistic clinical style, not like a rigid template, checklist, or legal summary.\\
11.\ Beyond the core facts in the visible structured event facts, you may draw on, reuse or modify other related background information from the anchored material so that the resulting event description feels more realistic and authentic --- but never at the cost of resolving a masked boundary condition.\\
12.\ Adopt the perspective of a nurse, physician, or another healthcare professional involved in or knowledgeable about the event, in either first-person or third-person.\\[4pt]

\textbf{\# Input information.}\\
\#\# Basic event element meanings (visible slots only): \texttt{\{\{visible\_bee\_skeleton\_json\}\}}\\
\#\# Structured event facts (visible only): \texttt{\{\{visible\_slot\_values\_json\}\}}\\
\#\# Missing variant summary: \texttt{\{\{missing\_variant\_summary\_text\}\}}\\
\#\# Masked fact slots and meanings: \texttt{\{\{masked\_bee\_meanings\_json\}\}}\\
\#\# Masked boundary conditions (each must remain truth-value indeterminate): \texttt{\{\{masked\_bc\_view\_json\}\}}\\
\#\# Clause and guidance context: \texttt{\{\{loyal\_extraction\_markdown\}\}}\\
\#\# Anchored material: \texttt{\{\{anchored\_material\_text\}\}}\\
\textsc{Retry Feedback} block (empty on first attempt; on retries, includes the previous failed missing narrative and the upstream verifier's \path{issues} array).\\[4pt]

\textbf{\# Output format.}\\
Return JSON only in this shape: \verb|{ "event_narrative": "..." }|. Return exactly one event description; the narrative must be specific and realistic; do not add keys other than \verb|event_narrative|.
\end{promptbox}
\end{figure*}

\paragraph{Missing-Variant Verifier.}
The Missing-Variant Verifier checks the missing-variant narrative
on three axes: (i) every visible fact is preserved (the
Extractor's output for visible slots matches the canonical
visible \path{slot_values} in substance), (ii) every masked
basic-event-element slot remains unrecoverable (the Extractor
returns \texttt{null} for it, \emph{and} the narrative does not
explicitly acknowledge the omission), and (iii) every masked
boundary condition's truth value remains unresolvable from the
visible content, including indirect channels. The
full prompt spans
Prompts~\ref{prompt:missing-verifier}--\ref{prompt:missing-verifier-cont}.

\begin{figure*}[t]
\begin{promptbox}[label=prompt:missing-verifier]{Missing-Variant Verifier prompt (part 1/2): role, inputs, behavioral guidance}
\textbf{\# Background.} You are verifying an intentionally incomplete healthcare event description. \textsc{<Mn29 Background>}.\\[4pt]

\textbf{\# Task and concepts.}\\
\ - \texttt{Full canonical slot\_values} is the complete structured event before masking.\\
\ - \texttt{Masked slot names} are the basic-event-element slots whose values were intentionally withheld from the narrative.\\
\ - \texttt{Masked boundary conditions} are clause-card-level truth-valued conditions that the masking pattern was supposed to make indeterminate. Each masked boundary condition has a \path{meaning} (a natural-language statement of what truth-value question it asks) and a \path{masked_condition_value} (the truth value this card's full instantiation happens to take for that condition). The \path{masked_condition_value} is given to you only as metadata about the underlying case --- it is NOT the answer the narrative should let you reach. A correctly written incomplete narrative must leave each masked boundary condition's truth value unresolvable from the visible content --- meaning a reasonable, clinically literate reader cannot confidently determine ANY truth value (whether or not it matches \path{masked_condition_value}) from the visible narrative, neither through the masked slots nor through any other surrounding wording.\\
\ - \texttt{Missing event narrative} is the actual incomplete narrative that was generated.\\
\ - \texttt{Extracted slot\_values} is the structured interpretation recovered from the incomplete narrative.\\
\ - \texttt{Basic event element meanings} explain the neutral meaning of each slot.\\

Your job is to decide whether the incomplete narrative behaved correctly: it should preserve the visible facts, keep the masked basic-event-element slots unrecoverable, and --- crucially --- keep each masked boundary condition's truth value unresolvable from the visible content, directly or indirectly.\\[4pt]

\textbf{\# What you will receive.}\\
\ - \texttt{Basic event element meanings}: neutral interpretations of the slots being checked.\\
\ - \texttt{Full canonical slot\_values before masking}: the complete source facts.\\
\ - \texttt{Masked slot names}: the basic-event-element slots whose values must be absent from the narrative.\\
\ - \texttt{Masked boundary conditions}: each masked boundary condition's \path{meaning} and \path{masked_condition_value}. Reason about each of these conditions individually. Do NOT use \path{masked_condition_value} as a target to confirm --- your task is to detect whether ANY confident truth value can be inferred from the narrative.\\
\ - \texttt{Missing event narrative}: the actual generated narrative whose wording you must inspect directly.\\
\ - \texttt{Extracted slot\_values from the incomplete narrative}: the facts recovered from the generated narrative.\\[2pt]

\emph{The behavioral-guidance rules, input placeholders, and output format are continued in Prompt~\ref{prompt:missing-verifier-cont}.}
\end{promptbox}
\end{figure*}

\begin{figure*}[t]
\begin{promptbox}[label=prompt:missing-verifier-cont]{Missing-Variant Verifier prompt (part 2/2): behavioral guidance, inputs, output format}
\textbf{\# Behavioral guidance.}\\
1.\ Judge semantic equivalence, not exact surface wording.\\
2.\ Pass only if every visible fact that should remain available is preserved in the extracted result.\\
3.\ Pass only if every masked basic-event-element slot remains absent or unrecoverable in the extracted result, meaning the extracted value for that slot is \verb|null|.\\
4.\ Fail if the narrative explicitly says that a masked fact is missing, unknown, unavailable, unclear, pending, not yet determined, not documented, or still under review.\\
5.\ Fail if the narrative directly discusses the absence of a masked fact, even if the extractor still returns \verb|null| for that slot.\\
6.\ Fail if the narrative explicitly states that the event is \emph{reportable}, \emph{non-reportable}, \emph{should be reported}, \emph{should not be reported}, or uses equivalent direct classification language.\\
7.\ Fail if any visible fact was lost or materially changed.\\
8.\ Fail if any masked basic-event-element fact leaked into the narrative and became recoverable.\\
9.\ \emph{Boundary-condition leakage check.} For EACH entry in \texttt{Masked boundary conditions} you MUST emit exactly one corresponding row in the required output field \path{per_bc_check}, in the same order and using the same \path{boundary_condition} name. Do NOT collapse, merge, skip, or summarize any masked boundary condition. For each row, independently reason about the following question: ``Given only the visible content of the incomplete narrative, can a reasonable, clinically literate reader confidently determine the truth value of this boundary condition?'' Answer this question by considering the narrative as a whole, including content that does NOT live inside the masked basic-event-element slots --- adjacent clinical descriptions, setting and scope language, severity or outcome wording, capacity/status descriptors, named devices/products/specimens/drugs/procedures, downstream review or imaging findings, post-hoc reconciliation language, timeline of catch or intervention, clinical-team causal attributions, and so on. If the answer is ``yes, confidently'' --- meaning the narrative directly states or strongly implies a single specific truth value for that boundary condition --- set \path{leak_verdict} to \verb|"leaked"|, populate \path{evidence_phrase} with a short quote (or close paraphrase) of the betraying narrative wording, and write a concise \path{explanation} describing how that wording resolves the boundary condition. If the answer is ``no'' or ``only with substantial speculation'' --- meaning the visible content is genuinely consistent with both truth values of that boundary condition --- set \path{leak_verdict} to \verb|"indeterminate"|, set \path{evidence_phrase} to \verb|null|, and use \path{explanation} to briefly note what range of interpretations the visible content remains consistent with. A boundary-condition leak counts even when every masked basic-event-element slot was correctly extracted as \verb|null|.\\
10.\ Be specific in \verb|issues|. For every \path{per_bc_check} row whose \path{leak_verdict} is \verb|"leaked"|, the \verb|issues| array MUST also contain a matching entry that names the boundary condition, quotes or paraphrases the betraying narrative wording, explains how that wording resolves the boundary condition, and instructs the narrator to remove or rewrite it on the next revision.\\[4pt]

\textbf{\# Input information.}\\
\#\# Basic event element meanings: \texttt{\{\{bee\_skeleton\_json\}\}}\\
\#\# Full canonical \path{slot_values} before masking: \texttt{\{\{full\_canonical\_slot\_values\_json\}\}}\\
\#\# Masked slot names: \texttt{\{\{masked\_slot\_names\_json\}\}}\\
\#\# Masked boundary conditions (each must remain truth-value indeterminate): \texttt{\{\{masked\_bc\_view\_json\}\}}\\
\#\# Missing event narrative: \texttt{\{\{missing\_event\_narrative\_text\}\}}\\
\#\# Extracted \path{slot_values} from the incomplete narrative: \texttt{\{\{extracted\_slot\_values\_json\}\}}\\[4pt]

\textbf{\# Output format.}\\
Return JSON only in this shape:\\
\verb!{ "per_bc_check": [ { "boundary_condition": "<exact name>", "leak_verdict":!\\
\verb!     "leaked" | "indeterminate", "evidence_phrase": "<short quote or null>",!\\
\verb!     "explanation": "<<=60 words>" } ], "pass": true | false, "issues": [] }!\\
\path{per_bc_check} is REQUIRED and must contain exactly one row per masked boundary condition (empty list if there are none); \verb|pass| MUST be \verb|false| whenever any row has \path{leak_verdict} equal to \verb|"leaked"|; do not add keys other than \path{per_bc_check}, \verb|pass|, and \verb|issues|.
\end{promptbox}
\end{figure*}

\subsection{MN29 policy and JQ anchor materials}
\label{app:mn29-jq}

\paragraph{MN29 reporting policy.}
\emph{Minnesota's 29 Reportable Adverse Health Events} (MN29) is a
state-mandated, statute-backed reporting framework codified in
Minn. Stat. \S\,144.7065 \citep{mdh_mn29_adverse_events} and
operationalized by the Minnesota Department of Health (MDH).
Hospitals and licensed ambulatory surgical centers in Minnesota
must report any event falling under the 29 enumerated adverse
event types to the commissioner within $15$ working days of
discovery and must subsequently file a root-cause analysis and a
corrective action plan (Minn. Stat. \S\,144.7067). The 29 events are grouped into 7
categories --- \emph{Surgical}, \emph{Product or Device},
\emph{Patient Protection}, \emph{Care Management},
\emph{Environmental}, \emph{Potential Criminal}, and
\emph{Radiologic} --- which collectively constitute the global
clause numbering ($1$--$29$) we use throughout the paper
(Table~\ref{tab:ds-clause-counts}). Each clause is one or two
sentences of statutory text, often with explicit carve-outs
(e.g.\ Surgical clause~1 excludes ``situations requiring prompt
action that occur in the course of surgery''; Patient Protection
clause~2 excludes ``adults who have decision-making capacity'').
The statute is paired with an extensive companion guidance
document maintained by MDH that defines load-bearing terms
(``serious injury,'' ``associated with,'' ``setting licensed
under the reporting facility,'' ``patient'' vs.\ no-longer-a-patient,
``reasonable differences in clinical judgment'') and lists
worked Q\&A examples. The MN29 clauses together with this
guidance form the policy substrate we extract into clause cards
(Schemas~\ref{schema:clause_card}--\ref{schema:missing_variants} and
Prompt~\ref{prompt:loyal-extraction}).

\paragraph{JQ anchor materials.}
For anchor-driven instantiation (\S\ref{sec:generation-pipeline}),
we draw on the public reporting outputs of the Japan Council for
Quality Healthcare (JQ), a neutral third-party organization that
has operated the \emph{Project to Collect Medical Near-Miss /
Adverse Event Information} since $2004$ \citep{jq_medsafe_database}.
JQ publishes periodic and annual reports as well as monthly
medical safety information notices. Each report aggregates
de-identified case excerpts contributed by Japanese medical
institutions (both mandatory reporters and voluntarily
participating institutions) covering a broad spectrum of
near-miss and adverse-event scenarios across surgery, device use,
medication administration, patient protection, environmental
hazards, and diagnostic workflows. We use the $2023$ and $2024$
\emph{medical accident information} releases. Each released case
provides a short structured incident description plus narrative
context (clinical setting, procedure, timing, and surrounding
circumstances), all written in Japanese; we translate each case
into English with an LLM (gpt-5.4-mini), coarsely categorize it by clinical
specialty, and then pair it with one MN29 clause card during
generation.

\subsection{Dataset statistics}
\label{app:dataset-stats}

This subsection breaks down the static clause-card library and
the $5{,}074$ generated benchmark cases.

\paragraph{Clause-card library.}
Tables~\ref{tab:ds-cards-by-cat}--\ref{tab:ds-card-shape}
summarize the static set of clause cards. The library contains
$146$ \textsc{Complete Cases} cards
($45$ \textsc{Reportable}, $101$ \textsc{Non\_Reportable}),
$66$ \textsc{missing-information} variants attached to a subset
of those parents cards, and $11$ \textsc{Uncertain} cards.
Table~\ref{tab:ds-cards-by-cat} shows the per-category split. 
Table~\ref{tab:ds-clause-counts} reports per-clause card and
variant counts at the global clause numbering. The $146$ complete cases
cards are distributed across all $29$ clauses; the $66$ missing
variants are concentrated on the $61$ parents whose clause text
admits a clinically realistic ``initial-report-is-incomplete''
failure mode.
Table~\ref{tab:ds-card-shape} reports the distribution of
\path{boundary_conditions} and \path{basic_event_elements} counts
per card.

\begin{table}[t]
\centering
\footnotesize
\setlength{\tabcolsep}{4pt}
\begin{tabular}{lrrrr}
\toprule
MN29 category & Rep. & Non-Rep. & Unc. & Total \\
\midrule
Surgical            & 16 & 24 & 2 & 42 \\
Product / Device    &  3 & 11 & 2 & 16 \\
Patient Protection  &  5 & 11 & 0 & 16 \\
Care Management     & 12 & 33 & 4 & 49 \\
Environmental       &  4 & 10 & 1 & 15 \\
Potential Criminal  &  4 &  9 & 1 & 14 \\
Radiologic          &  1 &  3 & 1 &  5 \\
\midrule
\textbf{Total}      & \textbf{45} & \textbf{101} & \textbf{11} & \textbf{157} \\
\bottomrule
\end{tabular}
\caption{Clause-card library by MN29 category and event type. ``Rep.'' =
\textsc{Reportable}, ``Non-Rep.'' = \textsc{Non\_Reportable},
``Unc.'' = \textsc{Uncertain}.}
\label{tab:ds-cards-by-cat}
\end{table}

\begin{table}[t]
\centering
\footnotesize
\setlength{\tabcolsep}{4pt}
\begin{tabular}{c l rr}
\toprule
\# & Category (local clause) & Complete & Missing var. \\
\midrule
1  & Surgical 1              &  9 & 2 \\
2  & Surgical 2              &  4 & 3 \\
3  & Surgical 3              & 12 & 5 \\
4  & Surgical 4              & 11 & 3 \\
5  & Surgical 5              &  4 & 1 \\
\midrule
6  & Product / Device 1      &  5 & 3 \\
7  & Product / Device 2      &  4 & 4 \\
8  & Product / Device 3      &  5 & 2 \\
\midrule
9  & Patient Protection 1    &  5 & 3 \\
10 & Patient Protection 2    &  5 & 3 \\
11 & Patient Protection 3    &  6 & 1 \\
\midrule
12 & Care Management 1       &  6 & 5 \\
13 & Care Management 2       &  4 & 2 \\
14 & Care Management 3       &  6 & 4 \\
15 & Care Management 4       &  5 & 2 \\
16 & Care Management 5       &  6 & 3 \\
17 & Care Management 6       &  1 & 0 \\
18 & Care Management 7       &  7 & 2 \\
19 & Care Management 8       &  4 & 2 \\
20 & Care Management 9       &  6 & 0 \\
\midrule
21 & Environmental 1         &  4 & 3 \\
22 & Environmental 2         &  2 & 0 \\
23 & Environmental 3         &  4 & 3 \\
24 & Environmental 4         &  4 & 0 \\
\midrule
25 & Potential Criminal 1    &  2 & 0 \\
26 & Potential Criminal 2    &  2 & 1 \\
27 & Potential Criminal 3    &  4 & 2 \\
28 & Potential Criminal 4    &  5 & 5 \\
\midrule
29 & Radiologic 1            &  4 & 2 \\
\midrule
   & \textbf{Total}          & \textbf{146} & \textbf{66} \\
\bottomrule
\end{tabular}
\caption{Per-clause counts of \textsc{Reportable}\,/\,\textsc{Non\_Reportable}
clause cards and their attached missing-information
variants. \textsc{Uncertain} cards (one per starred clause; $11$
clauses total) are listed separately in Table~\ref{tab:ds-cards-by-cat}.}
\label{tab:ds-clause-counts}
\end{table}

\begin{table}[t]
\centering
\footnotesize
\setlength{\tabcolsep}{3.2pt}
\begin{tabular}{l rr rrr l}
\toprule
\multirow{2}{*}{Field}
 & \multicolumn{2}{c}{Per-card stats}
 & \multicolumn{3}{c}{Histogram}
 & \multirow{2}{*}{Range} \\
\cmidrule(lr){2-3}\cmidrule(lr){4-6}
 & Mean & Med. & $\leq\!2$ & $3$--$4$ & $\geq\!5$ &  \\
\midrule
\multicolumn{7}{l}{\emph{Boundary conditions per card}} \\
\quad Reportable     & 4.44 & 5 &  4 & 18 & 23 & $2$--$7$ \\
\quad Non\_Reportable & 3.61 & 4 & 19 & 64 & 18 & $1$--$7$ \\
\quad Uncertain      & 3.00 & 3 &  1 & 10 &  0 & $2$--$4$ \\
\midrule
\multicolumn{7}{l}{\emph{Basic event elements per card}} \\
\quad Reportable     & 6.36 & 6 &  1 &  7 & 37 & $2$--$10$ \\
\quad Non\_Reportable & 5.27 & 5 &  7 & 31 & 63 & $2$--$11$ \\
\quad Uncertain      & 3.00 & 3 &  1 & 10 &  0 & $2$--$4$ \\
\bottomrule
\end{tabular}
\caption{Distribution of \texttt{boundary\_conditions} and
\texttt{basic\_event\_elements} counts per clause card, broken down
by event type. ``Histogram'' columns group raw counts into
$\leq 2$\,/\,$3$--$4$\,/\,$\geq 5$ bins for compactness.}
\label{tab:ds-card-shape}
\end{table}

\paragraph{Missing-variant structure.}
Across the $61$ parents that carry at least one missing variant,
the fan-out is small: $56$ parents have exactly one variant and
$5$ have two. Each variant withholds a small,
deliberately chosen subset of the parent's
\path{basic_event_elements}; Table~\ref{tab:ds-missing-shape}
reports the distribution of the number of withheld slots
(\path{masked_slot_names}). The modal variant masks a single
slot, but the long tail (up to six masked slots) captures cases
where multiple co-dependent facts must all be withheld for the
residual narrative to be genuinely under-determined.

\begin{table}[t]
\centering
\footnotesize
\setlength{\tabcolsep}{4pt}
\begin{tabular}{l rrrrr rr}
\toprule
$\#$\,withheld slots & 1 & 2 & 3 & 4 & 5--6 & Mean & Med. \\
\midrule
\# variants ($n\!=\!66$) & 32 & 18 & 12 & 2 & 2 & 1.86 & 2 \\
\bottomrule
\end{tabular}
\caption{Distribution of \texttt{masked\_slot\_names} length per
missing-information variant.}
\label{tab:ds-missing-shape}
\end{table}

\paragraph{Per-case dataset.}
Each anchor-card pairing that survives all generation-time
verification gates becomes one benchmark case (\textsc{complete}
or \textsc{missing}); \textsc{Uncertain} cases are generated
similarly from \textsc{Uncertain} cards without missing variants.
The benchmark contains $5{,}074$ cases in total, broken down by
MN29 category $\times$ case type in
Table~\ref{tab:ds-cases-by-cat} and by clause $\times$ case type
in Table~\ref{tab:ds-cases-by-clause}. 

\begin{table}[t]
\centering
\footnotesize
\setlength{\tabcolsep}{3pt}
\begin{tabular}{lrrrr}
\toprule
MN29 category & Compl. & Miss. & Unc. & Total \\
\midrule
Surgical            &   940 & 260 & 43 & 1{,}243 \\
Product / Device    &   321 & 192 & 49 &    562 \\
Patient Protection  &   364 & 170 &  0 &    534 \\
Care Management     & 1{,}088 & 395 & 92 & 1{,}575 \\
Environmental       &   333 & 125 & 25 &    483 \\
Potential Criminal  &   309 & 179 & 24 &    512 \\
Radiologic          &   100 &  41 & 24 &    165 \\
\midrule
\textbf{Total}      & \textbf{3{,}455} & \textbf{1{,}362} & \textbf{257} & \textbf{5{,}074} \\
\bottomrule
\end{tabular}
\caption{Case counts by MN29 category and case type.
``Compl.'' = \textsc{Complete}, ``Miss.'' = \textsc{Missing},
``Unc.'' = \textsc{Uncertain}.}
\label{tab:ds-cases-by-cat}
\end{table}

\begin{table}[t]
\centering
\footnotesize
\setlength{\tabcolsep}{3pt}
\begin{tabular}{c l rrr r}
\toprule
\# & Category (local clause) & Compl. & Miss. & Unc. & Total \\
\midrule
1  & Surgical 1              & 223 &  30 & 23 & 276 \\
2  & Surgical 2              &  99 &  62 &  0 & 161 \\
3  & Surgical 3              & 290 & 112 &  0 & 402 \\
4  & Surgical 4              & 235 &  38 & 20 & 293 \\
5  & Surgical 5              &  93 &  18 &  0 & 111 \\
\midrule
6  & Product / Device 1      &  99 &  60 & 25 & 184 \\
7  & Product / Device 2      &  97 &  84 & 24 & 205 \\
8  & Product / Device 3      & 125 &  48 &  0 & 173 \\
\midrule
9  & Patient Protection 1    & 122 &  75 &  0 & 197 \\
10 & Patient Protection 2    &  98 &  70 &  0 & 168 \\
11 & Patient Protection 3    & 144 &  25 &  0 & 169 \\
\midrule
12 & Care Management 1       & 149 & 116 & 20 & 285 \\
13 & Care Management 2       &  99 &  48 &  0 & 147 \\
14 & Care Management 3       & 137 &  31 &  0 & 168 \\
15 & Care Management 4       & 122 &  45 &  0 & 167 \\
16 & Care Management 5       & 150 &  73 & 25 & 248 \\
17 & Care Management 6       &  25 &   0 &  0 &  25 \\
18 & Care Management 7       & 172 &  47 & 24 & 243 \\
19 & Care Management 8       &  85 &  35 & 23 & 143 \\
20 & Care Management 9       & 149 &   0 &  0 & 149 \\
\midrule
21 & Environmental 1         &  89 &  57 & 25 & 171 \\
22 & Environmental 2         &  50 &   0 &  0 &  50 \\
23 & Environmental 3         &  96 &  68 &  0 & 164 \\
24 & Environmental 4         &  98 &   0 &  0 &  98 \\
\midrule
25 & Potential Criminal 1    &  46 &   0 &  0 &  46 \\
26 & Potential Criminal 2    &  49 &  24 & 24 &  97 \\
27 & Potential Criminal 3    &  98 &  48 &  0 & 146 \\
28 & Potential Criminal 4    & 116 & 107 &  0 & 223 \\
\midrule
29 & Radiologic 1            & 100 &  41 & 24 & 165 \\
\midrule
   & \textbf{Total}          & \textbf{3{,}455} & \textbf{1{,}362} & \textbf{257} & \textbf{5{,}074} \\
\bottomrule
\end{tabular}
\caption{Case counts by global MN29 clause and case type.}
\label{tab:ds-cases-by-clause}
\end{table}

\paragraph{Narrative length.}
Table~\ref{tab:ds-narr-len} reports the distribution of
narrative length (in words) by case type.
\textsc{Complete} narratives average about $170$ words and
\textsc{missing} narratives are systematically shorter (mean
$\sim 109$ words) because masking decision-critical facts
mechanically removes content. \textsc{Uncertain} narratives are
substantially longer (mean $\sim 281$ words) because the
boundary conditions of an \textsc{Uncertain} card frequently
include multiple corroborating clinical facts that must all be
expressed concretely for the underlying ambiguity to remain
genuine.

\begin{table}[t]
\centering
\scriptsize
\setlength{\tabcolsep}{2.6pt}
\begin{tabular}{lrrrrrrrr}
\toprule
Case type & $n$ & Mean & Med. & Min & $p_{25}$ & $p_{75}$ & $p_{95}$ & Max \\
\midrule
Complete  & 3{,}455 & 170 & 165 & 56  & 129 & 202 & 267 & 449 \\
Missing   & 1{,}362 & 109 & 102 & 38  &  83 & 129 & 179 & 260 \\
Uncertain &   257   & 281 & 284 & 155 & 256 & 307 & 344 & 419 \\
\bottomrule
\end{tabular}
\caption{Narrative length (words) by case type.}
\label{tab:ds-narr-len}
\end{table}

\subsection{Generation failure and retry breakdown}
\label{app:retry}


Among the $5{,}074$ accepted cases, almost all succeed on the
first attempt at each stage, but a non-trivial tail relies on the
retry loop (Table~\ref{tab:retry-winning}). The retry tail is
substantially heavier on the missing-narration stage than on
either complete-case stage: only about two thirds of missing
variants pass on attempt $1$, and roughly one in thirteen needs
the third and final attempt to converge. This is consistent with
how subtle the missing-variant contract is in practice: the
narrator must simultaneously preserve every visible fact and
leave every masked boundary condition truly indeterminate.

\begin{table}[t]
\centering
\scriptsize
\setlength{\tabcolsep}{2.4pt}
\begin{tabular}{l rrr}
\toprule
Stage & Attempt 1 & Attempt 2 & Attempt 3 \\
\midrule
Instantiation ($5{,}074$)      & 4{,}282 (84.4\%) & 641 (12.6\%) & 151 (3.0\%) \\
Complete narration ($5{,}074$) & 4{,}630 (91.2\%) & 365 (7.2\%)  &  79 (1.6\%) \\
Missing narration ($1{,}362$)  &    911 (66.9\%) & 350 (25.7\%) & 101 (7.4\%) \\
\bottomrule
\end{tabular}
\caption{Winning-attempt distribution among cases that ultimately
passed verification. Percentages are within-stage.}
\label{tab:retry-winning}
\end{table}

\subsection{Representative end-to-end generation example}
\label{app:e2e}

To make the construction pipeline concrete, this subsection
walks through one full \textsc{complete}-case generation and its
accompanying \textsc{missing} variant on a single real anchor
from the production benchmark, with intermediate verifier
results included. We show a case from
clause card \texttt{CR\_1\_CareManagement\_1} (MN29 Care
Management clause~1, medication error) anchored on a Japan-Council
adverse-event report of a promethazine-related intra-procedural
cardiac arrest.
Exhibits~\ref{exhibit:e2e-loyal}--\ref{exhibit:e2e-mvv-att1}
trace the case from the loyal extraction of the clause text
through the final accepted missing narrative. We do not
re-walk the \textsc{Uncertain} generation pipeline because it
follows the same \textsc{Phase 1} + \textsc{Phase 2} sequence as
a \textsc{complete} case, with no missing-variant follow-up.

\begin{figure*}[t]
\begin{exhibitbox}[label=exhibit:e2e-loyal]{Loyal extraction for Care Management clause~$1$ (excerpt)}
\textbf{Clause text} (verbatim from MN29):\\[2pt]
``Patient death or serious injury associated with a medication error, including, but not limited to, errors involving the wrong drug, the wrong dose, the wrong patient, the wrong time, the wrong rate, the wrong preparation, or the wrong route of administration, excluding reasonable differences in clinical judgment on drug selection and dose.''\\[6pt]

\textbf{Companion guidance loyally extracted from MDH guidance} (cited in the clause card's \texttt{governing\_legal\_basis}):
\begin{itemize}\setlength\itemsep{1pt}
\item \emph{General Recommendation~1 (serious-injury definition).} Defines ``serious injury'' from Minn.\ Stat.\ \S\,144.7063 Subd.~$4$ plus an Inclusion / Exclusion list: e.g.\ higher level of care for more than $48$\,h related to the event, loss / substantial limitation of bodily function lasting more than $7$ days, loss of a body part, fractures (excluding minor digit / nose / rib / wrist fractures unless they meet inclusion criteria), injuries requiring major intervention. ``Inclusion criteria trump exclusion criteria.''
\item \emph{General Recommendation~3 (``associated with'').} The phrase ``associated with'' means the event had a high clinical likelihood of contributing to or causing the outcome. The event \emph{should be reported} unless evidence (e.g.\ autopsy) or a clinical-team determination supports a different cause.
\item \emph{Care Management Event Recommendation~1 (what ``medication error'' captures).} Explicitly includes (a) over-/under-dosing, (b) administration of a medication to which the patient has a \emph{known allergy or serious contraindication}, (c) drug-drug interactions for which there is known potential for death or serious injury, and (d) improper vial use. Explicitly excludes ``allergies that could not reasonably have been known or discerned in advance,'' reactions that could not reasonably have been anticipated, reasonable clinical-judgment differences, and errors of omission from diagnostic delays.
\end{itemize}

\end{exhibitbox}
\end{figure*}

\begin{figure*}[t]
\begin{exhibitbox}[label=exhibit:e2e-card]{Clause card \texttt{CR\_1\_CareManagement\_1} (content elided for space)}
\begin{Verbatim}[fontsize=\footnotesize, commandchars=\\\{\}]
\{
  "clause_card_id":         "CR_1_CareManagement_1",
  "event_type":             "Reportable",
  "clause_id":              "CareManagement_1_MedicationError",
  "clause_card_definition": "A patient dies or sustains a serious injury after being given
                             a medication for which a known serious allergy, contraindication,
                             or drug-drug interaction existed and was reasonably knowable
                             before administration ... (Reportable under this clause.)",

  "fixed_fields": \{
    "governing_legal_basis": \{
      "value":   ["Care Management Events clause 1", "General Recommendation 1",
                  "General Recommendation 3", "Care Management Event Recommendation 1"],
      "meaning": "The clause text and guidance provisions that support the reportable
                  known-risk medication-exposure determination."
    \},
    "boundary_conditions": \{
      "patient_death_or_serious_injury_present":        \{ "value": true,  ... \},
      "outcome_associated_with_medication_exposure":    \{ "value": true,  ... \},
      "preexisting_known_serious_medication_risk_present": \{ "value": true,  ... \}
    \}
  \},

  "basic_event_elements": \{
    "medication_administered":                  \{ "field_type": "string",         ... \},
    "preexisting_known_medication_risk_fact":   \{ "field_type": "string",         ... \},
    "association_assessment_fact":              \{ "field_type": "string",         ... \},
    "outcome_type":                             \{ "field_type": "enum",
                                                   "allowed_values":
                                                       ["death", "serious_injury"], ... \},
    "serious_injury_qualification_fact_or_null":\{ "field_type": "string_or_null", ... \}
  \},

  "constraints_on_basic_event_elements_instantiation": [
    "preexisting_known_medication_risk_fact must describe a known serious allergy, ...",
    "association_assessment_fact must describe a high clinical likelihood that ...",
    "If outcome_type == death, serious_injury_qualification_fact_or_null must be null",
    "If outcome_type == serious_injury, serious_injury_qualification_fact_or_null ..."
  ],

  "missing_information_variants": [
    \{ "missing_variant_id":         "missing_preexisting_known_risk_basis",
      "masked_boundary_conditions": ["preexisting_known_serious_medication_risk_present"],
      "masked_basic_event_elements":["preexisting_known_medication_risk_fact"],
      "summary": "The initial event description says the patient had a severe outcome
                  after receiving a medication, but it does not yet establish whether
                  a serious allergy, contraindication, or dangerous drug-drug interaction
                  was already known before the medication was given.",
      ... \}
  ]
\}
\end{Verbatim}
\end{exhibitbox}
\end{figure*}

\begin{figure*}[t]
\begin{exhibitbox}[label=exhibit:e2e-inst-att1]{Instantiation attempt~$1$: candidate slot values and Instantiation Verifier verdict (\texttt{FAIL})}
\textbf{Candidate \texttt{slot\_values}} (attempt~$1$; the load-bearing field is \path{association_assessment_fact}, which the verifier flags below):
\begin{Verbatim}[fontsize=\footnotesize, commandchars=\\\{\}]
\{
  "medication_administered":
      "Promethazine hydrochloride 12.5 mg IV (single dose for nausea)",
  "preexisting_known_medication_risk_fact":
      "Prior to dosing, the medication reconciliation in the EMR listed a documented
       history of prolonged QTc (QTc 520 ms on ECG two weeks earlier) and the
       decision-support alert identified promethazine as contraindicated due to risk
       of malignant ventricular arrhythmia in patients with marked QT prolongation.",
  "association_assessment_fact":
      "Within 15 minutes of the dose, the patient had an abrupt collapse with loss
       of pulse on telemetry showing a new *polymorphic ventricular tachyarrhythmia*;
       the event occurred before any other new medication or procedure, electrolytes
       and blood glucose drawn immediately afterward were not significantly abnormal,
       and return of circulation occurred after resuscitation with no alternative
       acute trigger identified by the treating team.",
  "outcome_type": "serious_injury",
  "serious_injury_qualification_fact_or_null":
      "The patient required transfer from the medical ward to the ICU for
       post-resuscitation care and mechanical ventilation for 3 days (>48 hours)
       related to the event, meeting the higher-level-of-care serious-injury criterion."
\}
\end{Verbatim}
\textbf{Instantiation Verifier output:} \verb|"pass": false|.\\
\verb|"issues"|:
\begin{itemize}\setlength\itemsep{1pt}
\item ``\path{association_assessment_fact} violates the instantiation constraint by naming the specific reaction syndrome/risk-class (`polymorphic ventricular tachyarrhythmia'), which is disallowed for this field. The causation evidence should be described without naming the specific syndrome/risk class.''
\end{itemize}
This is the verdict-leak check at work: the clause card's instantiation constraints forbid the \path{association_assessment_fact} slot from naming any reaction syndrome or risk class because doing so would prejudge the masked-knowability axis that the missing variant relies on. The rhythm name (``polymorphic ventricular tachyarrhythmia'') is, in clinical context, tightly coupled to a \emph{known QT-prolongation contraindication}, so naming it would let a clinically literate reader infer the masked boundary condition even from the visible facts of a future missing variant. The Instantiator receives the \verb|issues| string as a \textsc{Retry Feedback} block and tries again. 
\end{exhibitbox}
\end{figure*}

\begin{figure*}[t]
\begin{exhibitbox}[label=exhibit:e2e-inst-att2]{Instantiation attempt~$2$: candidate slot values (accepted) and Instantiation Verifier verdict (\texttt{PASS})}
\begin{Verbatim}[fontsize=\footnotesize, commandchars=\\\{\}]
\{
  "medication_administered": "Promethazine hydrochloride 12.5 mg IV (single dose)",
  "preexisting_known_medication_risk_fact":
      "Before administration, the EMR problem list and a cardiology note documented
       marked QTc prolongation (QTc 520 ms on an ECG two weeks earlier), and
       computerized order entry displayed a contraindication warning for promethazine
       in patients with significant QT prolongation due to risk of life-threatening
       rhythm complications.",
  "association_assessment_fact":
      "About 10 minutes after the IV dose, the patient had sudden unresponsiveness
       with loss of palpable pulse requiring immediate CPR and defibrillation;
       no other new medication, contrast, or procedure occurred in that interval,
       point-of-care glucose and electrolytes obtained right after the event were
       not significantly abnormal, and the treating team documented that the temporal
       relationship to the dose with absence of an alternative acute trigger made
       the medication exposure a likely contributor.",
  "outcome_type": "serious_injury",
  "serious_injury_qualification_fact_or_null":
      "After return of circulation, the patient was transferred from the medical ward
       to the ICU for post-arrest care and required mechanical ventilation and
       vasopressor support for 72 hours (>48 hours), meeting the higher-level-of-care
       serious-injury criterion."
\}
\end{Verbatim}
\textbf{Instantiation Verifier output:} \verb|"pass": true|; \verb|"issues": []|. The accepted version surgically separates the two axes the verifier guards: the \emph{knowability} fact (the QT-prolongation contraindication and the order-entry warning) is named in \path{preexisting_known_medication_risk_fact}, where naming a known risk is allowed and required; while the \emph{association} fact uses only behaviorally-observable evidence (timing of collapse after the dose, lack of an alternative acute trigger, and exclusion of metabolic causes), with no rhythm-class label. This is the canonical structured event handed to the Event Narrator.
\end{exhibitbox}
\end{figure*}

\begin{figure*}[t]
\begin{exhibitbox}[label=exhibit:e2e-narr-att1]{Event Narrator attempt~$1$ (accepted): narrative, Extractor reconstruction, and Narrative Verifier verdict (\texttt{PASS})}
\textbf{Event narrative} (attempt~$1$, verbatim):\\
\emph{Patient was admitted to the medical ward for evaluation and symptom management. Prior to the event, the EMR problem list and a recent cardiology note documented marked QTc prolongation (QTc 520\,ms on ECG obtained two weeks earlier). During computerized order entry for antiemetic/sedation, the system displayed a contraindication warning for promethazine in patients with significant QT prolongation due to risk of life-threatening rhythm complications.}\\
\emph{Despite the documented QTc history and the on-screen warning, the patient received promethazine hydrochloride 12.5\,mg IV as a single dose. Approximately 10 minutes after administration, the patient became suddenly unresponsive and was found without a palpable pulse. A code was called immediately; CPR was initiated and the patient required defibrillation. Return of spontaneous circulation was achieved.}\\
\emph{In the immediate post-event assessment, there were no other new medications administered and no contrast exposure or procedures in the interval between the dose and the arrest. Point-of-care glucose and electrolytes drawn right after the event were not significantly abnormal. The treating team documented that, given the close temporal relationship to the promethazine dose and the absence of another acute trigger, the medication exposure was felt to be a likely contributor.}\\
\emph{Following stabilization, the patient was transferred from the medical ward to the ICU for post-arrest care and required mechanical ventilation and vasopressor support for approximately 72 hours.}\\[4pt]

\textbf{Extractor reconstruction} (all $5$ slots recovered in substance):
\begin{Verbatim}[fontsize=\footnotesize, commandchars=\\\{\}]
\{
  "medication_administered": "promethazine hydrochloride 12.5 mg IV",
  "preexisting_known_medication_risk_fact":
      "The EMR problem list and a recent cardiology note documented marked QTc
       prolongation (QTc 520 ms), and computerized order entry displayed a
       contraindication warning for promethazine in patients with significant QT
       prolongation due to risk of life-threatening rhythm complications.",
  "association_assessment_fact":            "...",
  "outcome_type":                           "serious_injury",
  "serious_injury_qualification_fact_or_null":
      "The patient had a cardiac arrest requiring CPR and defibrillation with ROSC,
       followed by transfer to the ICU for post-arrest care with mechanical ventilation
       and vasopressor support for approximately 72 hours."
\}
\end{Verbatim}

\textbf{Narrative Verifier output:} \verb|"pass": true|; \verb|"issues": []|. Extractor successfully recovered the canonical \path{slot_values} value and verifier found the extracted facts to be semantically equivalent to the canonical facts.
\end{exhibitbox}
\end{figure*}

\begin{figure*}[t]
\begin{exhibitbox}[label=exhibit:e2e-mask]{Missing-variant masking specification}
\textbf{Variant id:} \texttt{missing\_preexisting\_known\_risk\_basis}\\
\textbf{Masked boundary condition:} \path{preexisting_known_serious_medication_risk_present}.\\
\textbf{Masked basic-event-element slot:} \path{preexisting_known_medication_risk_fact} (single slot).\\
\textbf{Visible slots (preserved verbatim from the accepted instantiation):} \path{medication_administered}, \path{association_assessment_fact}, \path{outcome_type}, \path{serious_injury_qualification_fact_or_null}.\\[4pt]
\textbf{Variant summary}: ``The initial event description says the patient had a severe outcome after receiving a medication, but it does not yet establish whether a serious allergy, contraindication, or dangerous drug-drug interaction was already known before the medication was given.''\\[4pt]
\textbf{Why under-determined}: without the preexisting-known-risk facts, the residual narrative could fit \emph{either} this \textsc{Reportable} known-risk card \emph{or} the \textsc{Non\_Reportable} ``unknowable-reaction'' sibling card on the same clause.
\end{exhibitbox}
\end{figure*}

\begin{figure*}[t]
\begin{exhibitbox}[label=exhibit:e2e-mvv-att1]{Missing-Variant Narrator attempt~$1$ (accepted), Extractor reconstruction, and Missing-Variant Verifier verdict (\texttt{PASS})}
\textbf{Missing narrative} (attempt~$1$):\\
\emph{Patient was admitted to a mixed neurology/neurosurgery/general medicine ward and was awaiting a planned diagnostic study later that day. Per the care plan, the patient received promethazine hydrochloride 12.5\,mg IV as a single dose for symptom management prior to the study. Approximately 10 minutes after the IV dose was given, the patient suddenly became unresponsive and no palpable pulse was detected. A code was called and immediate CPR was initiated; the patient required defibrillation during the resuscitation.}\\
\emph{No other new medication, contrast administration, or procedure occurred in the interval between the IV dose and the collapse. Point-of-care glucose and electrolytes obtained immediately after the event were not significantly abnormal. After return of circulation, the patient was transferred from the medical ward to the ICU for post-arrest care and required mechanical ventilation and vasopressor support for 72 hours. In the treating team's documentation, the close temporal relationship to the IV dose and the lack of an alternative acute trigger were noted as making the medication exposure a likely contributor to the arrest.}\\[4pt]

\textbf{Extractor reconstruction} (the masked slot correctly returns \texttt{null}; visible slots recovered in substance):
\begin{Verbatim}[fontsize=\footnotesize, commandchars=\\\{\}]
\{
  "medication_administered":
      "promethazine hydrochloride 12.5 mg IV (single dose)",
  "preexisting_known_medication_risk_fact": null,
  "association_assessment_fact":
      "The patient collapsed about 10 minutes after the IV dose; no other new medication,
       contrast, or procedure occurred in the interval, point-of-care glucose/electrolytes
       were not significantly abnormal, and the treating team documented the close
       temporal relationship and lack of an alternative acute trigger as making the
       medication exposure a likely contributor.",
  "outcome_type": "serious_injury",
  "serious_injury_qualification_fact_or_null":
      "The patient experienced a cardiac arrest requiring CPR and defibrillation with
       subsequent transfer to the ICU for post-arrest care, including mechanical
       ventilation and vasopressor support for 72 hours."
\}
\end{Verbatim}

\textbf{Missing-Variant Verifier output:} \verb|"pass": true|; \verb|"issues": []|; \path{per_bc_check} row with \path{leak_verdict}: \texttt{"indeterminate"}; \path{explanation}: ``The narrative describes the medication administration and the subsequent arrest with temporal-association reasoning, but it does not state or imply anything about a preexisting allergy / contraindication / interaction being known or documented before dosing. Both possibilities (known serious risk present vs.\ not) remain consistent with the text.''
\end{exhibitbox}
\end{figure*}


\section{Evaluation Environment and Metrics}
\label{app:metrics}

\subsection{Per-Case Notation}
\label{app:notation}

For each benchmark case $i\in\mathcal{D}$ ($|\mathcal{D}|=5{,}074$),
the underlying clause card supplies the ground-truth fields
\begin{itemize}
\item $\tau_i$: case type, taking values in
    $\{\textsc{complete},\textsc{missing},\textsc{uncertain}\}$;
\item $y_i^{\star}\in\mathcal{Y}=\{\text{Reportable},\text{Non-reportable},\text{Uncertain}\}$:
    gold verdict;
\item $c_i^{\star}$: gold targeted clause (defined only when $y_i^{\star}=\text{Reportable}$);
\item $G_i^{\star}$: gold governing legal basis (a set of canonical
    clause/guidance identifiers; cf.~the clause-card component $G$);
\item $B_i=\{(b_{i,k},m_{i,k},v_{i,k})\}_{k=1}^{K_i}$: the clause card's
    boundary conditions, each a (name, natural-language meaning,
    truth value) triple;
\item $E_i'^{\,\star}\subseteq E_i$: the withheld basic event elements
    that define a \textsc{missing} variant (empty otherwise; cf.~$E'$
    in the missing-variant definition).
\end{itemize}
The evaluated LLM emits, for each case,
\begin{itemize}
\item $\hat y_i\in\mathcal{Y}\cup\{\text{null}\}$: predicted verdict
    (\text{null} on parse failure);
\item $\hat c_i$: predicted targeted clause;
\item $\hat G_i$: predicted set of clause/guidance citations;
\item $\hat r_i$: free-text rationale;
\item $a_i\in\{0,1\}$: whether the evaluated LLM issued at least one
    \textsc{Ask} action;
\item $\hat E_i'$: the set of basic event elements actually recovered
    through the evaluated LLM's \textsc{Ask} actions, as resolved by
    the Information Provider against the case's instantiated record.
\end{itemize}
We write $\mathrm{PRF}(\mathrm{tp},\mathrm{fp},\mathrm{fn})$ for the
standard binary triple
$\big(P=\tfrac{\mathrm{tp}}{\mathrm{tp}+\mathrm{fp}},\;
R=\tfrac{\mathrm{tp}}{\mathrm{tp}+\mathrm{fn}},\;
F_1=\tfrac{2PR}{P+R}\big)$.

\subsection{Metric Definitions}
\label{app:metric-defs}

\paragraph{M1 -- Verdict Accuracy.}
Three-way classification accuracy over $\mathcal{Y}$, on all $N=5{,}074$
cases:
\begin{equation}
\mathrm{M1} \;=\; \tfrac{1}{N}\sum_{i=1}^{N}\mathbf{1}[\hat y_i=y_i^{\star}].
\end{equation}
Parse failures and any verdict outside $\mathcal{Y}$ are counted as
incorrect. Per-case-type stratifications
(Tab.~\ref{tab:m1_by_casetype}) restrict the sum to cases with the
stratum's $\tau_i$.

\paragraph{M2 -- Clause Accuracy.}
On the subset where gold and predicted verdicts are both Reportable,
$\mathcal{R}=\{i:y_i^{\star}=\hat y_i=\text{Reportable}\}$,
\begin{equation}
\mathrm{M2} \;=\; \tfrac{1}{|\mathcal{R}|}\sum_{i\in\mathcal{R}}\mathbf{1}[\hat c_i=c_i^{\star}].
\end{equation}

\paragraph{M3 -- Evidence Citation F1.}
On the triage-correct subset with non-empty gold legal basis,
$\mathcal{E}=\{i:\hat y_i=y_i^{\star},\;|G_i^{\star}|>0\}$, set
$\mathrm{tp}_i=|G_i^{\star}\cap\hat G_i|$,
$\mathrm{fp}_i=|\hat G_i\setminus G_i^{\star}|$,
$\mathrm{fn}_i=|G_i^{\star}\setminus\hat G_i|$, and pool across cases:
\begin{equation}
\mathrm{M3}\;=\;F_1\!\left(
\textstyle\sum_{i\in\mathcal{E}}\mathrm{tp}_i,\;
\sum_{i\in\mathcal{E}}\mathrm{fp}_i,\;
\sum_{i\in\mathcal{E}}\mathrm{fn}_i
\right).
\end{equation}

\paragraph{M4 -- Boundary Condition Hit Rate.}
On the triage-correct subset with at least one boundary condition,
$\mathcal{H}=\{i:\hat y_i=y_i^{\star},\;K_i>0\}$, an LLM judge 
takes the rationale $\hat r_i$ and the boundary-condition list $B_i$
and returns the subset $H_i\subseteq\{b_{i,k}\}$ whose underlying
concept the rationale actually invokes and treats consistently with
$v_{i,k}$. Empty rationales bypass the judge with $|H_i|=0$. The
metric pools BCs across cases:
\begin{equation}
\mathrm{M4} \;=\;
\frac{\sum_{i\in\mathcal{H}}|H_i|}{\sum_{i\in\mathcal{H}}K_i}.
\end{equation}

\paragraph{M5 -- Missing Case Detection F1.}
Binary F1 on all $N=5{,}074$ cases with positive class
$\mathbf{1}[\tau_i=\textsc{missing}]$ and prediction
$\mathbf{1}[a_i=1]$:
\begin{align*}
\mathrm{tp} &= |\{i:\tau_i=\textsc{missing},\,a_i=1\}|, \\
\mathrm{fp} &= |\{i:\tau_i\neq\textsc{missing},\,a_i=1\}|, \\
\mathrm{fn} &= |\{i:\tau_i=\textsc{missing},\,a_i=0\}|,
\end{align*}
\begin{equation}
\mathrm{M5} \;=\; F_1(\mathrm{tp},\mathrm{fp},\mathrm{fn}).
\end{equation}
Asking on a complete or uncertain case is a false positive; not asking
on a missing case is a false negative. Including uncertain cases in
the negative class penalizes asking on gray-zone events, which is
itself a behavioral error.

\paragraph{M6 -- Missing Slot Identification F1.}
On missing cases for which the evaluated LLM actually asked,
$\mathcal{S}=\{i:\tau_i=\textsc{missing},\,a_i=1,\,|E_i'^{\,\star}|>0\}$,
the predicted slot set $\hat E_i'$ is compared to the gold withheld
set $E_i'^{\,\star}$:
$\mathrm{tp}_i=|E_i'^{\,\star}\cap\hat E_i'|$,
$\mathrm{fp}_i=|\hat E_i'\setminus E_i'^{\,\star}|$,
$\mathrm{fn}_i=|E_i'^{\,\star}\setminus\hat E_i'|$,
and
\begin{equation}
\mathrm{M6}\;=\;F_1\!\left(
\textstyle\sum_{i\in\mathcal{S}}\mathrm{tp}_i,\;
\sum_{i\in\mathcal{S}}\mathrm{fp}_i,\;
\sum_{i\in\mathcal{S}}\mathrm{fn}_i
\right).
\end{equation}
Restricting to cases with $a_i=1$ isolates the quality of the
evaluated LLM's questions from the upstream decision to ask (already
captured by M5).

\paragraph{M7 -- Uncertain Detection F1.}
Binary F1 on all $N=5{,}074$ cases with positive class
$\mathbf{1}[\tau_i=\textsc{uncertain}]$ and prediction
$\mathbf{1}[\hat y_i=\text{Uncertain}]$:
\begin{align*}
\mathrm{tp} &= |\{i:\tau_i=\textsc{uncertain},\,\hat y_i=\text{Uncertain}\}|, \\
\mathrm{fp} &= |\{i:\tau_i\neq\textsc{uncertain},\,\hat y_i=\text{Uncertain}\}|, \\
\mathrm{fn} &= |\{i:\tau_i=\textsc{uncertain},\,\hat y_i\neq\text{Uncertain}\}|,
\end{align*}
\begin{equation}
\mathrm{M7} \;=\; F_1(\mathrm{tp},\mathrm{fp},\mathrm{fn}).
\end{equation}

\paragraph{M8 -- Reportable Detection F1.}
Binary F1 on all $N=5{,}074$ cases with positive class
$\mathbf{1}[y_i^{\star}=\text{Reportable}]$ and prediction
$\mathbf{1}[\hat y_i=\text{Reportable}]$:
\begin{align*}
\mathrm{tp} &= |\{i:y_i^{\star}=\text{Reportable},\,\hat y_i=\text{Reportable}\}|, \\
\mathrm{fp} &= |\{i:y_i^{\star}\neq\text{Reportable},\,\hat y_i=\text{Reportable}\}|, \\
\mathrm{fn} &= |\{i:y_i^{\star}=\text{Reportable},\,\hat y_i\neq\text{Reportable}\}|,
\end{align*}
\begin{equation}
\mathrm{M8} \;=\; F_1(\mathrm{tp},\mathrm{fp},\mathrm{fn}).
\end{equation}
The positive class covers both complete and missing cases that
resolve to Reportable.

\subsection{Judge Model}
\label{app:judge-infra}

The M4 judge is \texttt{openai/gpt-5.2} with \texttt{temperature}=0
and JSON-only outputs. The prompt enforces
a strict schema; non-conforming outputs are retried up to three
times. Hits returned outside the per-case boundary-condition vocabulary
are dropped before aggregation. Cases with no boundary conditions are
skipped. The system prompt and user prompt template used by the M4 judge are
shown in Prompts~\ref{prompt:m4-system}
and~\ref{prompt:m4-user}, respectively.

\begin{figure*}[t]
\centering
\begin{promptbox}[label=prompt:m4-system]{M4 Judge -- System Prompt}
You are a careful legal/clinical reasoner judging whether a model's
free-text rationale for a reportability decision actually invokes the
underlying boundary conditions of the relevant decision unit.

\medskip
You will be given:\\
1. The full text of the model's rationale.\\
2. A list of boundary conditions for the case. Each entry has a name,
a natural-language meaning, and the boundary condition's truth value
(true or false) in this specific case.

\medskip
Your job. For each boundary condition, decide whether the rationale
actually invokes that boundary condition's underlying concept and
arrives at (or is consistent with) its truth value. A ``hit'' requires
both: (a) the rationale touches the same concept the boundary
condition is about, AND (b) the rationale is consistent with the truth
value (e.g.\ if the BC is true, the rationale should treat it as
satisfied; if false, the rationale should treat it as not satisfied).
Mere mention of related vocabulary without engaging the concept is
NOT a hit. Hidden or implicit mentions count as long as the rationale
clearly relies on that concept to reach its verdict.

\medskip
Return JSON only.

\medskip
\# Output schema

\medskip
\{\\
\hspace*{1em}"hits": [<boundary\_condition\_name\_string>, ...],\\
\hspace*{1em}"explanations": \{\\
\hspace*{2em}"<boundary\_condition\_name>": "<one short sentence stating why the rationale hits or misses this BC>"\\
\hspace*{1em}\}\\
\}

\medskip
Constraints:\\
- ``hits'' must be a subset of the provided BC names; do not invent.\\
- Every BC supplied to you must have an entry in ``explanations''.\\
- No keys other than ``hits'' and ``explanations''.\\
- Use exact BC names (case-sensitive, verbatim).
\end{promptbox}
\end{figure*}

\begin{figure*}[t]
\centering
\begin{promptbox}[label=prompt:m4-user]{M4 Judge -- User Prompt Template}
Rationale to evaluate:

\medskip
\{rationale\}

\medskip
Boundary conditions for the case (each is a named criterion with a
true/false truth value and a natural-language meaning):

\medskip
- name: \{bc\_name\}\\
\hspace*{1em}value: \{bc\_value\}\\
\hspace*{1em}meaning: \{bc\_meaning\}\\
- name: ...

\medskip
Emit exactly one JSON object as specified by the system message ---
no Markdown fences, no prose.
\end{promptbox}
\end{figure*}

\section{Tested LLMs and Evaluation Infrastructure}
\label{app:llms-infra}

\subsection{Evaluated LLMs}
\label{app:tested-llms}

We evaluate 15 LLMs spanning four families: closed-source
frontier models, open-source frontier models, small general-purpose
models, and medical-specialty models. We briefly describe each model
below; reported parameter counts are taken from each model's release
notes / technical report.

\paragraph{Closed-source frontier.}
\begin{itemize}\itemsep0pt
\item \textbf{GPT-5.5} --- OpenAI's most recent frontier reasoning
    model and the strongest member of the GPT-5 family; emits an
    internal reasoning trace before the final answer. In our runs we
    leave \texttt{reasoning\_effort} at the API default of
    \texttt{medium}.
\item \textbf{GPT-5} --- OpenAI's general-purpose frontier reasoning
    model in the GPT-5 family. In our runs \texttt{reasoning\_effort}
    is at the API default of \texttt{medium}.
\item \textbf{Claude Opus 4.7} --- Anthropic's largest model in the
    Claude 4.x generation; a hybrid model that supports extended
    thinking. In our runs we do not pass a \texttt{thinking}
    parameter, so extended thinking is disabled (Anthropic's
    default).
\item \textbf{Claude Sonnet 4.6} --- Anthropic's mid-tier Claude 4.x
    model with the same extended-thinking capability. Extended
    thinking is likewise disabled in our runs.
\item \textbf{Gemini 3.1 Pro} --- One of Google DeepMind's largest reasoning
    models in the Gemini 3.x line, accessed via the preview release.
    In our runs the thinking budget is left at the API default (the
    model decides its own per-request thinking budget).
\item \textbf{Gemini 2.5 Flash} --- Google DeepMind's faster Gemini
    2.5 variant. Gemini 2.5 supports an optional ``thinking'' mode;
    in our runs we do not pass a \texttt{thinking\_config}, so the
    model uses Google's default per-request behavior for this slug.
\end{itemize}

\paragraph{Open-source frontier.}
\begin{itemize}\itemsep0pt
\item \textbf{DeepSeek-R1} --- a $671$B-parameter Mixture-of-Experts
    reasoning model from DeepSeek with $\approx\!37$B active
    parameters per token.
\item \textbf{Qwen3-235B-A22B-Instruct} --- Alibaba's $235$B-parameter
    MoE instruction-tuned model with $\approx\!22$B active parameters
    per token, from the Qwen3 series (release tag
    \texttt{2507}).
\item \textbf{GPT-OSS-120B} --- OpenAI's open-weight $117$B-parameter
    MoE reasoning model with $\approx\!5.1$B active parameters per
    token.
\end{itemize}

\paragraph{Small general-purpose.}
\begin{itemize}\itemsep0pt
\item \textbf{GPT-5-nano} --- the smallest member of the GPT-5
    reasoning-model family. In our runs \texttt{reasoning\_effort} is
    at the API default of \texttt{medium}.
\item \textbf{Llama-3.1-8B-Instruct} --- Meta's $8$B-parameter dense
    instruction-tuned Llama-3.1 model.
\item \textbf{Mistral-Small-3.2-24B-Instruct} --- Mistral AI's
    $24$B-parameter dense instruction-tuned model (release tag
    \texttt{2506}); a non-reasoning model.
\end{itemize}

\paragraph{Medical-specialty.}
\begin{itemize}\itemsep0pt
\item \textbf{HuatuoGPT-o1-8B} --- an $8$B-parameter medical
    reasoning model from FreedomIntelligence, built on top of
    Llama-3.1-8B-Instruct via verifiable medical-reasoning
    fine-tuning and reinforcement learning. Emits an internal
    reasoning trace.
\item \textbf{HuatuoGPT-o1-70B} --- a $70$B-parameter version in the
    same HuatuoGPT-o1 line, built on top of Llama-3.1-70B-Instruct
    with the same medical reasoning recipe.
\item \textbf{MedGemma-27B-text-it} --- Google's $27$B-parameter
    medical-domain text instruction-tuned model from the MedGemma
    release, based on the Gemma 3 backbone; a non-reasoning model.
\end{itemize}

\subsection{Evaluation Loop and Information Provider}
\label{app:eval-loop}

Each case is evaluated under a multi-turn budget of at most $T=10$
turns. Whenever this cap is reached the evaluated LLM is forced to
emit an \textsc{Answer} action on the final turn via the directive in
Prompt~\ref{prompt:eval-force}.

The Information Provider role is held fixed to \textbf{GPT-5.2}
across all 15 evaluation runs. Holding the Information Provider
fixed isolates differences between evaluated LLMs from confounds in
the fact-lookup oracle (the provider is described as a stateless,
fact-list-restricted role in
Appendix~\ref{app:ip-prompts}).

\subsection{Evaluated LLM Prompts}
\label{app:eval-prompts}

The evaluated LLM is given the role of a clinical safety reviewer
working against the MN29 statute and its accompanying guidance, with a
tool-style interface for issuing \textsc{Ask} actions to the
Information Provider. Its system prompt
(Prompt~\ref{prompt:eval-system}) bundles four components in order: a
role specification with the three permitted verdicts, the tool
definition for \textsc{Ask}/\textsc{Answer}, the JSON output schema
and field rules, and the verbatim reference materials (the
$29$-entry targeted-clause vocabulary, the canonical evidence
vocabulary, and the full MN29 clause text and official guidance
document). The user-turn templates that drive the multi-turn loop ---
the initial event narrative, the Information-Provider followup, and
the force-answer directive --- are shown in
Prompts~\ref{prompt:eval-initial}, \ref{prompt:eval-followup}, and
\ref{prompt:eval-force}.

\begin{figure*}[t]
\begin{promptbox}[label=prompt:eval-system]{Evaluated LLM --- system prompt}
You are a clinical safety reviewer judging whether a single medical
event is reportable under Minnesota's 29 Reportable Adverse Health
Events (MN29). You will be shown the MN29 clause text (29 statutory
clauses) and the official guidance document (Recommendations +
Appendices). For each event you must decide one of three verdicts:

\medskip
- ``Reportable'' --- the event clearly satisfies one of the 29 MN29
  clauses given the available facts and the guidance.\\
- ``Non-reportable'' --- the event clearly does NOT satisfy any of
  the 29 MN29 clauses given the available facts and the guidance.\\
- ``Uncertain'' --- even with the available facts, MN29 + guidance
  themselves do not yield a determinate Reportable / Non-reportable
  answer, and the case should be escalated for human expert review.
  Pick ``Uncertain'' ONLY when the law/guidance itself is what leaves
  the question open, NOT when key facts are missing --- for missing
  facts use the ASK action.

\medskip
\# Tools

\medskip
You have one external tool: an information provider that can look up
specific facts about the event. Use it when you believe a specific
factual gap is preventing you from deciding (there is a lack of
information to make a decision).

\medskip
- ASK: send a concrete factual question to the information provider.
  Each question must target a specific fact (e.g.\ ``What was the
  patient's documented age at admission?''). Do NOT ask broad/vague
  questions; the provider will refuse them. The provider has NO
  memory between turns --- every question must be self-contained.\\
- ANSWER: emit your final verdict. After ANSWER no further
  interaction takes place.

\medskip
\# Output format

\medskip
On EVERY turn you MUST output exactly one JSON object and nothing
else (no Markdown fences, no prose, no comments). The JSON object has
this exact schema:

\medskip
\{\\
\hspace*{1em}``action'': ``ASK'' \textbar{} ``ANSWER'',\\
\hspace*{1em}``ask\_question'': <string or null>,\\
\hspace*{1em}``final\_verdict'': ``Reportable'' \textbar{} ``Non-reportable'' \textbar{} ``Uncertain'' \textbar{} null,\\
\hspace*{1em}``targeted\_clause'': <string or null>,\\
\hspace*{1em}``clause\_and\_guidance\_evidence'': [<identifier\_string>, ...] \textbar{} null,\\
\hspace*{1em}``rationale'': <string or null>\\
\}

\medskip
Field rules:

\medskip
- When \texttt{action} is ``ASK'': \texttt{ask\_question} must be a
  specific factual question; \texttt{final\_verdict},
  \texttt{targeted\_clause}, \texttt{clause\_and\_guidance\_evidence},
  and \texttt{rationale} must all be \texttt{null}.\\
- When \texttt{action} is ``ANSWER'': \texttt{ask\_question} must be
  \texttt{null}; \texttt{final\_verdict} must be one of the three
  verdict strings; \texttt{targeted\_clause} identifies the single
  MN29 clause driving a Reportable verdict and must be \texttt{null}
  for the other two verdicts; \texttt{clause\_and\_guidance\_evidence}
  is the list of MN29 sections you relied on (required for all three
  verdicts), each drawn verbatim from the Evidence vocabulary below
  (no free-text, no duplicates); \texttt{rationale} is a concise
  explanation tying determining facts in the narrative (and any
  answers from the information provider) to the cited clause /
  guidance / Recommendation.

\medskip
\# Targeted-clause vocabulary (29 entries; case-sensitive, verbatim)

\medskip
\textit{[29 canonical clause labels, e.g.\ ``Surgical Events clause 1'',
``Product or Device Events clause 5''.]}

\medskip
\# Evidence vocabulary (case-sensitive, verbatim)

\medskip
\textit{[Canonical identifiers for the 29 clauses plus General /
Per-event Recommendations, Appendices, and Definitions sections of
the MN29 guidance document; $\approx\!150$ entries.]}

\medskip
\# Reference: MN29 clause text

\medskip
\textit{[Full text of the MN29 statute]}

\medskip
\# Reference: MN29 guidance text

\medskip
\textit{[Full text of the official MN29 guidance document
(Recommendations + Appendices + Definitions)]}
\end{promptbox}
\end{figure*}

\begin{figure*}[t]
\begin{promptbox}[label=prompt:eval-initial]{Evaluated LLM --- initial user turn}
Event narrative to judge:

\medskip
\{event\_narrative\}

\medskip
Emit exactly one JSON object as specified.
\end{promptbox}
\end{figure*}

\begin{figure*}[t]
\begin{promptbox}[label=prompt:eval-followup]{Evaluated LLM --- user turn after Information Provider reply}
\textit{(One of the three prefixes below is selected based on the
provider's response status.)}

\medskip
\texttt{status=answered}:\\
``Information provider response (status=answered):''

\medskip
\texttt{status=refused\_too\_vague}:\\
``Information provider response (status=refused\_too\_vague). Your
question was too broad / not specific enough; rephrase to ask about
ONE concrete fact:''

\medskip
\texttt{status=unknown}:\\
``Information provider response (status=unknown). The provider has no
record of the requested fact:''

\medskip
Then the provider's natural-language reply is appended verbatim:

\medskip
\{answer\_to\_eval\}

\medskip
Emit your next action as a single JSON object.
\end{promptbox}
\end{figure*}

\begin{figure*}[t]
\begin{promptbox}[label=prompt:eval-force]{Evaluated LLM --- force-answer directive (turn $T$)}
You have reached the maximum number of turns. You MUST now output an
ANSWER action; further ASK actions are not allowed. Emit a single
JSON object with \texttt{action}=``ANSWER'' and the required fields.
\end{promptbox}
\end{figure*}

\subsection{Information Provider Prompts}
\label{app:ip-prompts}

The Information Provider is a stateless fact-lookup role with no
memory between calls. On each call it sees the case's complete
\emph{basic event element} list (a field name, a natural-language
meaning, and a concrete value for each element) together with the
evaluated LLM's single factual question, and must return one of three
statuses --- \texttt{answered}, \texttt{refused\_too\_vague}, or
\texttt{unknown} --- in a strict JSON schema. Its system prompt is
shown in Prompt~\ref{prompt:ip-system} and its user-turn template in
Prompt~\ref{prompt:ip-user}.

\begin{figure*}[t]
\begin{promptbox}[label=prompt:ip-system]{Information Provider --- system prompt}
You are a stateless information lookup service. You answer ONE
factual question per call. You have NO memory between calls.

\medskip
Your knowledge consists ONLY of the structured fact list provided in
the user message. Each item is a field name with its meaning and
concrete value. You may NOT use general medical knowledge,
common-sense extrapolation, or any reasoning beyond restating facts
that are present in the list.

\medskip
Behavior rules

\medskip
1. \textbf{Faithfulness.} If the question references a specific fact
   that exists in the list, answer with the literal value of that
   fact (or a close paraphrase that does NOT add information). Cite
   the field name(s) you relied on in \texttt{fields\_used}.\\
2. \textbf{Refusal on vague questions.} If the question is broad,
   asks for ``all you know'' / ``everything about the event'' / ``any
   other context'' / or does not point to a single concrete fact,
   set \texttt{status} to ``refused\_too\_vague'', tell the caller to
   ``ask more specifically'', but you should never advise the caller
   on exactly what to ask, in order to prevent information leakage.
   Leave \texttt{fields\_used} empty.\\
3. \textbf{Unknown.} If the question is specific but the fact list
   contains no relevant entry, set \texttt{status} to ``unknown'',
   say so plainly, and leave \texttt{fields\_used} empty. Do NOT
   invent. Do not leak other information unrelated to the
   question.\\
4. \textbf{Faithful translation.} NEVER summarize a concrete value
   down to a yes/no. If the caller asks ``Was X true?'' and the list
   contains a specific value (e.g.\ ``X = the patient was admitted to
   ICU for 52 hours''), reply with that concrete value rather than
   just ``Yes''.\\
5. \textbf{No leakage.} \textbf{NEVER} reveal information unrelated
   to the question.\\
6. \textbf{Brevity.} Keep \texttt{answer\_to\_eval} concise while not
   losing information.

\medskip
\# Output format

\medskip
Return exactly one JSON object and nothing else:

\medskip
\{\\
\hspace*{1em}``status'': ``answered'' \textbar{} ``refused\_too\_vague'' \textbar{} ``unknown'',\\
\hspace*{1em}``answer\_to\_eval'': ``<natural-language reply, faithful to the fact list>'',\\
\hspace*{1em}``fields\_used'': [<field\_name\_string>, ...]\\
\}

\medskip
Constraints:

\medskip
- \texttt{status} MUST be one of the three strings above.\\
- When \texttt{status} is ``answered'', \texttt{fields\_used} MUST be
  a non-empty list of field names that came from the provided fact
  list (verbatim).\\
- When \texttt{status} is ``refused\_too\_vague'' or ``unknown'',
  \texttt{fields\_used} MUST be an empty list.
\end{promptbox}
\end{figure*}

\begin{figure*}[t]
\begin{promptbox}[label=prompt:ip-user]{Information Provider --- user-turn template}
Fact list (each item: field name + meaning + concrete value):

\medskip
- field: \{field\_name\}\\
\hspace*{1em}meaning: \{field\_meaning\}\\
\hspace*{1em}value: \{field\_value\}\\
- field: ...

\medskip
Caller's question:

\medskip
\{question\}

\medskip
Emit exactly one JSON object as specified by the system message.
\end{promptbox}
\end{figure*}

\section{Extended Results}
\label{app:ext-results}

This appendix extends the main paper's Table~\ref{tab:overall_results},
\ref{tab:m1_by_casetype}, and \ref{tab:behavioral_profile} along
four axes: (i) precision / recall / F1 decompositions of every
F1-style metric in the leaderboard; (ii) evidence-citation (M3) and
rationale--boundary-condition (M4) results stratified by case type;
(iii) the full verdict-routing distribution on the $257$
\textsc{uncertain} cases; and (iv) information-seeking behavior,
token-budget vs.\ performance trade-offs, per-clause difficulty,
and qualitative failure-mode case studies.

\subsection{Precision / Recall Decomposition}
\label{app:prf}

Table~\ref{tab:prf_breakdown} decomposes the five F1-style metrics
in Table~\ref{tab:overall_results} --- \textbf{M3} evidence
citation, \textbf{M5} missing-information detection, \textbf{M6}
missing-slot identification, \textbf{M7} uncertain detection, and
\textbf{M8} reportable detection --- into their underlying
precision and recall. The decomposition reveals patterns that the
F1 column alone hides:
\begin{itemize}
\item \textbf{Evidence citation (M3)} is consistently
    precision-heavy and recall-light for the frontier group:
    citation precision sits in the $74$--$89\%$ range while recall
    only reaches $52$--$68\%$. In other words, when these models
    cite a clause / recommendation / appendix identifier the cited
    item is usually correct, but each prediction \emph{under-cites}:
    it covers only a subset of the governing legal basis that a
    careful human reviewer would attach to the case.
\item \textbf{Missing-information detection (M5)} has its
    precision--recall imbalance flip by model class. Closed-source
    frontier models combine high precision with high recall;
    open-source and small models trade off (e.g.\ GPT-OSS-120B
    couples high precision with very low recall, while
    Llama-3.1-8B is the opposite); and medical-specialty models
    collapse to near-zero recall because they almost never ask.
\item \textbf{Uncertain detection (M7)} is the metric that most
    cleanly separates ``willing to abstain'' models (GPT-5.5,
    Gemini 3.1 Pro, Claude Opus 4.7) from models that rarely emit
    the \textsc{Uncertain} verdict (Mistral-Small-3.2 and all three
    medical-specialty models all sit below $4\%$ F1; cf.\ also
    Table~\ref{tab:uncertain_routing}).
\item \textbf{Reportable detection (M8)} is high across the board
    on recall, because most models default to a Reportable-leaning
    prior. The discriminating quantity is therefore M8 precision,
    on which frontier models stay high while the medical-specialty
    group drops sharply.
\end{itemize}

\begin{table*}[t]
\centering
\footnotesize
\setlength{\tabcolsep}{3pt}
\renewcommand{\arraystretch}{1.05}
\begin{tabular}{l|rrr|rrr|rrr|rrr|rrr}
\hline
\textbf{Model} & \multicolumn{3}{c|}{\textbf{M3}} & \multicolumn{3}{c|}{\textbf{M5}} & \multicolumn{3}{c|}{\textbf{M6}} & \multicolumn{3}{c|}{\textbf{M7}} & \multicolumn{3}{c}{\textbf{M8}} \\
 & \multicolumn{3}{c|}{Evidence cit.} & \multicolumn{3}{c|}{Missing det.} & \multicolumn{3}{c|}{Missing Slot ident.} & \multicolumn{3}{c|}{Uncertain det.} & \multicolumn{3}{c}{Reportable det.} \\
\cline{2-4}\cline{5-7}\cline{8-10}\cline{11-13}\cline{14-16}
 & P & R & F1 & P & R & F1 & P & R & F1 & P & R & F1 & P & R & F1 \\
\hline
\multicolumn{16}{l}{\emph{Closed-source frontier}} \\
GPT-5.5 & 81.3 & 68.5 & 74.3 & 80.5 & 84.0 & 82.2 & 89.5 & 68.0 & 77.3 & 99.4 & 66.5 & 79.7 & 85.6 & 98.7 & 91.7 \\
GPT-5 & 79.8 & 56.9 & 66.4 & 82.1 & 73.8 & 77.7 & 88.4 & 65.1 & 75.0 & 35.7 & 7.8 & 12.8 & 80.0 & 95.0 & 86.9 \\
Claude Opus 4.7 & 84.8 & 56.0 & 67.4 & 79.3 & 88.0 & 83.4 & 86.1 & 70.7 & 77.6 & 57.5 & 56.8 & 57.1 & 85.8 & 96.7 & 90.9 \\
Claude Sonnet 4.6 & 74.2 & 59.1 & 65.8 & 63.0 & 79.0 & 70.1 & 83.0 & 70.4 & 76.2 & 52.4 & 46.3 & 49.2 & 69.7 & 98.4 & 81.6 \\
Gemini 3.1 Pro & 89.0 & 51.8 & 65.5 & 86.0 & 90.6 & 88.2 & 91.6 & 66.9 & 77.4 & 85.3 & 65.4 & 74.0 & 91.1 & 97.1 & 94.0 \\
Gemini 2.5 Flash & 77.2 & 55.9 & 64.9 & 85.8 & 59.8 & 70.5 & 90.5 & 59.2 & 71.6 & 37.5 & 26.8 & 31.3 & 72.5 & 95.4 & 82.4 \\
\hline
\multicolumn{16}{l}{\emph{Open-source frontier}} \\
DeepSeek-R1 & 75.7 & 44.8 & 56.3 & 87.1 & 53.8 & 66.5 & 90.3 & 52.3 & 66.3 & 35.0 & 37.7 & 36.3 & 73.5 & 92.7 & 82.0 \\
Qwen3-235B & 71.1 & 52.8 & 60.6 & 46.5 & 66.4 & 54.7 & 78.3 & 62.1 & 69.3 & 16.2 & 21.0 & 18.3 & 63.1 & 90.7 & 74.4 \\
GPT-OSS-120B & 57.6 & 37.5 & 45.4 & 89.7 & 22.9 & 36.5 & 88.8 & 60.0 & 71.6 & 25.0 & 6.2 & 10.0 & 63.5 & 81.7 & 71.4 \\
\hline
\multicolumn{16}{l}{\emph{Small models}} \\
GPT-5-nano & 54.4 & 33.0 & 41.1 & 81.2 & 22.5 & 35.3 & 84.3 & 47.7 & 61.0 & 8.5 & 12.5 & 10.1 & 58.3 & 85.8 & 69.4 \\
Llama-3.1-8B & 50.2 & 22.1 & 30.7 & 32.0 & 86.3 & 46.7 & 24.0 & 25.2 & 24.6 & 6.4 & 29.2 & 10.5 & 49.2 & 75.3 & 59.5 \\
Mistral-Small-3.2 (24B) & 47.0 & 52.1 & 49.4 & 60.6 & 18.0 & 27.7 & 82.2 & 39.7 & 53.5 & 3.7 & 1.6 & 2.2 & 44.1 & 96.3 & 60.5 \\
\hline
\multicolumn{16}{l}{\emph{Medical-specialty}} \\
HuatuoGPT-o1 (8B) & 11.1 & 17.5 & 13.6 & 40.0 & 0.1 & 0.3 & 0.0 & 0.0 & 0.0 & 2.9 & 1.9 & 2.3 & 36.4 & 90.1 & 51.9 \\
MedGemma (27B) & 49.8 & 39.1 & 43.8 & 57.1 & 0.3 & 0.6 & 100.0 & 23.1 & 37.5 & 0.0 & 0.0 & 0.0 & 37.7 & 96.2 & 54.2 \\
HuatuoGPT-o1 (70B) & 63.0 & 43.8 & 51.7 & 72.7 & 1.2 & 2.3 & 88.2 & 41.7 & 56.6 & 8.2 & 2.3 & 3.6 & 44.2 & 96.2 & 60.5 \\
\hline
\end{tabular}
\caption{Precision / Recall / F1 breakdown for the five F1-style metrics on the full 5{,}074-case benchmark (micro aggregation, in percent). \textbf{M3} is the targeted-clause-conditional evidence-citation micro F1 over canonical clause/guidance identifiers; \textbf{M5} is missing-information detection F1; \textbf{M6} is missing-slot-identification F1 on the did-ask subset (M6); \textbf{M7} is uncertain detection F1; \textbf{M8} is reportable detection F1. Detailed definitions are given in App.~\ref{app:metrics}.}
\label{tab:prf_breakdown}
\end{table*}

\subsection{Per-Case-Type Stratification of Evidence Citation and Boundary-Condition Hit}
\label{app:m3m4-by-type}

Tables~\ref{tab:m3_by_casetype} and \ref{tab:m4_by_casetype} extend
the verdict-triage (M1) case-type stratification in
Table~\ref{tab:m1_by_casetype} to the two evidence-grounding
metrics: evidence citation (M3) and the rationale--boundary-condition
hit rate (M4). 

For both metrics, performance is broadly comparable between
\textsc{complete} and \textsc{missing} cases across models,
demonstrating that missing facts alone do not degrade clause-level
grounding. However, grounding quality generally drops on the
escalation-aware \textsc{uncertain} slice, indicating that statutory
ambiguity is harder for models to reason about than factual absence.

\begin{table}[t]
\centering
\footnotesize
\setlength{\tabcolsep}{3pt}
\renewcommand{\arraystretch}{1.05}
\begin{tabular}{lrrr}
\hline
\textbf{Model} & \textbf{Complete} & \textbf{Missing} & \textbf{Uncertain} \\
\hline
\multicolumn{4}{l}{\emph{Closed-source frontier}} \\
GPT-5.5 & 74.4 & 73.5 & 78.7 \\
GPT-5 & 66.3 & 67.2 & 39.1 \\
Claude Opus 4.7 & 67.2 & 68.4 & 63.7 \\
Claude Sonnet 4.6 & 65.5 & 67.5 & 58.8 \\
Gemini 3.1 Pro & 65.4 & 65.3 & 70.0 \\
Gemini 2.5 Flash & 64.5 & 66.7 & 57.4 \\
\hline
\multicolumn{4}{l}{\emph{Open-source frontier}} \\
DeepSeek-R1 & 55.8 & 58.0 & 54.8 \\
Qwen3-235B & 59.8 & 63.5 & 49.8 \\
GPT-OSS-120B & 44.5 & 48.8 & 44.2 \\
\hline
\multicolumn{4}{l}{\emph{Small models}} \\
GPT-5-nano & 41.0 & 42.3 & 24.2 \\
Llama-3.1-8B & 30.2 & 33.2 & 24.5 \\
Mistral-Small-3.2 (24B) & 49.4 & 49.6 & 51.6 \\
\hline
\multicolumn{4}{l}{\emph{Medical-specialty}} \\
HuatuoGPT-o1 (8B) & 13.4 & 14.2 & 29.3 \\
MedGemma (27B) & 43.3 & 45.7 & 0.0 \\
HuatuoGPT-o1 (70B) & 50.7 & 56.3 & 56.4 \\
\hline
\end{tabular}
\caption{M3 evidence-citation micro F1 stratified by case type (\%; complete = 3{,}455 cases, missing = 1{,}362, uncertain = 257). Extends Table~\ref{tab:m1_by_casetype} (M1, verdict triage accuracy) to evidence citation.}
\label{tab:m3_by_casetype}
\end{table}

\begin{table}[t]
\centering
\footnotesize
\setlength{\tabcolsep}{3pt}
\renewcommand{\arraystretch}{1.05}
\begin{tabular}{lrrr}
\hline
\textbf{Model} & \textbf{Complete} & \textbf{Missing} & \textbf{Uncertain} \\
\hline
\multicolumn{4}{l}{\emph{Closed-source frontier}} \\
GPT-5.5 & 88.6 & 87.1 & 81.8 \\
GPT-5 & 83.9 & 84.3 & 60.0 \\
Claude Opus 4.7 & 78.3 & 78.9 & 78.5 \\
Claude Sonnet 4.6 & 86.2 & 84.8 & 85.9 \\
Gemini 3.1 Pro & 78.7 & 80.0 & 63.1 \\
Gemini 2.5 Flash & 79.0 & 73.9 & 57.4 \\
\hline
\multicolumn{4}{l}{\emph{Open-source frontier}} \\
DeepSeek-R1 & 77.2 & 75.3 & 60.5 \\
Qwen3-235B & 82.4 & 80.3 & 57.8 \\
GPT-OSS-120B & 69.4 & 65.9 & 33.3 \\
\hline
\multicolumn{4}{l}{\emph{Small models}} \\
GPT-5-nano & 73.8 & 68.3 & 53.1 \\
Llama-3.1-8B & 64.0 & 59.1 & 31.5 \\
Mistral-Small-3.2 (24B) & 72.4 & 67.1 & 33.3 \\
\hline
\multicolumn{4}{l}{\emph{Medical-specialty}} \\
HuatuoGPT-o1 (8B) & 60.2 & 53.3 & 26.7 \\
MedGemma (27B) & 75.2 & 68.6 & 0.0 \\
HuatuoGPT-o1 (70B) & 68.4 & 63.5 & 44.4 \\
\hline
\end{tabular}
\caption{M4 rationale--boundary-condition LLM-judge hit rate stratified by case type (\%; same denominators as M3 modulo per-case judge eligibility).}
\label{tab:m4_by_casetype}
\end{table}

\subsection{Verdict Routing on Uncertain Cases}
\label{app:uncertain-routing}

Table~\ref{tab:uncertain_routing} reports, for each of the $257$
\textsc{uncertain} cases, the predicted verdict distribution per
model. Three groupings emerge cleanly:
\begin{itemize}
\item \textbf{Abstention-willing} (GPT-5.5, Gemini 3.1 Pro, Claude
    Opus 4.7, Claude Sonnet 4.6, DeepSeek-R1): routing to
    \textsc{Uncertain} on $37.7\%$--$66.5\%$ of the slice, while
    confining \textsc{Non-reportable} mis-routings to $\le 11\%$.
\item \textbf{Over-reporting} (GPT-5, GPT-5-nano, Qwen3-235B,
    GPT-OSS-120B, Gemini 2.5 Flash, Llama-3.1-8B,
    Mistral-Small-3.2, and all three medical-specialty models):
    routing $> 65\%$ of \textsc{uncertain} cases to
    \textsc{Reportable}; GPT-OSS-120B and Mistral-Small-3.2 push
    this to $> 75\%$, and MedGemma~27B routes $98.4\%$ to
    \textsc{Reportable} with zero \textsc{Uncertain} predictions.
\end{itemize}
The over-reporting pattern is the operational risk surfaced in
\S\ref{sec:experiments} of the main paper: a triage system that
escalates ambiguous cases to ``report'' rather than ``escalate to
human review'' floods the downstream reporting queue with cases that
the statute itself does not adjudicate, and is therefore unsafe to
deploy in either direction without additional gating.

\begin{table}[t]
\centering
\footnotesize
\setlength{\tabcolsep}{1.4pt}
\renewcommand{\arraystretch}{1.05}
\begin{tabular}{lrrr}
\hline
\textbf{Model} & \textbf{$\to$Unc.} & \textbf{$\to$Rep.} & \textbf{$\to$Non-rep.} \\
\hline
\multicolumn{4}{l}{\emph{Closed-source frontier}} \\
GPT-5.5 & 171 (66.5) & 83 (32.3) & 2 (0.8) \\
GPT-5 & 20 (7.8) & 186 (72.4) & 51 (19.8) \\
Claude Opus 4.7 & 146 (56.8) & 103 (40.1) & 8 (3.1) \\
Claude Sonnet 4.6 & 119 (46.3) & 134 (52.1) & 4 (1.6) \\
Gemini 3.1 Pro & 168 (65.4) & 61 (23.7) & 28 (10.9) \\
Gemini 2.5 Flash & 69 (26.8) & 172 (66.9) & 16 (6.2) \\
\hline
\multicolumn{4}{l}{\emph{Open-source frontier}} \\
DeepSeek-R1 & 97 (37.7) & 141 (54.9) & 19 (7.4) \\
Qwen3-235B & 54 (21.0) & 178 (69.3) & 25 (9.7) \\
GPT-OSS-120B & 16 (6.2) & 194 (75.5) & 47 (18.3) \\
\hline
\multicolumn{4}{l}{\emph{Small models}} \\
GPT-5-nano & 32 (12.5) & 213 (82.9) & 12 (4.7) \\
Llama-3.1-8B & 75 (29.2) & 167 (65.0) & 15 (5.8) \\
Mistral-Small-3.2 (24B) & 4 (1.6) & 232 (90.3) & 21 (8.2) \\
\hline
\multicolumn{4}{l}{\emph{Medical-specialty}} \\
HuatuoGPT-o1 (8B) & 5 (1.9) & 230 (89.5) & 16 (6.2) \\
MedGemma (27B) & 0 (0.0) & 253 (98.4) & 2 (0.8) \\
HuatuoGPT-o1 (70B) & 6 (2.3) & 224 (87.2) & 27 (10.5) \\
\hline
\end{tabular}
\caption{Verdict routing on the 257 \textsc{uncertain} cases: raw counts (and row percentages) of the predicted verdict for each model. Only \textsc{Uncertain} is the correct routing.}
\label{tab:uncertain_routing}
\end{table}

\subsection{Information-Seeking Behavior on Missing Cases}
\label{app:ask-behavior}

Table~\ref{tab:ask_distribution} reports the distribution of the
number of \textsc{Ask} calls each model issues on the $1{,}362$
\textsc{missing} cases, plus the per-model average. Three regimes
are visible:
\begin{itemize}
\item \textbf{Calibrated questioning} (closed-source frontier): the
    Claude, Gemini, and GPT-5.5 lines average roughly one
    \textsc{Ask} per missing case, with most cases receiving exactly
    one question. These are the models that most consistently treat
    an absent fact as a one-question lookup.
\item \textbf{Frequent skipping} (Gemini 2.5 Flash, DeepSeek-R1,
    GPT-5-nano, GPT-OSS-120B, Mistral-Small-3.2, and all three
    medical-specialty models): \textsc{Ask} is omitted on a large
    fraction of missing cases, with the medical-specialty models
    effectively never asking.
\item \textbf{Ask-indiscriminately} (Llama-3.1-8B): the model
    issues 4 or more \textsc{Ask} calls on the majority of
    \textsc{missing} cases, yet --- as the slot-identification
    column in Table~\ref{tab:prf_breakdown} shows --- recovers very
    few of the actually-missing fields, indicating that most of
    these questions fail to target the gap.
\end{itemize}

\begin{table}[t]
\centering
\footnotesize
\setlength{\tabcolsep}{2.5pt}
\renewcommand{\arraystretch}{1.05}
\begin{tabular}{lrrrrrr}
\hline
\textbf{Model} & \textbf{0} & \textbf{1} & \textbf{2} & \textbf{3} & \textbf{$\geq$4} & \textbf{Avg.} \\
\hline
\multicolumn{7}{l}{\emph{Closed-source frontier}} \\
GPT-5.5 & 16.0 & 71.1 & 9.2 & 2.3 & 1.3 & 1.02 \\
GPT-5 & 26.2 & 57.1 & 9.0 & 3.7 & 4.0 & 1.06 \\
Claude Opus 4.7 & 12.0 & 63.4 & 17.9 & 4.5 & 2.2 & 1.22 \\
Claude Sonnet 4.6 & 21.0 & 46.5 & 19.1 & 9.7 & 3.7 & 1.31 \\
Gemini 3.1 Pro & 9.4 & 79.4 & 9.4 & 1.2 & 0.6 & 1.04 \\
Gemini 2.5 Flash & 40.2 & 54.2 & 4.9 & 0.6 & 0.1 & 0.66 \\
\hline
\multicolumn{7}{l}{\emph{Open-source frontier}} \\
DeepSeek-R1 & 46.2 & 52.7 & 1.1 & 0.0 & 0.0 & 0.55 \\
Qwen3-235B & 33.6 & 41.0 & 15.7 & 6.8 & 2.9 & 1.07 \\
GPT-OSS-120B & 77.1 & 20.9 & 1.4 & 0.4 & 0.3 & 0.26 \\
\hline
\multicolumn{7}{l}{\emph{Small models}} \\
GPT-5-nano & 77.5 & 17.8 & 1.4 & 0.4 & 2.9 & 0.44 \\
Llama-3.1-8B & 13.7 & 0.2 & 4.8 & 21.2 & 60.0 & 3.70 \\
Mistral-Small-3.2 (24B) & 82.0 & 16.7 & 1.1 & 0.1 & 0.1 & 0.20 \\
\hline
\multicolumn{7}{l}{\emph{Medical-specialty}} \\
HuatuoGPT-o1 (8B) & 99.9 & 0.0 & 0.1 & 0.0 & 0.0 & 0.00 \\
MedGemma (27B) & 99.7 & 0.2 & 0.1 & 0.0 & 0.0 & 0.00 \\
HuatuoGPT-o1 (70B) & 98.8 & 1.2 & 0.0 & 0.0 & 0.0 & 0.01 \\
\hline
\end{tabular}
\caption{Distribution of the number of \textsc{Ask} calls per model on the $1{,}362$ \textsc{missing} cases. Columns $0$, $1$, $2$, $3$, $\geq\!4$ report the percentage of missing cases that received that many \textsc{Ask} calls from the evaluated LLM. ``Avg.'' is the per-model average number of \textsc{Ask} calls.}
\label{tab:ask_distribution}
\end{table}

\subsection{Token Budget vs Verdict Accuracy}
\label{app:token-vs-perf}

\begin{figure}[H]
\centering
\includegraphics[width=\linewidth]{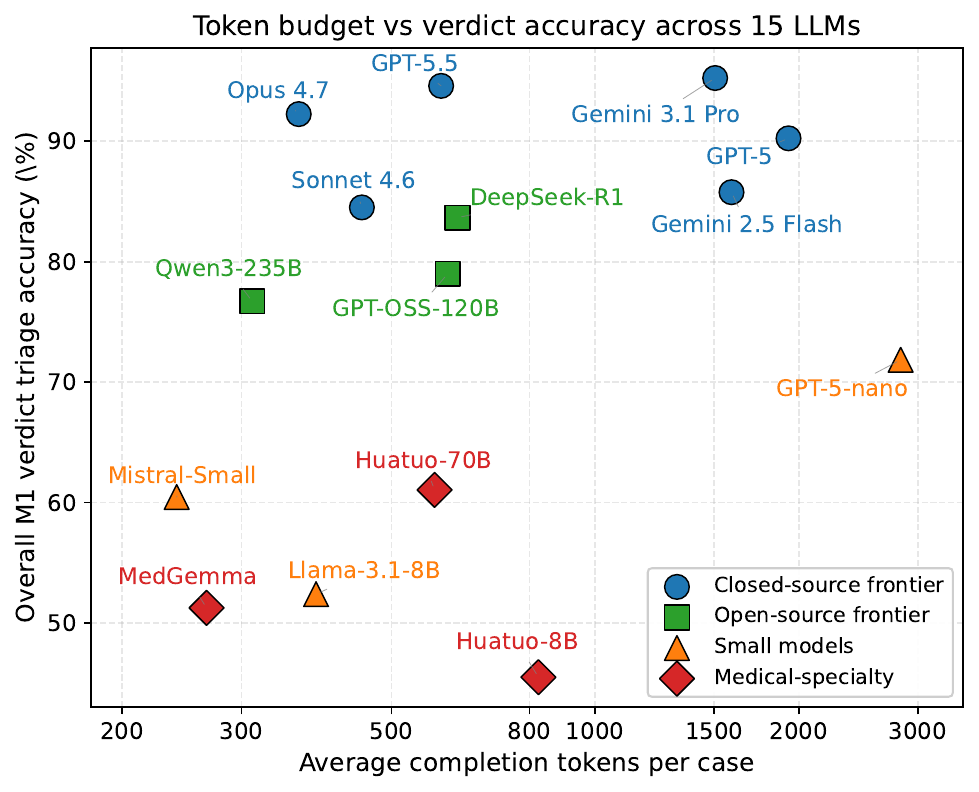}
\caption{Average completion tokens per case
vs.\ verdict-triage accuracy.
``Completion tokens'' counts only the evaluated-LLM side.}
\label{fig:token_vs_perf}
\end{figure}

Figure~\ref{fig:token_vs_perf} plots, for each of the $15$ evaluated
LLMs, the average completion tokens it spends per case (eval-side,
log scale; the horizontal axis grows from left to right) against
its overall verdict-triage accuracy (M1). 

\textbf{Larger token budgets are not the limiting factor for
verdict accuracy.} The two strongest models, Gemini 3.1 Pro and
GPT-5.5, sit in very different token-consumption regimes, and the
highest-consuming model in our pool, GPT-5-nano, nevertheless ranks
in the bottom half. Within the open-source frontier group,
DeepSeek-R1 and GPT-OSS-120B consume similar tokens yet differ
non-trivially in M1; within the medical-specialty group,
HuatuoGPT-o1 (8B) uses several times as many tokens as MedGemma
(27B) yet trails it in M1.


\subsection{Per-MN29-Clause Difficulty}
\label{app:per-clause}

\begin{figure}[H]
\centering
\includegraphics[width=\linewidth]{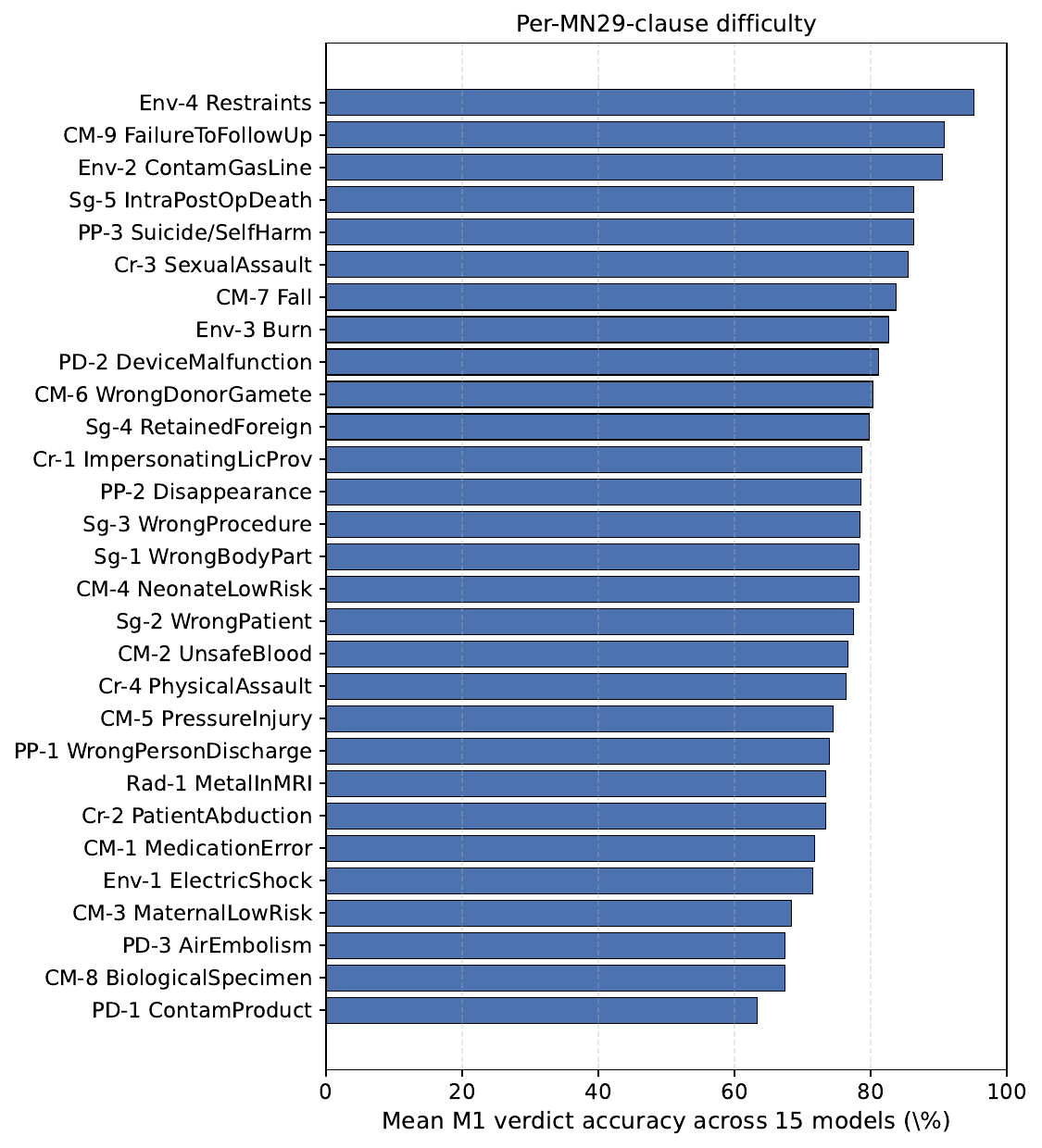}
\caption{Per-MN29-clause difficulty ranking. Each bar reports the mean verdict-triage accuracy, averaged across the $15$
models.}
\label{fig:per_clause_difficulty}
\end{figure}

Figure~\ref{fig:per_clause_difficulty} ranks the $29$ MN29
sub-clauses by their mean verdict-triage accuracy (M1) averaged
across the $15$ models.

The five hardest sub-clauses are
\emph{ProductDevice-1 (Contaminated Product)},
\emph{CareManagement-8 (Biological Specimen)},
\emph{ProductDevice-3 (Intravascular Air Embolism)},
\emph{CareManagement-3 (Maternal Low-risk Pregnancy)}, and
\emph{Environmental-1 (Electric Shock)}. These clauses share a
common structural property: their reportability hinges on
multi-condition boundary criteria with non-trivial \emph{exclusions}
(e.g.\ ``contaminated product where the contamination is associated
with patient death or serious injury''), which models routinely
satisfy partially and then commit to the wrong verdict.

The five easiest sub-clauses are
\emph{Environmental-4 (Restraints or Bedrails)},
\emph{CareManagement-9 (Failure to Follow Up / Communicate Test
Results)},
\emph{Environmental-2 (Wrong or Contaminated Gas Line)},
\emph{Surgical-5 (Intra/Post-Op Death of Normal Healthy Patient)},
and
\emph{PatientProtection-3 (Suicide / Attempted Suicide /
Self-Harm)}. These clauses have crisp, single-condition triggers
that are explicit in the narrative.

\section{Expert Review Questionnaire}
\label{app:expert-questionnaire}

\begin{figure*}[p]
\centering
\includegraphics[width=0.82\linewidth]{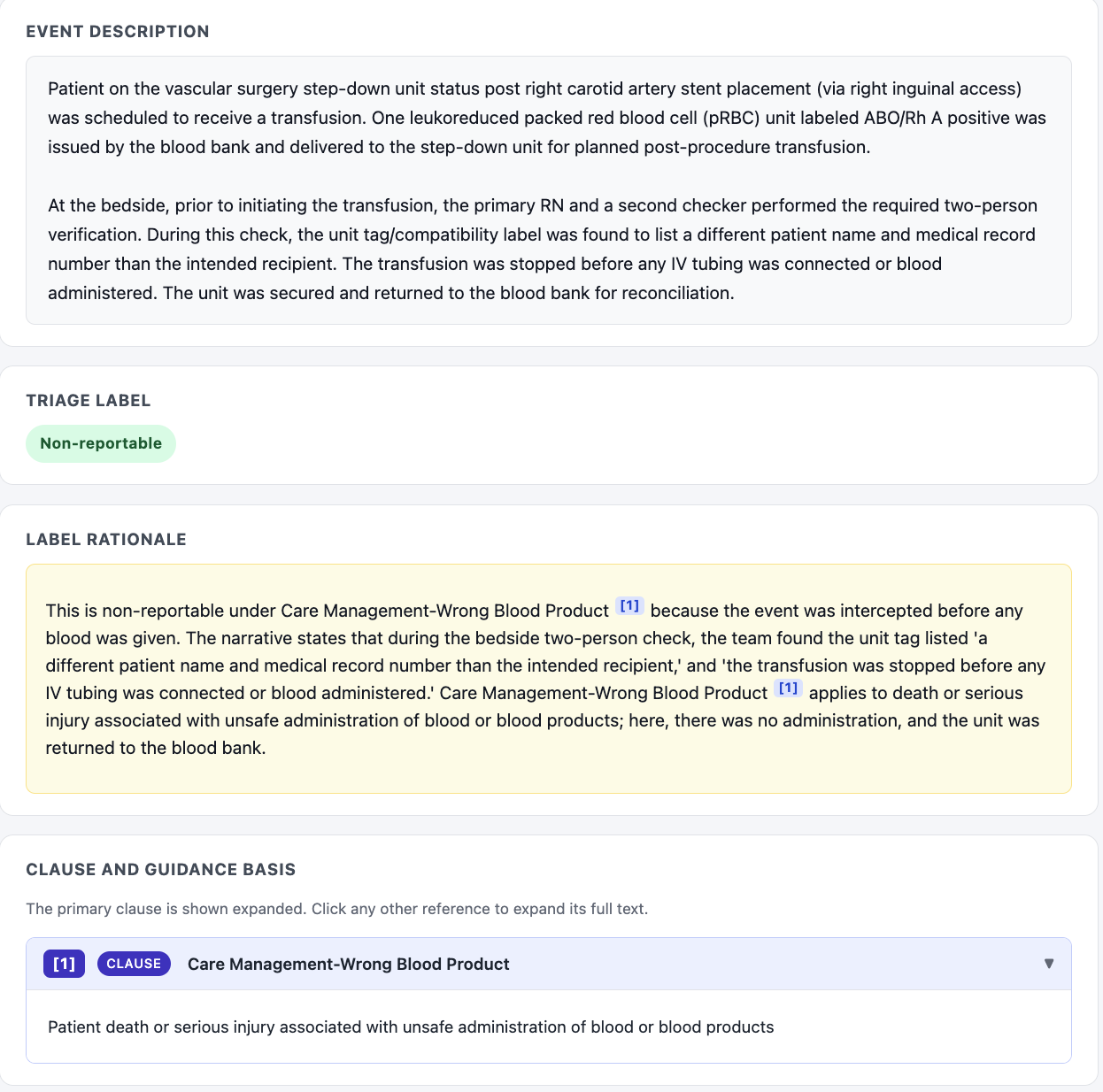}
\caption{Expert-review questionnaire, Complete Case example screenshot part 1.}
\label{fig:expert-questionnaire-p1}
\end{figure*}

\begin{figure*}[p]
\centering
\includegraphics[width=0.82\linewidth]{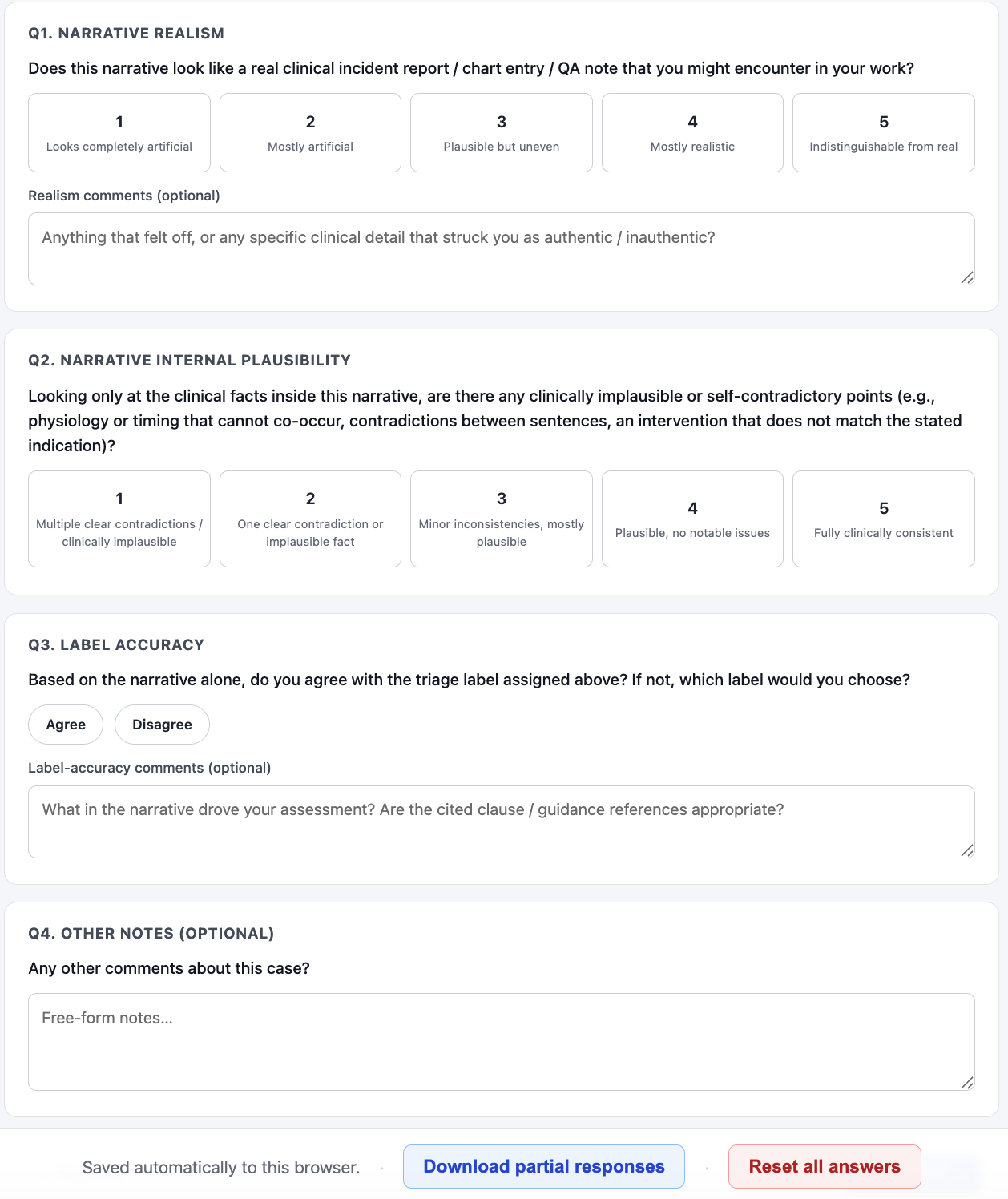}
\caption{Expert-review questionnaire, Complete Case example screenshot part 2.}
\label{fig:expert-questionnaire-p2}
\end{figure*}

To validate benchmark quality, two patient safety experts jointly
reviewed a stratified sample of $90$ generated cases ($30$ complete,
$30$ missing, $30$ uncertain) using a dedicated web questionnaire.
The experts rated each case on a $1$--$5$ scale
for clinical realism and internal plausibility, and indicated whether
they agreed with the pipeline's by-construction ground-truth verdict
(\textsc{Reportable}, \textsc{Non\_Reportable}, \textsc{Missing}, or
\textsc{Uncertain}). The aggregate ratings are reported in the main
paper. Figures~\ref{fig:expert-questionnaire-p1}--\ref{fig:expert-questionnaire-p2}
exhibit the example screenshot of the questionnaire used during
this review.

\end{document}